\documentclass[11pt]{article}

\usepackage[margin=1in]{geometry}
\usepackage{amsmath,amssymb}
\usepackage{tikz}
\usetikzlibrary{arrows.meta,positioning,shapes.geometric,calc}
\usepackage{graphicx}
\graphicspath{{./}{figures/}}
\usepackage{booktabs}
\usepackage{longtable}
\usepackage{amsfonts}
\usepackage{array}
\usepackage{tabularx}
\usepackage{xcolor}
\usepackage{hyperref}
\usepackage[super,sort&compress,comma]{natbib}
\usepackage{mathptmx}
\usepackage{lineno}
\usepackage{setspace}
\usepackage{microtype}
\usepackage{caption}


\hypersetup{
colorlinks=true,
linkcolor=blue!70!black,
citecolor=blue!70!black,
urlcolor=blue!70!black
}

\newcommand{\HyperionCorpusRows}{14{,}518{,}531}
\newcommand{\HyperionOmegaLeaves}{249}
\newcommand{\HyperionXiLeaves}{1{,}001}

\newcommand{\HyperionCellsForHalf}{23}

\newcommand{\HyperionCoarseOccupied}{99.3\%}
\newcommand{\HyperionFineEmpty}{67.2\%}

\newcommand{\HyperionVTwentyOneCorpusRows}{14{,}518{,}531}
\newcommand{\HyperionVTwentyOneOmegaStates}{32}
\newcommand{\HyperionVTwentyOneXiStates}{18}
\newcommand{\HyperionVTwentyOneOmegaNonNullStates}{31}
\newcommand{\HyperionVTwentyOneXiNonNullStates}{17}
\newcommand{\HyperionVTwentyOneOmegaLeaves}{249}
\newcommand{\HyperionVTwentyOneXiLeaves}{1{,}001}
\newcommand{\HyperionVTwentyOneCellsForHalf}{23}
\newcommand{\HyperionVTwentyOneOccupiedFine}{81{,}875}
\newcommand{\HyperionVTwentyOneFineEmpty}{67.2\%}
\newcommand{\HyperionVTwentyOneCoarseOccupied}{99.3\%}
\newcommand{\HyperionVTwentyOneCellsForHalfFraction}{0.03\%}
\newcommand{\HyperionVTwentyOneCellsForHalfMinimum}{21}
\newcommand{\HyperionVTwentyOneCellsForHalfMaximum}{25}
\newcommand{\HyperionVTwentyOneHalfCoreJaccard}{95.7\%}
\newcommand{\HyperionVTwentyOneMechanismRows}{4{,}355{,}138}
\newcommand{\HyperionVTwentyOneMechanismOccupiedFine}{80{,}658}
\newcommand{\HyperionVTwentyOneMechanismCellsForHalf}{328}
\newcommand{\HyperionVTwentyOneMechanismFineEmpty}{67.3\%}
\newcommand{\HyperionVTwentyOneMechanismCellsForHalfMinimum}{310}
\newcommand{\HyperionVTwentyOneMechanismCellsForHalfMaximum}{348}
\newcommand{\HyperionVTwentyOneMechanismHalfCoreJaccard}{91.4\%}
\newcommand{\HyperionVTwentyOneMechanismFraction}{30.00\%}

\renewcommand{\HyperionCorpusRows}{\HyperionVTwentyOneCorpusRows}
\renewcommand{\HyperionOmegaLeaves}{\HyperionVTwentyOneOmegaLeaves}
\renewcommand{\HyperionXiLeaves}{\HyperionVTwentyOneXiLeaves}

\renewcommand{\HyperionCellsForHalf}{\HyperionVTwentyOneCellsForHalf}
\renewcommand{\HyperionCoarseOccupied}{\HyperionVTwentyOneCoarseOccupied}
\renewcommand{\HyperionFineEmpty}{\HyperionVTwentyOneFineEmpty}

\providecommand{\HyperionVTwentyOneMechanismRows}{4{,}355{,}138}
\providecommand{\HyperionVTwentyOneMechanismCellsForHalf}{328}
\providecommand{\HyperionVTwentyOneMechanismCellsForHalfMinimum}{310}
\providecommand{\HyperionVTwentyOneMechanismCellsForHalfMaximum}{348}
\providecommand{\HyperionVTwentyOneMechanismHalfCoreJaccard}{91.4\%}
\newif\ifHyperionVTwentyOneRoadMetrics
\HyperionVTwentyOneRoadMetricstrue
\newcommand{\HyperionVTwentyOneRoadEdges}{250{,}000}
\newcommand{\HyperionVTwentyOneRoadPapers}{18{,}915}
\newcommand{\HyperionVTwentyOneOmegaRoadRetention}{74.7\%}
\newcommand{\HyperionVTwentyOneOmegaRoadNull}{24.9\%}
\newcommand{\HyperionVTwentyOneXiRoadRetention}{71.9\%}
\newcommand{\HyperionVTwentyOneXiRoadNull}{15.8\%}

\newcommand{\HyperionVTwentyOneOperatorEdges}{1{,}000{,}000}
\newcommand{\HyperionVTwentyOneCarrierEdges}{1{,}000{,}000}
\newcommand{\HyperionVTwentyOneCoupledEdges}{9{,}885}
\newcommand{\HyperionVTwentyOneOperatorTransferEdges}{1{,}000{,}000}
\newcommand{\HyperionVTwentyOneCarrierTransferEdges}{1{,}000{,}000}
\newcommand{\HyperionVTwentyOneControlEdges}{1{,}000{,}000}
\newcommand{\HyperionVTwentyOneSequenceEdges}{323{,}894}
\newcommand{\HyperionVTwentyOneDAGDirectedEdges}{1{,}550{,}834}
\newcommand{\HyperionVTwentyOneDAGInputEdges}{4{,}333{,}779}
\newcommand{\HyperionVTwentyOneDAGDirectedFraction}{35.78\%}
\newcommand{\HyperionVTwentyOneDAGLayerZero}{2{,}242{,}098}
\newcommand{\HyperionVTwentyOneDAGLayerZeroFraction}{15.44\%}
\newcommand{\HyperionVTwentyOneDAGLayerOne}{5{,}019{,}024}
\newcommand{\HyperionVTwentyOneDAGLayerOneFraction}{34.57\%}
\newcommand{\HyperionVTwentyOneDAGLayerTwo}{4{,}470{,}319}
\newcommand{\HyperionVTwentyOneDAGLayerTwoFraction}{30.79\%}
\newcommand{\HyperionVTwentyOneDAGLayerThree}{2{,}244{,}019}
\newcommand{\HyperionVTwentyOneDAGLayerThreeFraction}{15.46\%}
\newcommand{\HyperionVTwentyOneDAGLayerFour}{525{,}937}
\newcommand{\HyperionVTwentyOneDAGLayerFourFraction}{3.62\%}
\newcommand{\HyperionVTwentyOneDAGLayerFive}{17{,}134}
\newcommand{\HyperionVTwentyOneDAGLayerFiveFraction}{0.12\%}
\providecommand{\HyperionVTwentyOneDAGLayerFive}{17{,}134}
\providecommand{\HyperionVTwentyOneDAGLayerFiveFraction}{0.12\%}
\providecommand{\HyperionVTwentyOneMechanismOccupiedFine}{80{,}658}
\providecommand{\HyperionVTwentyOneMechanismFineEmpty}{67.3\%}

\providecommand{\Rspace}{\mathbb{R}}

\providecommand{\Om}{\Omega}
\providecommand{\Xis}{\Xi}
\providecommand{\Lam}{\Lambda}
\providecommand{\Gam}{\Gamma}
\providecommand{\Tau}{\mathsf{T}}
\providecommand{\Cur}{\mathsf{J}}

\providecommand{\HyperionCorpusRows}{14{,}518{,}531}
\providecommand{\HyperionOmegaLeaves}{249}
\providecommand{\HyperionXiLeaves}{1{,}001}

\providecommand{\HyperionCellsForHalf}{23}

\providecommand{\HyperionCoarseOccupied}{99.3\%}
\providecommand{\HyperionFineEmpty}{67.2\%}

\providecommand{\HyperionVTwentyOneCorpusRows}{14{,}518{,}531}
\providecommand{\HyperionVTwentyOneOmegaStates}{32}
\providecommand{\HyperionVTwentyOneXiStates}{18}
\providecommand{\HyperionVTwentyOneOmegaNonNullStates}{31}
\providecommand{\HyperionVTwentyOneXiNonNullStates}{17}
\providecommand{\HyperionVTwentyOneOmegaLeaves}{249}
\providecommand{\HyperionVTwentyOneXiLeaves}{1{,}001}
\providecommand{\HyperionVTwentyOneCellsForHalf}{23}
\providecommand{\HyperionVTwentyOneOccupiedFine}{81{,}875}
\providecommand{\HyperionVTwentyOneFineEmpty}{67.2\%}
\providecommand{\HyperionVTwentyOneCoarseOccupied}{99.3\%}
\providecommand{\HyperionVTwentyOneCellsForHalfFraction}{0.03\%}
\providecommand{\HyperionVTwentyOneCellsForHalfMinimum}{21}
\providecommand{\HyperionVTwentyOneCellsForHalfMaximum}{25}
\providecommand{\HyperionVTwentyOneHalfCoreJaccard}{95.7\%}
\providecommand{\HyperionVTwentyOneMechanismRows}{4{,}355{,}138}
\providecommand{\HyperionVTwentyOneMechanismOccupiedFine}{80{,}658}
\providecommand{\HyperionVTwentyOneMechanismCellsForHalf}{328}
\providecommand{\HyperionVTwentyOneMechanismFineEmpty}{67.3\%}
\providecommand{\HyperionVTwentyOneMechanismCellsForHalfMinimum}{310}
\providecommand{\HyperionVTwentyOneMechanismCellsForHalfMaximum}{348}
\providecommand{\HyperionVTwentyOneMechanismHalfCoreJaccard}{91.4\%}
\providecommand{\HyperionVTwentyOneMechanismFraction}{30.00\%}

\providecommand{\HyperionVTwentyOneMechanismRows}{4{,}355{,}138}
\providecommand{\HyperionVTwentyOneMechanismCellsForHalf}{328}
\providecommand{\HyperionVTwentyOneMechanismCellsForHalfMinimum}{310}
\providecommand{\HyperionVTwentyOneMechanismCellsForHalfMaximum}{348}
\providecommand{\HyperionVTwentyOneMechanismHalfCoreJaccard}{91.4\%}
\providecommand{\HyperionVTwentyOneMechanismFraction}{30.00\%}
\providecommand{\HyperionVTwentyOneMechanismOccupiedFine}{80{,}658}
\providecommand{\HyperionVTwentyOneMechanismFineEmpty}{67.3\%}
\newif\ifHyperionVTwentyOneRoadMetrics
\HyperionVTwentyOneRoadMetricstrue
\providecommand{\HyperionVTwentyOneRoadEdges}{250{,}000}
\providecommand{\HyperionVTwentyOneRoadPapers}{18{,}915}
\providecommand{\HyperionVTwentyOneOmegaRoadRetention}{74.7\%}
\providecommand{\HyperionVTwentyOneOmegaRoadNull}{24.9\%}
\providecommand{\HyperionVTwentyOneXiRoadRetention}{71.9\%}
\providecommand{\HyperionVTwentyOneXiRoadNull}{15.8\%}

\newif\ifHyperionVTwentyOneDAGMetrics
\HyperionVTwentyOneDAGMetricstrue
\providecommand{\HyperionVTwentyOneOperatorEdges}{1{,}000{,}000}
\providecommand{\HyperionVTwentyOneCarrierEdges}{1{,}000{,}000}
\providecommand{\HyperionVTwentyOneCoupledEdges}{9{,}885}
\providecommand{\HyperionVTwentyOneOperatorTransferEdges}{1{,}000{,}000}
\providecommand{\HyperionVTwentyOneCarrierTransferEdges}{1{,}000{,}000}
\providecommand{\HyperionVTwentyOneControlEdges}{1{,}000{,}000}
\providecommand{\HyperionVTwentyOneSequenceEdges}{323{,}894}
\providecommand{\HyperionVTwentyOneDAGDirectedEdges}{1{,}550{,}834}
\providecommand{\HyperionVTwentyOneDAGInputEdges}{4{,}333{,}779}
\providecommand{\HyperionVTwentyOneDAGDirectedFraction}{35.78\%}
\providecommand{\HyperionVTwentyOneDAGLayerZero}{2{,}242{,}098}
\providecommand{\HyperionVTwentyOneDAGLayerZeroFraction}{15.44\%}
\providecommand{\HyperionVTwentyOneDAGLayerOne}{5{,}019{,}024}
\providecommand{\HyperionVTwentyOneDAGLayerOneFraction}{34.57\%}
\providecommand{\HyperionVTwentyOneDAGLayerTwo}{4{,}470{,}319}
\providecommand{\HyperionVTwentyOneDAGLayerTwoFraction}{30.79\%}
\providecommand{\HyperionVTwentyOneDAGLayerThree}{2{,}244{,}019}
\providecommand{\HyperionVTwentyOneDAGLayerThreeFraction}{15.46\%}
\providecommand{\HyperionVTwentyOneDAGLayerFour}{525{,}937}
\providecommand{\HyperionVTwentyOneDAGLayerFourFraction}{3.62\%}
\providecommand{\HyperionVTwentyOneDAGLayerFive}{17{,}134}
\providecommand{\HyperionVTwentyOneDAGLayerFiveFraction}{0.12\%}
\newif\ifHyperionVTwentyOneDefectFateMetrics
\HyperionVTwentyOneDefectFateMetricsfalse
\newif\ifHyperionVTwentyOnePredictionMetrics
\HyperionVTwentyOnePredictionMetricsfalse

\title{Constructing physical mechanisms with artificial intelligence}

\author{
Vladimir A. Baulin$^{1,*}$,
Austin Cook$^{1}$,
\and Shagor Rahman$^{1}$
}
\date{}

\begin{document}
\maketitle

\begin{center}
\vspace{-6mm}
\noindent $^{1}$Synthetix Institute, Keizersgracht 241, 1016 EA Amsterdam, The Netherlands\\
$^*$Corresponding author: \texttt{vbaulin@synthetix.institute}
\end{center}

\begin{abstract}
\noindent
Mechanisms recur across the sciences: the equation for heat in a solid also describes molecular diffusion or a Brownian particle. A mechanism is rarely written in one equation; its operation, boundaries, parameters and closures are stated at different steps of a derivation, where methods that read text or fit formulas miss it. We built an equation-native system that reads the equations of 2.5 million arXiv manuscripts, and show that mechanisms can be compiled from them and constructed by deterministic calculation. A constructor detaches, transfers, unfolds, composes and joins their parts across disciplines; where parts do not fit, the failure is an equation for the missing term. Constructions turn a threshold reactor into a memory that a weak signal writes, an oscillator into a switch through two changes that each fail alone, and a rhythm into the regulator of a chemical amount, each with conditions for an experimental test.
\end{abstract}

Science is organized by what it studies: disciplines, materials and regimes of phenomena. The division may look practical, since each object demands its own experiments and its own vocabulary, and it decides where a result is published and who reads it. This convenience comes at the price of fragmented knowledge and isolated domains. The concepts that have lasted longest, however, are defined by what does not depend on the object: they are shared across disciplines and can be generalized by removing details. An electron is identified by its quantum numbers and by the way its state transforms, wherever it is found. A thermodynamic phase comprises many microscopic arrangements with one macroscopic order, and its laws are largely insensitive to microscopic detail~\cite{Anderson1972}. Potentials related by a gauge transformation describe the same observables, and the freedom to choose the local phase of a field, first taken as a convention, turned out to be the origin of the fields through which particles interact~\cite{YangMills1954,AharonovBohm1959}. A coordinate chart is one local description of a geometry, and only what survives a change of chart is physical. In each case the concept is what remains when the description is changed, and for this reason it can be recognized in systems that share no constituents: the same symmetry groups organize the spectra of nuclei, atoms and solids~\cite{Wigner1960}.

In 1855 Adolf Fick obtained the law by which salt diffuses in water by copying Fourier's equation for the flow of heat in a solid~\cite{Fourier1822,Fick1855}. Fifty years later Einstein found the same equation in the probability of finding a Brownian particle, and used it to measure the size of molecules~\cite{Einstein1905}. Renormalization later gave one reason why such transfers succeed near a critical point, where materials which differ microscopically flow to the same fixed points~\cite{Kadanoff1966,Wilson1971,WilsonKogut1974}. What moved from heat to salt and from salt to probability was a mechanism: one operation, the Laplacian, acting on a temperature, a concentration and a probability, so that the three systems are realizations of it.

We built a system native to equations rather than to text and read with it the equations of 2.5 million arXiv manuscripts, the \LaTeX{} sources of two public datasets, each symbol resolved against its definition and each equation linked to those it is derived from. Read in this way, what a paper records separates into three layers. The first is attestation: who reported what, with which claims and which data. Citation networks, ontology graphs, retrieval engines and language models work in this surface layer~\cite{Fortunato2018Science,Lo2020S2ORC}. The training pairs of a language model are neighbouring parts of one document, and representations learned from such pairs approach a shared description of the world only slowly, with scale~\cite{Huh2024Platonic,Balestriero2026Platonic}. The second layer is the mathematical representation: the equations and the protocols by which they were obtained and tested. The form of the equations changes with notation and coordinates, so that one physical process is written in many forms, and equations that are written alike need not act alike; an equation is therefore not a stable unit of knowledge. Various tools try to reproduce this layer, from symbolic regression to the sparse identification of dynamics and physics-informed networks; they search for the right formula among variables chosen in advance~\cite{Schmidt2009SR,Brunton2016SINDy,Raissi2019PINN,Cornelio2023Descartes,RomeraParedes2024Nature}.

The third layer contains the mechanisms: what the equations do, once boundaries, the measured quantity and the drive have made the operation definite. This layer is not visible in the text, because authors state the operation in one equation, the boundary in another and the measurement in a third, each fixed at a different step of the work, so that a mechanism is assembled along a derivation chain and no single sentence or formula contains it. Methods that work in the first two layers recover what authors wrote and how they interpreted it. A mechanism is more abstract and more stable: it does not depend on the authors' interpretation, it can be moved to a new system, and it connects isolated studies in one picture even when their equations and words differ. Statements change with their authors, and equations change with notation and coordinates, as a chart changes over one geometry. A mechanism is what remains when the description is changed, and like any invariant it can be written only through one of its descriptions.

Scientists reach new mechanisms starting with models they already know, and derivations follow traditions and habits: consecutive equations of a derivation keep their state space 4.6 times and their operation 3.0 times more often than when papers are shuffled, and the kind of the next step is predictable, whereas the equation that it reaches is predicted no better than by its frequency (Supplementary Sections~S11, S18 and S19). A model is explored along the control that its authors chose, and over the range that interested them. Most studies of Lorenz's model of convection~\cite{Lorenz1963} follow the chaos far above the onset of convection, and a study that sweeps the Rayleigh parameter only through that range does not meet the pitchfork at which convection begins, and with it the choice between two senses of rotation, although its equations contain it. Changes are made one at a time and kept only when each has an effect on its own, although two changes that fail separately can together create a switch. Chemical kinetics and population dynamics evolve amounts that cannot become negative~\cite{Schlogl1972,Gardner2000Toggle,Briat2016,Aoki2019Integral}, oscillator theory evolves phases that return to themselves after a full turn~\cite{Adler1946}, and because each discipline keeps to its own state space, mechanisms that need both are rarely written. A search, finally, begins from a target that someone wants, so that a mechanism which answers no posed question is not sought. The mechanisms that remain unwritten are hidden less by nature than by the routes that lead to the written ones.

We search these manuscripts for mechanisms, reconstruct each in a form that can be carried to another system, and compile those whose equations close, a much smaller set, into executable objects. Such an object can be reused whatever the notation and the discipline in which it was found. Together the objects form a library of parts, operations, carriers, boundaries, observables, protocols and materials, taken from different fields and written in one form, and the library shows which parts have never been combined. A constructor builds new models from it, one change of a part at a time. It detaches a part from one model and attaches it to another, unfolds a protocol, composes changes and joins state spaces, and it calculates the physics that results: whether the mechanism holds in the changed system, which term is missing when it does not, and what law the completed model obeys. The operations on these objects are openly executable in the workbench FieldBridge~\cite{FieldBridge2026}.

\begin{figure}[!htbp]
\centering
\includegraphics[width=0.80\textwidth]{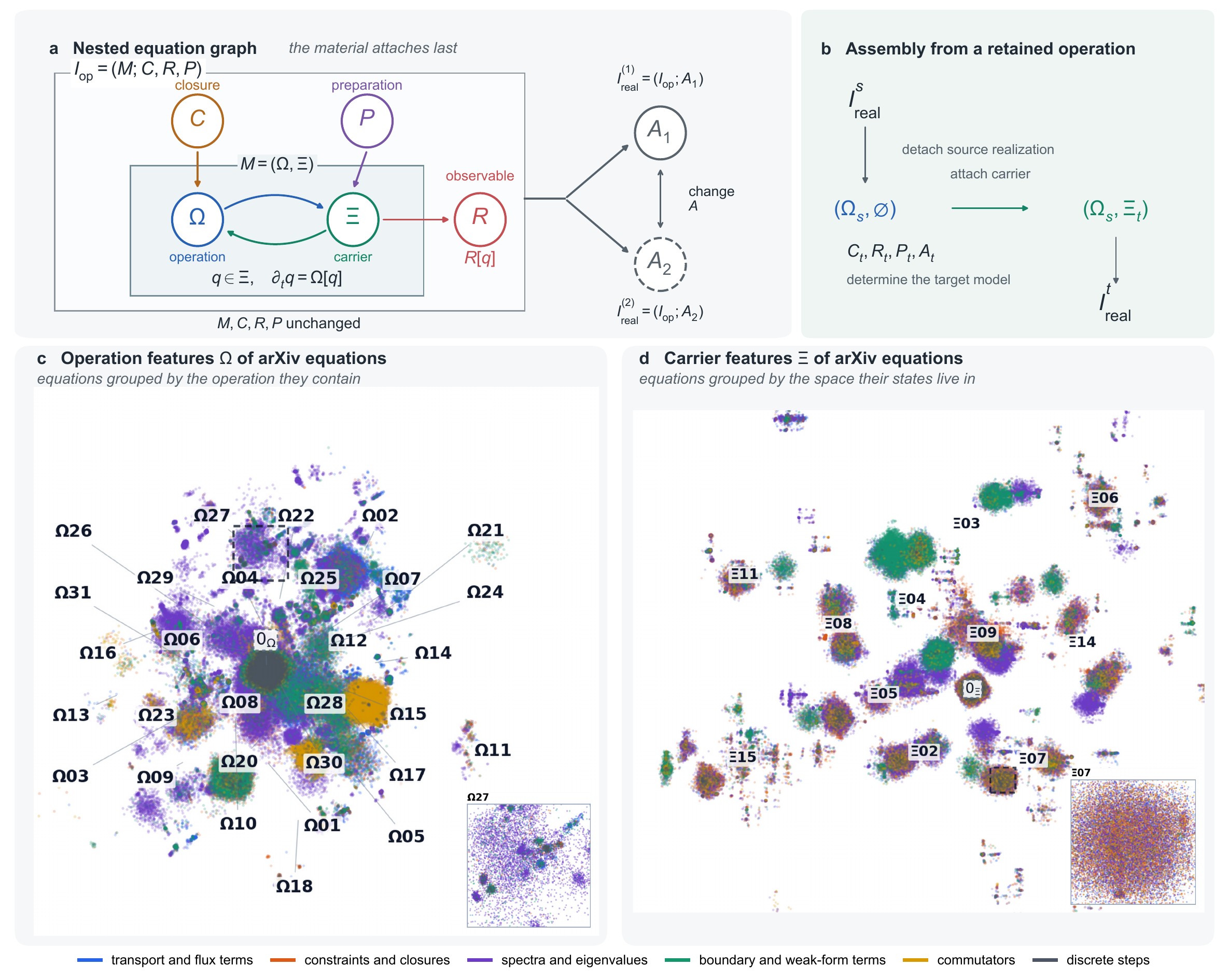}\\[2pt]
\includegraphics[width=0.74\textwidth]{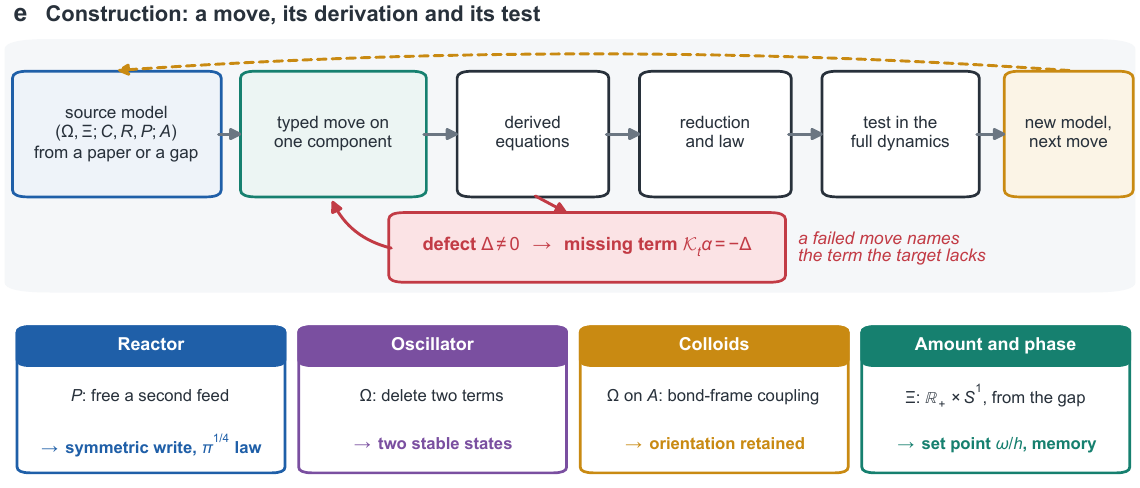}
\caption{\textbf{A mechanism is a nested object, and the written record can be read in its parts.}
\textbf{a}, The core $M=(\Omega,\Xi)$ pairs an operation $\Omega$ with the carrier $\Xi$ of the states it evolves, $\partial_t q=\Omega[q]$ with $q\in\Xi$. Boundary and constitutive relations $C$, the preparation and drive $P$ and the observable $R$ complete it to $I_{\mathrm{op}}=(M;C,R,P)$. The material attaches last: exchanging $A_1$ for $A_2$ changes the realization $I_{\mathrm{real}}=(I_{\mathrm{op}};A)$ and leaves $M$, $C$, $R$ and $P$ as they were, as long as both materials share the closure, the boundaries of the domain and the regime of validity. Outside this regime the new material brings couplings, slow modes or escape barriers that $I_{\mathrm{op}}$ does not represent, and $I_{\mathrm{op}}$ must be revised.
\textbf{b}, An operation kept from a source model is detached from its realization, attached to a new carrier $\Xi_t$ and completed by the target's own relations, observable, drive and material.
\textbf{c,d}, Each point is an equation from arXiv, placed by the operation (\textbf{c}) or the carrier (\textbf{d}) recognized in its text. Labels name the fitted classes, colours give the kind of term the equation contains, and the insets enlarge the boxed classes; the layout is described in Supplementary Section~S8.
\textbf{e}, A construction changes one part of a source model, derives the new equations, reduces them to a law and tests the law in the full dynamics. When a move fails, its defect $\Delta\neq0$ fixes the missing term through $\mathcal{K}_t\alpha=-\Delta$. Below are the four constructions of this work, the part each one changes and what it produces.}
\label{fig:atlas}
\end{figure}

\section*{Results}

\subsection*{The structure of a mechanism}

In heat conduction along a bar of length $L$, $\partial_t u = D\,\partial_x^2 u$, the operation $\Omega = D\,\partial_x^2$ alone does not determine what the bar does. If the bar is insulated at its ends ($\partial_x u=0$), the evolution has cosine modes with a zero eigenvalue, total heat is conserved and the bar relaxes to its mean temperature. If the ends touch an ice bath ($u=0$), the modes are sines, the slowest decays at rate $D(\pi/L)^2$ and all heat leaks out. The symbol is the same; conservation in one case and loss in the other are fixed only when the operation is joined to its carrier of states and closed by its boundary.

A change of the constitutive relation acts in the same way. If the heat flux relaxes over a finite collision time, instead of following the temperature gradient at once, the flux becomes a dynamical variable, diffusion becomes a damped wave, and the initial rate of change of the field must be prepared together with the field itself: a change in one part of the description forces changes in the others (Methods).

\begin{figure}[!htbp]
\centering
\includegraphics[width=0.87\textwidth]{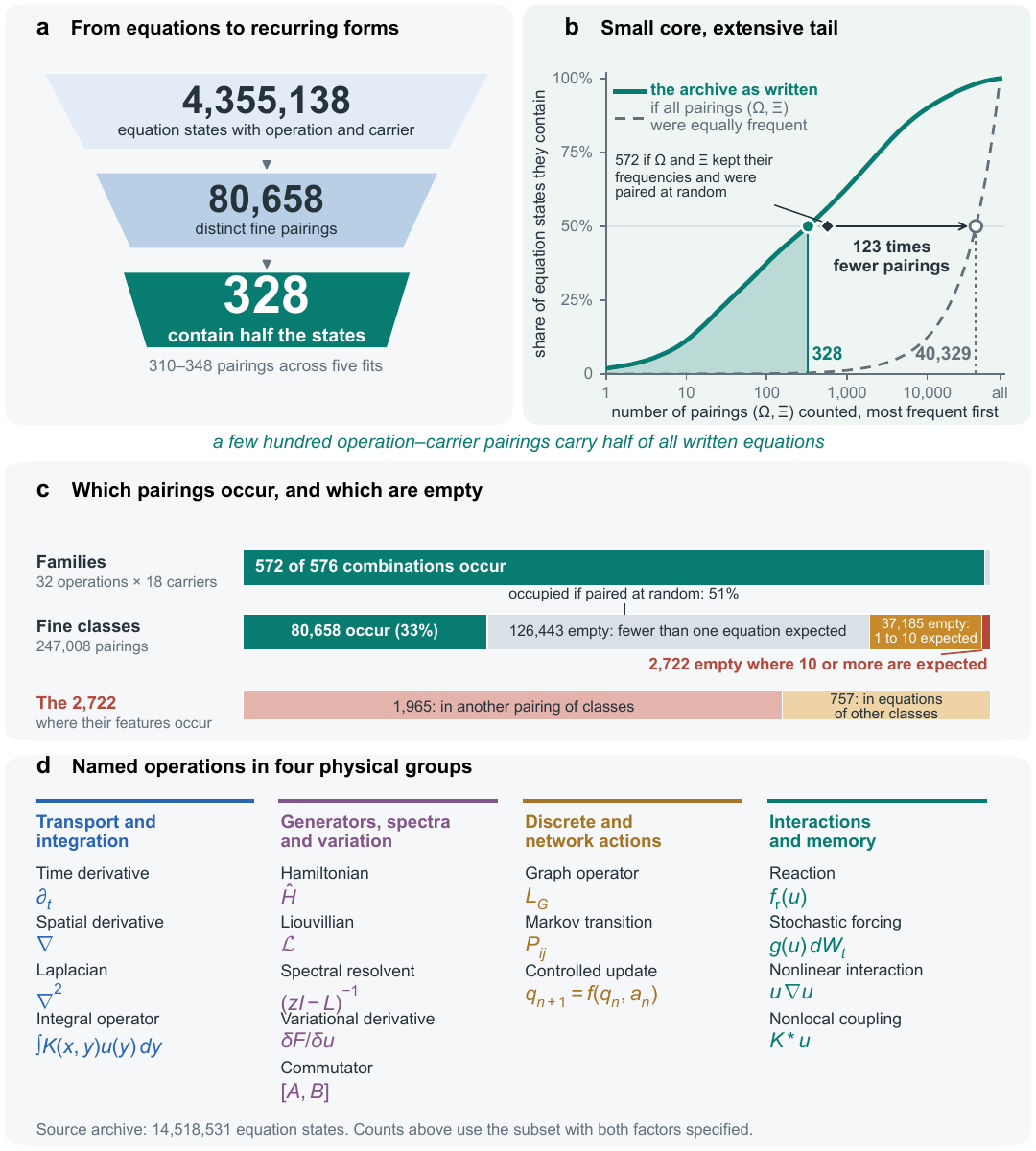}
\caption{\textbf{A few hundred pairings of an operation with a carrier contain half of the written equations.}
\textbf{a}, Of the \HyperionVTwentyOneMechanismRows{} equation states whose operation and carrier are both specified, \HyperionVTwentyOneMechanismOccupiedFine{} distinct fine pairings occur, and \HyperionVTwentyOneMechanismCellsForHalf{} of them contain half of the states (\HyperionVTwentyOneMechanismCellsForHalfMinimum{}--\HyperionVTwentyOneMechanismCellsForHalfMaximum{} across five fits). The source archive contains \HyperionCorpusRows{} equation states from 2,511,195 manuscripts.
\textbf{b}, The pairings are counted from the most frequent to the least, and the curve gives the share of the equation states that the counted pairings contain (solid, horizontal axis logarithmic). Half of the states are reached after \HyperionVTwentyOneMechanismCellsForHalf{} pairings, 0.41\% of those that occur. The dashed curve is the same count for an archive in which all \HyperionVTwentyOneMechanismOccupiedFine{} pairings were equally frequent: half of the states would then need half of the pairings, 40,329, which is 123 times more. The diamond marks the count if the classes of operation and of carrier kept their frequencies and were paired at random, 572. The bar at one half spans the five fits.
\textbf{c}, Pairings that occur and pairings that are empty. Among 32 families of operations and 18 of carriers, 572 of 576 combinations occur. Among the fine classes, 80,658 of 247,008 pairings occur; the tick marks the share that random pairing would fill. The empty pairings are divided by the number of equations that random pairing would place in them: fewer than one, one to ten, and ten or more (2,722). The lower bar shows where the combination of features of these 2,722 pairings is written: in another pairing of classes (1,965), or in single equations that the index assigns to other classes (757; for one of these the features are carried by separate equations of the same papers).
\textbf{d}, The operations recognized in equations, grouped by the physics they describe, with an example of each.}
\label{fig:funnel}
\end{figure}

These dependencies mark the joints at which a mechanism can be taken apart and rebuilt (Fig.~\ref{fig:atlas}a). At the centre is a core
\begin{equation}
M = (\Omega, \Xi)
\label{eq:bare-core}
\end{equation}
that pairs an operation $\Omega$ (a differential generator, a commutator, a variational principle or a stochastic jump operator) with the carrier $\Xi$ on which it acts (a field, non-negative amounts $\mathbb{R}_+^n$, a phase on the circle $S^1$, the vertices of a graph or a density operator). The core is completed by
\begin{equation}
I_{\mathrm{op}} = (M; C, R, P) = ((\Omega, \Xi); C, R, P),
\label{eq:op-completion}
\end{equation}
where $C$ contains the boundary conditions, couplings to baths and constitutive relations, $R$ the observable that connects states to measurements, and $P$ the preparation and the time-dependent drive. Binding the completed structure to a material $A$ gives the realized model:
\begin{equation}
I_{\mathrm{real}} = \underbrace{\Biggl(\; \overbrace{\underbrace{\Big(\overbrace{\Omega}^{\text{operation}}, \overbrace{\Xi}^{\text{carrier}}\Big)}_{\text{core } M} ;\; \underbrace{\overbrace{C}^{\text{closure}},\, \overbrace{R}^{\text{observable}},\, \overbrace{P}^{\text{protocol}}}_{\text{operational completion}}}^{I_{\mathrm{op}} \text{ (operational identity)}} ;\; \underbrace{A}_{\substack{\text{material} \\ \text{realization}}} \;\Biggr)}_{\text{complete realized physical model}}.
\label{eq:identity}
\end{equation}
The nesting states what can act on what: the material enters the evolution only through the closures, observables and protocols it declares, and these are defined on the carrier of the operation. Removing the material and then the completion projects a realized model onto its core,
\begin{equation}
I_{\mathrm{real}} \;\xrightarrow{\;Q_A\;}\; I_{\mathrm{op}} \;\xrightarrow{\;Q_{CRP}\;}\; M, \qquad Q = Q_{CRP} \circ Q_A,
\label{eq:compression-quotient}
\end{equation}
and models from distant fields fall onto the same core. In the written record this projection appears as a compression: of the \HyperionVTwentyOneMechanismRows{} equations whose operation and carrier are both explicit, half fall into \HyperionVTwentyOneMechanismCellsForHalf{} pairings of an operation with a carrier, 0.41\% of those that occur (Fig.~\ref{fig:funnel}a,b). The concentration comes mostly from a few frequent operations and a few frequent carriers: four classes of operation and 24 classes of carrier contain half of the equations, and with these frequencies a random pairing of operations with carriers would need 572 pairings for half of them, against the \HyperionVTwentyOneMechanismCellsForHalf{} observed. The projection forgets detail, so one core has many realizations. Construction runs the other way and must supply that detail: among all realizations of a core, it chooses one that a new material can carry.

\subsection*{Transfer of a mechanism between systems}

The simplest transfer carries an operation $\Omega$ onto a new carrier $\Xi$ and closes it with the relations $C$ of the target. The carrier is not a free choice: its geometry, boundaries and connectivity restrict which equations are admissible, and together with $C$ they fix what is conserved, so that an operation can be reattached only where these conditions are met. Detaching the Laplacian from temperature and attaching it to the probability density of a particle turns heat conduction into Brownian motion, and the conservation of heat into the conservation of probability. Attaching it to the vertices of a network with reciprocal currents $J_{ij}=Dw_{ij}(u_i-u_j)$ gives diffusion on a graph, which conserves the total amount and dissipates its energy exactly as the continuum does (Methods). Attaching drift and fluctuations to the parameters of a statistical model gives Langevin sampling, whose stationary state in continuous time and in a confining loss is the Gibbs density; the sampling algorithms of machine learning are diffusion carried into a space of parameters~\cite{Welling2011SGLD}.

Each of these transfers holds exactly, on the graph in the conservation law and the dissipation, and the condition for it can be written for any pair of models. A transfer is exact when evolving and then mapping gives the same result as mapping and then evolving. For a source evolving on $\Xi_s$ under $\dot{q}=F_s(t,q)$ with observable $R$, and a map $\alpha:\mathbb{R}\times\Xi_s\to\Xi_t$ onto a target evolving under $\dot y=F_t(t,y)$, the two routes differ by the compatibility defect
\begin{equation}
\Delta(t,q)=F_t(t,\alpha(t,q))-D_q\alpha(t,q)F_s(t,q)
-\partial_t\alpha(t,q).
\label{eq:nonlinear-residual-exact}
\end{equation}
The last two terms are the rate of change of the mapped state computed from the source, and the first is the rate that the target equation assigns to the same state. The transfer is exact when the defect vanishes, together with $R_t(\alpha(q))=R_s(q)$ for the measurement. Written for the operations themselves, with $\Omega_s$ and $\Omega_t$ acting on the source and target states and $\beta$ the map that $\alpha$ induces on their rates of change ($\beta=D_q\alpha$ for a deterministic flow; for a map that depends on time, $\beta\Omega_s$ includes $\partial_t\alpha$), the same defect is the remainder in
\begin{equation}
\Omega_t \alpha = \beta \Omega_s + \Delta_{\alpha,\beta}.
\label{eq:intertwining}
\end{equation}
For a stochastic flow, the second-order term of the chain rule appears in $\Delta_{\alpha,\beta}$. For a linear projection $\Pi$ onto fewer variables, with generators $L_a$ and $\overline{L}_a$ under protocol input $a$, the defect vanishes when evolution and projection commute~\cite{Shalizi2001,HoKalman1966,Petreczky2011}:
\begin{equation}
\Pi L_a = \overline{L}_a \Pi, \qquad R = \overline{R}\Pi.
\label{eq:intertwining-condition}
\end{equation}
The transfer of diffusion to a probability density satisfies this condition, and the transfer to a graph satisfies it to the accuracy of the grid (Methods). Most transfers do not, and the defect that remains is the quantity on which the constructor acts.

\subsection*{Definition of a mechanism and outcomes of a transfer}

A network of three compartments that exchange mass at rates $W_{ij}$ shows the cases that a transfer can meet. Only the deviation of the first compartment from equal occupation, $x=p_0-1/3$, is observed; the imbalance between the other two, $z=p_1-p_2$, is not, and the map from the network to its observed variable discards it. When the first compartment exchanges equally with the other two, the observed deviation relaxes on its own, $\dot x=-3x$: evolution and projection commute (Eq.~\eqref{eq:intertwining-condition}), the defect vanishes and $z$ can be forgotten. When the exchange is unequal, with $W_{01}=4$, $W_{02}=1$ and $W_{12}=0.7$, the two variables are coupled,
$\dot x=-a_x x+bz$ and $\dot z=cx-a_z z$ with $a_x=7.5$, $a_z=3.9$, $b=1.5$ and $c=4.5$, and the defect is $\Delta=-bz$: the observed rate depends on a variable that the description has dropped.

Two models share a mechanism when an invertible map between their states, whose derivative and that of its inverse stay bounded in time, carries one evolution into the other with zero defect, together with the measurement. A mechanism is therefore not an equation but a class of realizations related by such maps, and what one realization conserves the others conserve: a quantity $Q$ conserved by the source is conserved by the target as $Q\circ\alpha^{-1}$. A map with zero defect that discards states, as in the symmetric network, is a reduction and not an equivalence: the target is a smaller description of the source, and a conserved quantity of the target has to be supplied separately. When the source and the target are the same model, an invertible map with zero defect is a symmetry of that model: a symmetry is the transfer of a mechanism to itself, and a transfer extends the idea of symmetry to pairs of different systems. Two models can agree at four nested levels: their parts can correspond, their dimensionless groups can coincide, the map can carry one evolution into the other, and the observables can agree as well (Methods). Only the last level, which contains the third, makes them one mechanism; each lower level can hold without it.

A non-zero defect may only show that the two descriptions are inequivalent, and then nothing follows from it. It may instead close, that is, transform as a definite object under the allowed maps, obey a relation of its own and change a calculated or measured consequence. A closed defect is a physical quantity that the smaller description left out, such as a field strength, a boundary flux, a memory kernel, an interaction or a further degree of freedom. The defect of the unequal network closes: the imbalance $z$ obeys an equation of its own, driven by the observed deviation. A transfer by an invertible map that carries the measurement therefore has three outcomes,
\begin{equation}
\begin{aligned}
\Delta=0 &\;\Longrightarrow\; \text{the same mechanism in another physical form},\\
\Delta\neq0,\ \text{unstructured} &\;\Longrightarrow\; \text{inequivalent descriptions},\\
\Delta\neq0,\ \text{closed} &\;\Longrightarrow\; \text{additional physical structure}.
\end{aligned}
\label{eq:defect-outcomes}
\end{equation}
In the last case a difference between descriptions acquires its own law and consequence and becomes part of the target theory.

A missing term that depends only on the retained state of the target exists if and only if the source states that the map identifies require the same rate in the target; for linear dynamics with generator $L_s$ this condition reads $\ker\alpha\subseteq\ker(\alpha L_s)$. When it fails, no instantaneous term suffices, and the target must keep a hidden coordinate or the history of its state. The symmetric network satisfies the condition. The unequal network violates it, because states that differ only in $z$ require different rates of $x$, and the missing term becomes a memory kernel (Fig.~\ref{fig:obstruction}b), which is derived in the next section. Where the condition holds, the defect fixes the missing term only on the states that the map reaches; on the remaining states of the target the term is free. Restoring a discarded distinction is therefore not the inverse of discarding it. Many completions carry the same mechanism; they are different models, and the constructor keeps the one with the smallest additional term, field or coordinate, among those it searches, that represents the defect (Methods).

A construction usually chains several transfers, and exact local transfers need not compose into a consistent global one. Two sequences of transfers $p$ and $q$ between the same endpoints compose to maps $T_p$ and $T_q$, and their relative transport $\mathfrak H(p,q)=T_q^{-1}\circ T_p$ need not be the identity. In a constructed square, every edge transfers exactly, with a composite defect of $3\times10^{-14}$, yet the two routes between opposite corners differ by a map of norm 8.44 (Methods). When both routes are exact and the evolutions and the maps are linear, the difference of their maps itself carries the source evolution into the target evolution, and when the maps can be inverted the relative transport commutes with the evolution of the source: it is a symmetry of the source (Methods). Exact transfers can therefore disagree only by a symmetry of the mechanism, as two gauges of one field disagree only by a transformation that leaves the field equations unchanged. Whether the path matters to an experiment depends on what the transfers were required to carry. If both routes also carry the observable, $R_tT_p=R_s=R_tT_q$, then $R_s\circ\mathfrak H=R_s$, and that observable cannot tell the routes apart. The path is measurable when the routes are exact for the evolution alone, or when a second observable, which the transfers were not required to carry, distinguishes states that the symmetry relates.

If two preparations that a reduced description assigns to the same state respond differently to the same later drive, the description has omitted a coordinate that carries the path. Under curvature, transport around a closed loop returns the state rotated. Under frustration, the signs of the couplings around a loop multiply to $-1$, and no state satisfies every bond. Under memory, evolution over a time $t+s$ differs from evolution over $t$ followed by evolution over $s$, because the second step starts with the hidden coordinates reset. These are three physically different failures of composition, and the corresponding missing terms are a connection, a competing bond energy and a memory kernel (Methods). The comparison with a gauge field holds for the first of these, where every transfer has an inverse and the transport around a loop is defined. An elimination and a frustrated bond have no inverse, and for them the failure is stated by the composition of the reduced evolution or by the product of signs. A difference between paths that changes a response is therefore part of the mechanism.

\subsection*{Moves of the constructor and the equation for the missing term}

The constructor builds new models by changing one part of the nested object at a time, while the other parts keep their meaning, and each change, called a move, is a physical operation on a model. A mechanism is relocated when its core is \emph{detached} from its material and \emph{attached} to a new carrier or material, or when its operation is \emph{bound} to the geometry of a material. The description of the state changes when variables that are not observed are \emph{eliminated} and when the remaining ones are \emph{changed} by a nonlinear map. The protocol can be \emph{unfolded}, so that several parameters are driven together; a parameter can be \emph{promoted} to the control, and the range of the control can be \emph{extended}. The operation itself can be edited, by \emph{deleting} or adding a term and by \emph{composing} such edits. Separate objects are joined when a \emph{bridge} couples two carriers that the compiled record does not couple, when a \emph{clock}, a phase that runs at its own frequency, is coupled to an amount, and when a unit is \emph{replicated} around a ring. Two moves compare instead of changing: a material can be \emph{compared} with its \emph{counterpart}, the simplest system that obeys a structural rule, so that what the material adds beyond the rule becomes visible, and a mechanism can be \emph{co-discovered}, that is, reached by separate derivations in different fields. Table~\ref{tab:moves} lists the moves used in this work, with the defect that each produced and the term that closed it.
In every case the changed model is related to the original by a map of states, and the defect of Eq.~\eqref{eq:intertwining} states whether the mechanism survives the change. All moves rest on one physical rule:
\begin{quote}
\emph{A description can discard a distinction between states only while that distinction cannot affect any prediction under the allowed operations. Changing the interaction, the protocol or the measurement can make the discarded distinction active again.}
\end{quote}
The first half of the rule states what a description may discard, and it is the condition of zero defect for a reduction; the second states what must be restored, and it is the last case of Eq.~\eqref{eq:defect-outcomes}. Every move tests the rule by comparing two routes to the same result: a transfer compares evolving and then mapping with mapping and then evolving, an elimination compares reducing and then evolving with evolving and then reducing, and a protocol compares two orders of the same operations. The defect is in each case the difference between the two routes, and for this reason one equation serves all three.

When the defect closes, the target lacks a term, and adding a term $\mathcal{K}_t$ to the target operation restores the mechanism if
\begin{equation}
\mathcal{K}_t \alpha = -\Delta_{\alpha,\beta} \quad\Longrightarrow\quad (\Omega_t + \mathcal{K}_t)\alpha = \beta \Omega_s.
\label{eq:correction-constraint}
\end{equation}
This is an equation for the missing term, which may be a coupling, a memory kernel or a drive. A failed transfer whose defect closes therefore states what the target must contain for the mechanism to hold there.

\begin{figure}[!htbp]
\centering
\includegraphics[width=\textwidth]{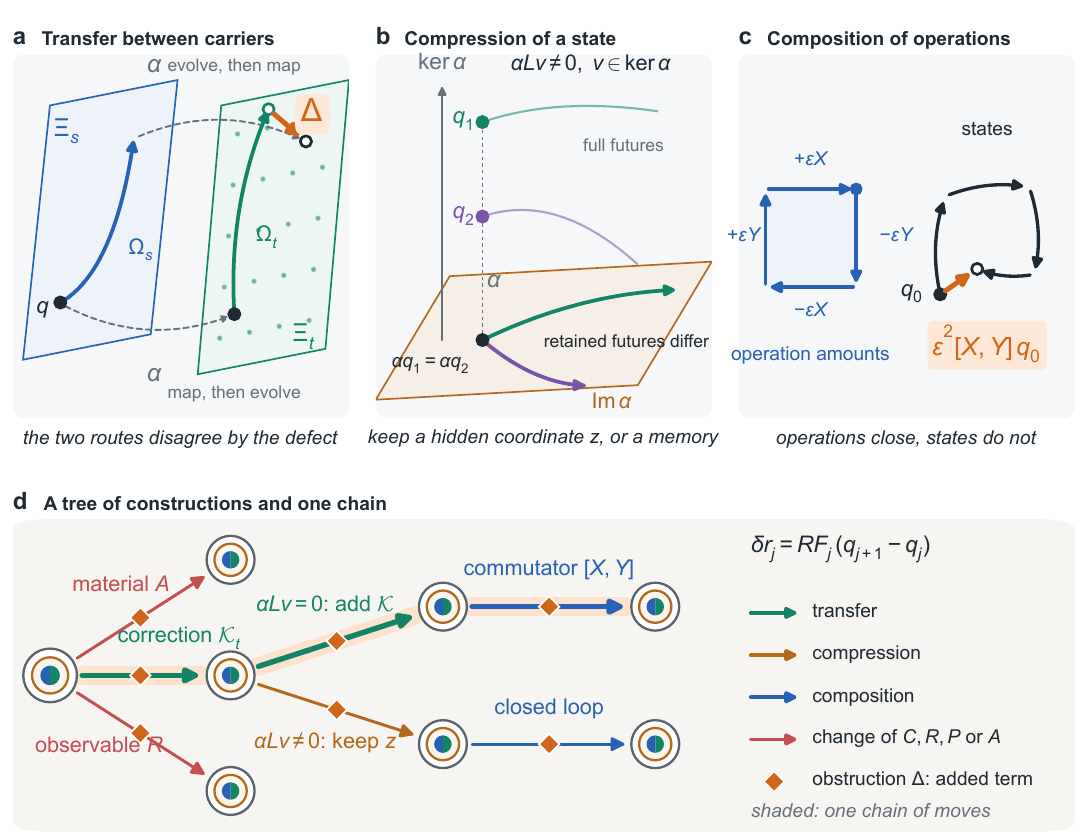}
\caption{\textbf{A failed transfer states which term is missing.}
\textbf{a}, A map $\alpha$ carries states of the source $\Xi_s$ to the target $\Xi_t$. Evolving a state $q$ with $\Omega_s$ and then mapping it differs from mapping it and then evolving it with $\Omega_t$ by the defect $\Delta$ (Eq.~\ref{eq:intertwining}).
\textbf{b}, Two source states $q_1$ and $q_2$ that the map identifies, $\alpha q_1=\alpha q_2$, differ by a vector $v$ in $\ker\alpha$. If they require different rates in the target, $\alpha L v\neq0$, no term that depends only on the target state can close the defect, and the target must keep the coordinate that tells them apart.
\textbf{c}, Shears $-\epsilon Y$, $-\epsilon X$, $+\epsilon Y$ and $+\epsilon X$, applied in this order, return the applied strain to zero, but the state does not return: it is displaced by $\epsilon^2[X,Y]q_0$.
\textbf{d}, A construction is a sequence of moves from a starting model: transfers, compressions, compositions and changes of the boundary relations $C$, the observable $R$, the protocol $P$ or the material $A$. At an obstruction (diamond) a term is added: an instantaneous term $\mathcal{K}$ when $\alpha L v=0$ for every $v$ in $\ker\alpha$, a new coordinate $z$ when $\alpha L v\neq0$ for some $v$. Along a chain of moves, the response changes by $\delta r_j=RF_j(q_{j+1}-q_j)$ at step $j$; the shaded chain is one route from a published model to a new one.}
\label{fig:obstruction}
\end{figure}

Three classical corrections of physics, each found in its own field by its own argument, follow from the defect of this one equation (Fig.~\ref{fig:obstruction}): two are missing terms, and the third is the sum of the defects along a closed cycle.

\paragraph{Memory from an eliminated variable.}
In the unequal exchange network of the previous section the defect is $\Delta=-bz$. States that differ only in $z$ are the case $\alpha Lv\neq0$ of Fig.~\ref{fig:obstruction}b, so no instantaneous term can close the defect. Solving for $z$ instead gives
\begin{equation}
\dot{x}(t) = -a_x x(t) + b\,e^{-a_z t}z(0) + \int_0^t bc\,e^{-a_z(t-s)}x(s)\,ds.
\label{eq:compartment-memory}
\end{equation}
Forgetting $z$ therefore has two consequences, both fixed by the defect together with the equation that $z$ obeys: a transient that depends on how the hidden compartments were prepared, and a memory kernel through which the past of $x$ returns by way of the hidden compartments. This is the structure that Mori and Zwanzig derived for any eliminated variable~\cite{Mori1965,Zwanzig1961}.

\paragraph{Drift from a change of variables.}
The modified Bessel process $dX_t=dB_t+[(\theta+1/2)/X_t-aX_t]\,dt$ cannot be rewritten in the variable $y=x^2$ by the ordinary chain rule~\cite{Hufnagel2020Bessel}. A Brownian increment satisfies $(dX_t)^2=dt$, so $y$ gains a drift of one unit per unit time that the chain rule does not produce, and the defect is this constant drift along $y$. Solving Eq.~\eqref{eq:correction-constraint} gives the missing term, the It\^o drift $\mathcal{K}_t=\partial_y$. With it, the rewritten equation has the generator of the original on every smooth observable; without it, the mean of $y$ falls short by $(1-e^{-2at})/(2a)$, which is 0.43 at $a=1$ and $t=1$.

\paragraph{Transport from a closed cycle.}
In a protocol of $m$ stages with linear maps, the defects of the stages add up along the chain, each carried forward by the maps of the later stages and fed by those of the earlier ones:
\begin{equation}
\Delta_{m\cdots 1} = \sum_{j=1}^{m} \left(\prod_{k>j} \beta_k\right) \Delta_j \left(\prod_{i<j} \alpha_i\right).
\label{eq:composed-defect}
\end{equation}
Around a closed cycle, with each stage compared with the identity map, these contributions need not cancel. Shearing an elastic medium vertically and horizontally by $-\epsilon$ and then undoing both shears, $U_\epsilon=e^{\epsilon X}e^{\epsilon Y}e^{-\epsilon X}e^{-\epsilon Y}$, brings the applied strain back to zero but not the state: $U_\epsilon q_0-q_0=\epsilon^2[X,Y]q_0+O(\epsilon^3)$. The bracket $[X,Y]$ stretches the medium along one axis and compresses it along the other, a deformation that neither shear produces on its own (Fig.~\ref{fig:obstruction}c; Methods). The displacement grows with the area enclosed by the cycle, not with its amplitude; the same geometry lets a falling cat turn over without angular momentum~\cite{Murray1993Nonholonomic}.

\begin{table}[!htbp]
\centering
\caption{\textbf{Moves of the constructor, the defects they produce and the missing terms obtained by solving $\mathcal{K}_t\alpha = -\Delta_{\alpha,\beta}$.} The rows order the findings of this work by the part of a reduction that failed. In the last two rows what is missing is an object and not a single term.}
\label{tab:moves}
\footnotesize
\renewcommand{\arraystretch}{1.22}
\begin{tabularx}{\textwidth}{>{\raggedright\arraybackslash\bfseries}p{2.3cm} >{\raggedright\arraybackslash}p{2.5cm} >{\raggedright\arraybackslash}p{3.9cm} >{\raggedright\arraybackslash}X}
\toprule
Move & Part changed & Defect $\Delta \neq 0$ & Missing term $\mathcal{K}_t$ \\
\midrule
Eliminate an unobserved variable & State space, by a projection $\Pi$ & Evolution and projection do not commute, $\Pi L\neq\overline{L}\Pi$: the observed variable exchanges with the unobserved ones in $\ker\Pi$ & A memory kernel $\int_0^t K(t-s)x(s)\,ds$ and a transient set by the unmeasured preparation (Eq.~\ref{eq:compartment-memory}; Fig.~\ref{fig:obstruction}b) \\
Change variables nonlinearly & State space of a stochastic equation, by a map $\alpha$ & The ordinary chain rule misses the second-order term $\frac{1}{2}\mathrm{Tr}(\sigma\sigma^{\mathsf T}D^2\alpha)$ & The It\^o drift $\mathcal{K}_t = \partial_y$ in the radial coordinate, or the shift of a washboard tilt to $A_e = A - \sigma^2/2$ \\
Unfold the protocol & Drive, $P_s \to P_t$ & A quadratic term $a_2 u^2$ that makes the two emerging states unequal, and the lag $-\partial_t\alpha = \dot{m}(t)$ of a moving reference state & The two-feed path $c^2 da + db = 0$ through the cusp~\cite{RemleinSeifert2024}, and the extra feed $+\dot{m}(t)$ that restores the $\pi^{1/4}$ law (Eqs.~\ref{eq:schlogl-path}--\ref{eq:schlogl-moving}; Fig.~\ref{fig:constructor}) \\
Compose two deletions & Operation, $\Omega \to (\mathcal{D}_1\circ\mathcal{D}_2)\Omega$ & Removing the rotation alone gives a shear; removing the growth alone keeps the cycle & Removing both leaves the pitchfork $\dot{y}=\omega(\mu y-y^3)$ on $x=0$ (Eqs.~\ref{eq:hopf-changes}--\ref{eq:hopf-open-pair}) \\
Bind to the geometry of a material & Material, $I_{\mathrm{op}} \to (I_{\mathrm{op}};A)$ & Rotating every orientation by the same angle, $\theta_i \mapsto \theta_i + \delta\theta$, costs nothing, so rotational diffusion erases a written orientation & A term tied to the bond angles, $-b_{ij}\cos 2(\theta_i+\theta_j-2\phi_{ij})$, which fixes the orientations up to a common rotation by $\pi/2$ and frustrates closed loops with incompatible bond angles~\cite{BaulinColloidMemory} \\
Bridge amounts and phases & Carrier, $\mathbb{R}_+ \times S^1$ & An amount requires every loss to vanish at the wall, $F_x|_{x=0} \ge 0$, while a phase requires $2\pi$-periodic terms; an additive phase forcing drives the amount negative & The multiplicative term $g x \cos\varphi$, with the frequency feedback $\dot{\varphi}=\omega-hx$ \\
Replace individuals by a density & Noise, in the closure $C$ & A noise of constant intensity keeps the drift and the transition and misstates the rate of rare switches: for the fall of a budworm population to extinction the action of the direct path is $0.085$ in place of $0.58$, and the exponent of the mean time to extinction, which passes through the outbreak state, is $0.86$ in place of $0.76$ & The rates $b$ and $d$ of the elementary events of birth and death, which give the lifetime $\exp(NS)$ with $S=\int\ln(d/b)\,\mathrm{d}x$~\cite{FieldBridge2026} \\
Reduce to the neighbourhood of a transition & State space, by a local reduction & Models with one normal form and one law of writing differ in what they keep: of nine with a supercritical pitchfork, three lose the written sign when the control is kept at its final value for as long as the write lasted, and a fourth when it is kept at its published value & What lies far from the transition: the attractor at the final value of the control, the lifetime of each state against the noise and, for a runaway, the stable manifold of an edge state (Methods) \\
\bottomrule
\end{tabularx}
\end{table}

In the nested language these three known corrections are one calculation, and the same calculation applies to models for which the compiled record contains no correction. It reaches them by moves that a derivation within one field rarely makes: two feeds driven together, two changes that each fail alone, two state spaces joined across their boundaries and an interaction bound to the geometry of a material (Table~\ref{tab:moves}). None of these moves requires a new technique: each is set aside by a habit of exploration, not excluded by the equations. For the constructor every part of the nested object is equally open to change, a move that fails is kept as the source of the next one, and the defect decides whether the changed model is admissible and which term it lacks. The workbench FieldBridge carries out these moves on given equations, computes the defect and solves for the missing term~\cite{FieldBridge2026}.

\subsection*{Construction of a writing protocol in a chemical reactor}

\begin{figure}[!htbp]
\centering
\includegraphics[width=\linewidth]{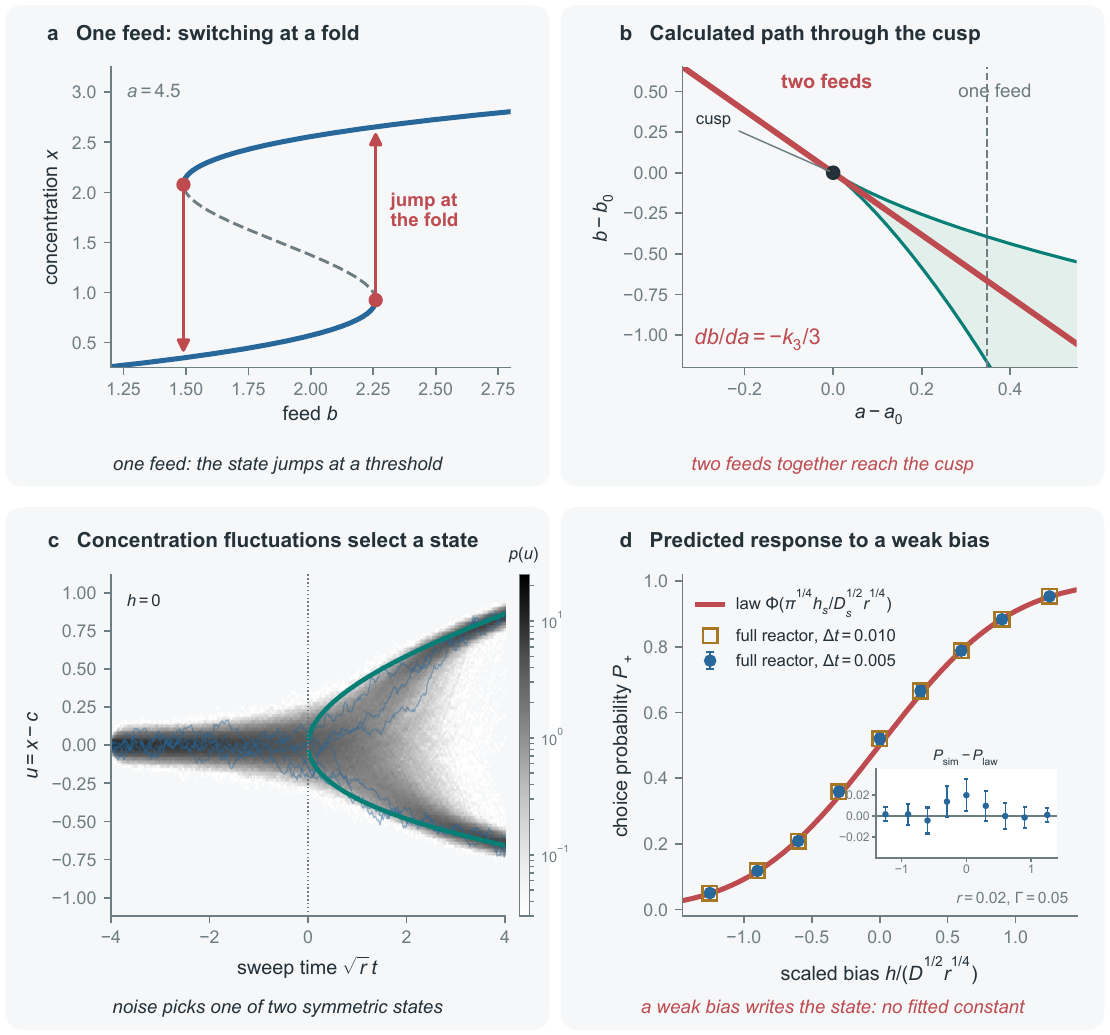}
\caption{\textbf{A two-feed protocol, constructed from the equations and tested in simulations of a chemical reactor.}
\textbf{a}, With one feed, the stable concentrations (solid) and the unstable one (dashed) form an S-shaped curve ($a=4.5$, $k_3=5.75$), and the reactor jumps between branches at the folds (arrows).
\textbf{b}, The two folds meet at the cusp. Keeping the critical concentration fixed while both feeds change gives the two-feed path (red), $\mathrm db/\mathrm da=-k_3/3$; the dashed line is the one-feed sweep of \textbf{a}.
\textbf{c}, Along this path at zero bias, the concentration leaves the single state and divides between the two new stable states (green): probability density and six trajectories of the full stochastic equation.
\textbf{d}, A weak bias in the second feed selects the upper state with the probability of Eq.~\eqref{eq:symmetric-write-law}, drawn with no fitted constant. Points are simulations of the full reactor at nine biases, 4,096 trajectories each, with error bars of 1.96 binomial standard errors; halving the time step from 0.010 to 0.005 changes no probability by more than 0.00024. The inset shows the departures from the law. The coefficient fitted to the points separately agrees with $\pi^{1/4}$ (Methods).}
\label{fig:constructor}
\end{figure}

A published chemical reactor switches only at a threshold; keeping its reactions and changing only how they are driven makes it a memory that a weak signal writes. The law that the written state then obeys was derived for symmetric switches~\cite{KondepudiNelson1985}; the construction supplies the protocol that brings the reactor under that law and, further below, the feed that lets the switch follow a moving reference. The Schl\"ogl reactor can remain at either of two stable concentrations under the same feed conditions. Its dimensionless concentration equation is
\begin{equation}
\dot x=-x^3+ax^2-k_3x+b,
\label{eq:schlogl-source}
\end{equation}
in which $a$ and $b$ are set by chemostatted feeds~\cite{Schlogl1972}. Varying $b$ alone carries the occupied state to a fold: that stable concentration disappears and the reactor jumps to the other branch (Fig.~\ref{fig:constructor}a). The aim here is a different response: the reactor should start near a single stable concentration and then develop two, of which a weak signal selects one, a response called here a symmetric write.

Starting from the one-feed model, the constructor follows its stable states and identifies the fold. In the model $a$ is a second adjustable feed, and promoting it to a control lets the constructor search for the cusp where the two folds meet (Fig.~\ref{fig:constructor}b). There the drift and its first two concentration derivatives vanish together, which determines a concentration $c$ and both feed values. Requiring $c$ to remain stationary during the subsequent sweep gives $c^2\,\mathrm da/\mathrm ds+\mathrm db/\mathrm ds=0$, the relation by which the two feeds must be changed together.

At the cusp, $a_0=\sqrt{3k_3}$, $c=\sqrt{k_3/3}$ and $b_0=c^3$. The joint feed path
\begin{equation}
a=a_0+s,\qquad b=b_0-\frac{k_3}{3}s
\label{eq:schlogl-path}
\end{equation}
keeps $x=c$ a steady state; this line through the cusp, with the pitchfork along it, was identified by Remlein and Seifert~\cite{RemleinSeifert2024}. Substituting $x=c+u$ gives exactly
\begin{equation}
\dot u=2cs\,u+s\,u^2-u^3.
\label{eq:schlogl-cusp}
\end{equation}
For $s<0$, disturbances from $x=c$ decay, whereas for $s>0$ they grow until the cubic term limits them on either side of $c$ (Fig.~\ref{fig:constructor}c). Near onset, $u\sim s^{1/2}$, so $su^2$ is smaller than the linear and cubic terms: the two emerging states are symmetric to leading order, and their wells become unequal only farther from onset. The reactions are unchanged; the construction changes only how they are driven.

The reduction succeeds here because a single direction loses stability, the cubic term limits the growth, and the joint path makes the transition two-sided. The obstructions that a reduction can meet fall into classes fixed by the linear problem at the transition. When a model is to have two states that a weak signal can select, the number of directions that lose stability together and their symmetry fix its reduced equation. The reduction fails, and its failure names what is missing, when a second direction loses stability, when the cubic term has the wrong sign, when a fold has no symmetry that makes it two-sided, or when the unstable pair oscillates. A symmetry can forbid such a switch outright: a ring of three identical units whose only symmetry is the rotation cannot reverse a real mode, because a rotation $g$ with $gv=-v$ would give $v=g^3v=-v$, so such a ring stores a phase instead of a sign~\cite{FieldBridge2026}.

A small offset in the second feed favours one of the emerging concentrations, and because the choice is made while the state is still within reach of the noise, the outcome is a probability. For bias $h_s$, additive diffusion coefficient $D_s$ and critical growth rate swept at rate $r$, linear amplification predicts
\begin{equation}
P_+=\Phi\!\left(\frac{\pi^{1/4}h_s}{D_s^{1/2}r^{1/4}}\right),
\label{eq:symmetric-write-law}
\end{equation}
where $\Phi$ is the standard-normal distribution function~\cite{KondepudiNelson1985}. A weaker noise or a slower sweep makes a given signal more effective at selecting the upper concentration. The law holds while the barrier between the emerging states outgrows the noise, that is, while $\Gamma=D_s|a_3|/r$ is of order one or smaller, with $a_3$ the cubic coefficient; at larger $\Gamma$ the state crosses between the wells after the bifurcation, and the coefficient falls below $\pi^{1/4}$, to about 1.12 at $\Gamma=10$ and 0.78 at $\Gamma=100$~\cite{FieldBridge2026}. Here the growth rate is $2cs$, so $r=2c\,\dot s$, and the simulated reactor has $\Gamma=0.05$. Stochastic simulations of the full reactor at nine biases follow this law (Fig.~\ref{fig:constructor}d; Methods).

The law contains no property of the reactor other than $h_s$, $D_s$ and $r$, and the reason lies in the scales of the passage. During the passage the state is of size $D_s^{1/2}r^{-1/4}$, and the passage lasts a time $r^{-1/2}$. Measured in these units, the bias appears only as $h_s/(D_s^{1/2}r^{1/4})$ and the cubic term only as $\Gamma$, while every higher power of the state and every mode that relaxes faster than $\sqrt r$ drops out as the noise and the sweep rate decrease (Methods). A reactor, a superconducting circuit and any other system whose passage reduces to one critical coordinate with a noise of constant intensity therefore write by the same law with the same constant, and, for a sweep that starts and ends far from the transition on the scale $r^{-1/2}$, a measured constant that differs from $\pi^{1/4}$ shows that a slow mode or a large $\Gamma$ has been left out. The construction turns a writing law into two feed trajectories that an experiment can impose; along the original one-feed path the same reactor switches only at a threshold.

The same reactor shows how a failed move names its missing term. Suppose that the critical concentration is to follow a prescribed trajectory $m(t)$ instead of staying at $c$. With $\alpha(t,x)=x-m(t)$ in Eq.~\eqref{eq:nonlinear-residual-exact}, the feeds $a(t)=3m(t)$ and $b(t)=k_3m(t)-2m(t)^3$ remove the quadratic term but leave a defect set by $\dot m(t)$, the rate at which the reference concentration moves. Adding $\dot m(t)$ to the second feed gives exactly
\begin{equation}
\dot u=[3m(t)^2-k_3]\,u-u^3,\qquad u=x-m(t).
\label{eq:schlogl-moving}
\end{equation}
Physically, the feed must also supply, at each instant, the change of concentration that the moving reference requires; a feed is non-negative, so the reference can be followed while $k_3m-2m^3+\dot m\ge0$, as it is along the trajectory used here. With this term, integrations of the full reactor follow Eq.~\eqref{eq:schlogl-moving}; without it they depart by up to 0.63 (Methods).

\subsection*{Changes of published models}

Joining the two feeds along one path changed the reactor, and the changed model was kept: its reduction showed that the quadratic term $su^2$, which separates it from a symmetric switch, is small near onset. Requiring the state to follow a moving reference prescribed what the reactor should do, and the defect gave the feed that it lacks. The first use of a move edits a mechanism and the second completes one, and both apply to the published models that the record supplies.

The habits that leave mechanisms unwritten are habits of exploration: a model is followed along the control and over the range that its authors chose, a change is kept only when it works on its own, and each discipline keeps to its own state space. Each habit corresponds to a move of the constructor: extending the range or promoting another parameter to the control lifts the first, composing changes that fail alone lifts the second, and a bridge between carriers lifts the third. The reactor has already shown one of them, a feed that its protocol keeps fixed promoted to a control. Here the constructor applied three changes to 76 models extracted from papers in 17 arXiv archives, removing one term, making another parameter the control, or extending the range of the control, without a target chosen in advance, and asked whether the model has two stable states of which a weak signal can select one, that is, a switch that can be written (Methods). These changes edit a mechanism: no correspondence with the source model is prescribed, and cancelling the defect of a deleted term under the identity map would only return the term. Pairs of such changes reach what single changes do not, and joining two carriers leaves the state space of one discipline; both follow in the next section.

\paragraph{Beyond the authors' control.} Lorenz's model shows what extending a range does. Most studies of it follow the chaos far above the onset of convection, and one record runs the Rayleigh parameter from 24.74 to 60 only. Extending this range downwards returns the pitchfork at which convection begins, and with it the two stable senses of rotation~\cite{Lorenz1963}. The pitchfork is in the equations and outside the range that the paper studied: a range belongs to a study and not to its equations. An extension recovers in the same way the fold of the R\"ossler flow~\cite{Rossler1976}, studied for its chaos over a narrow range of one parameter, at which its two equilibria appear, one stable and one a saddle; this is a transition, but not a choice between two states (Methods; Supplementary Section~S29). A range is extended only as far as the parameter keeps its meaning: a rate, a saturation constant or a mass is not continued through zero (Methods).

\paragraph{From recovered transitions to predicted ones.} Both transitions are in textbooks, and for that reason they test the procedure: given a protocol that stops short of a transition, the constructor returned the transition and its kind. The calculation does not depend on whether the answer is known. Every compiled model declares the range that its authors explored, and most declare parameters that their protocol never varies; each such range and each fixed parameter is a question about the equations, which a continuation answers. Where the source reports a single regime and the control is a quantity that an experiment can set, a transition returned outside the published range is a prediction for the system itself, for as long as the equations and their coefficients hold there: a value of the control, the kind of transition and, where two states appear, the law by which a weak signal chooses between them (Eq.~\eqref{eq:symmetric-write-law}). Whether the papers that use the model already know the transition can be read from the same record, because a transition leaves a trace in equations, a vanishing eigenvalue or a second branch of solutions. The reactor is the worked case: its fold was known, its cusp lay along a feed that the one-feed protocol keeps fixed, and promoting that feed to a control gave a memory whose law the simulations confirm.

\subsection*{Changes of the operation, the carrier and the material}

\begin{figure}[!htbp]
\centering
\includegraphics[width=\linewidth]{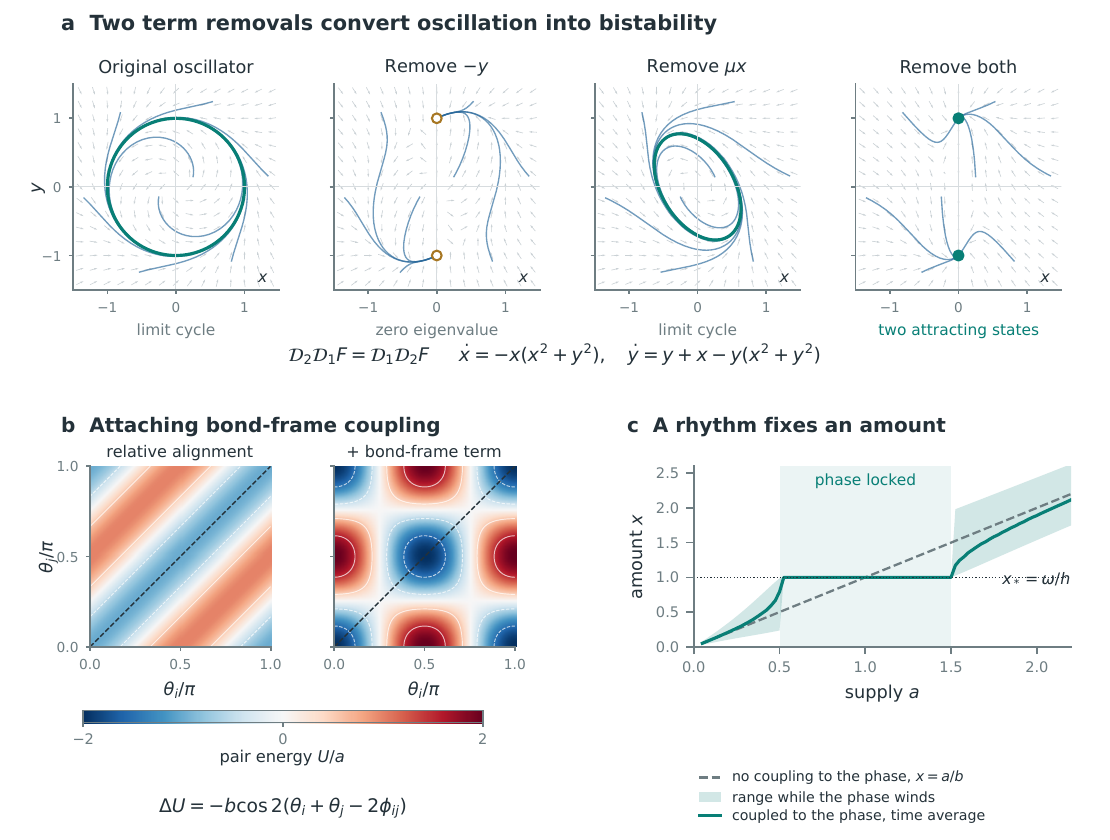}
\caption{\textbf{Two changes of an oscillator, an interaction bound to a material, and an amount fixed by a rhythm.}
\textbf{a}, Phase portraits of the Hopf normal form at $\mu=\omega=1$: the original model, either single deletion, and both deletions. Blue curves are integrated trajectories; green curves are limit cycles, and filled green points are the two attracting states after both deletions. Open points mark the stationary states of the model without rotation, which have a zero eigenvalue.
\textbf{b}, Pair energies before and after the bond term is attached, with $a=1$, $\phi_{ij}=0$ and $b=0$ or 1. The dashed diagonal is a common rotation of both particles: its energy is constant before and depends on the angle after.
\textbf{c}, Amount of Eq.~\eqref{eq:amount-phase} against the supply $a$, integrated with $b=1$, $g=0.5$ and $\omega=h=1$. Points are time averages and the band is the range of the oscillation. The dashed line is the amount $a/b$ without the coupling and the dotted line is the locked value $x_*=\omega/h$; the shaded interval is the window $|a-b\omega/h|<g$ in which the phase locks. Outside it the phase winds and the amount oscillates about a mean that follows the supply.}
\label{fig:colloid-case}
\end{figure}

\paragraph{Two changes of an operation.} The Hopf normal form
\begin{equation}
\dot x=\omega(\mu x-y-x\rho^2),\qquad \dot y=\omega(\mu y+x-y\rho^2),\qquad \rho^2=x^2+y^2,
\label{eq:hopf-changes}
\end{equation}
oscillates for $\mu>0$. Removing the rotation $-\omega y$ alone turns the rotation into a shear, and both directions lose stability together; the trajectories then approach $y=\sqrt{\mu}$ or $y=-\sqrt{\mu}$ according to the sign of $x$, two states with a zero eigenvalue that attract from one side only. Removing the growth $\omega\mu x$ alone leaves the oscillation in place for $\mu<2$, which contains the published range of the control. A derivation that keeps a change only when it has an effect on its own discards both. Removing both gives
\begin{equation}
\dot x=-\omega x\rho^2,\qquad \dot y=\omega(\mu y+x-y\rho^2),
\label{eq:hopf-open-pair}
\end{equation}
in which $y$ is the only direction that becomes unstable, the direction $x$ staying neutral at the origin: on $x=0$, $\dot y=\omega(\mu y-y^3)$, where the cubic term limits $y$ on both sides and the oscillator becomes a symmetric switch between $y=\pm\sqrt{\mu}$ (Fig.~\ref{fig:colloid-case}a). The two deletions commute, so the switch is their joint effect and does not depend on their order. The constructor forms every such pair and judges it afterwards: on the same 76 published models, 116 of 2{,}087 pairs reach an outcome that no single change reaches, and the seven that construct a switch are all in the Hopf normal form: four pairs of removals and three in which a removal follows an extension of the range (Methods).

\begin{figure}[!htbp]
\centering
\begin{minipage}[c]{0.565\textwidth}\includegraphics[width=\linewidth,trim=0 0 225.4 0,clip]{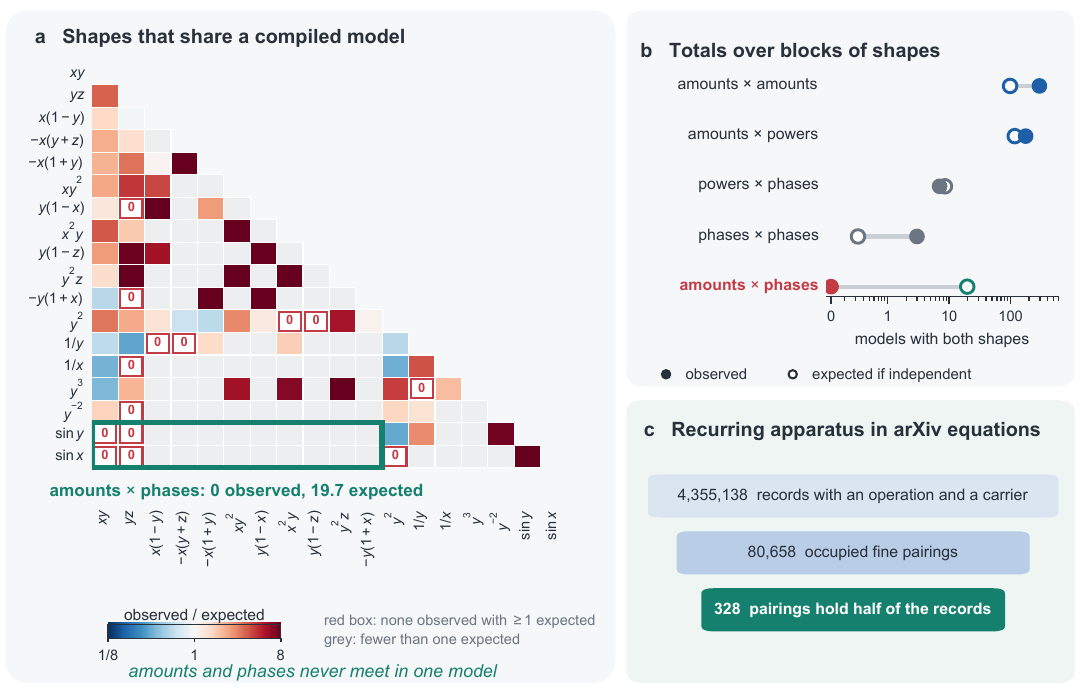}\end{minipage}\hfill
\begin{minipage}[c]{0.42\textwidth}\includegraphics[width=\linewidth,trim=300 142.2 0 0,clip]{figures/hyperion_nmi_written_record.pdf}\end{minipage}
\caption{\textbf{Amounts and phases never appear together in a compiled model.}
\textbf{a}, Each cell compares how often two interaction shapes occur in the same model with the count expected if shapes were combined independently, over 2,745 closed dynamical models compiled from 1,899 manuscripts in 21 fields. Red boxes mark pairs never observed although at least one was expected; grey cells expect fewer than one. The block that couples amounts to phases (green frame) is empty: no model, against 19.7 co-occurrences expected over its 22 pairs of shapes.
\textbf{b}, Observed (filled) and expected (open) counts summed over blocks of shapes. Only the block of amounts with phases is empty.}
\label{fig:record}
\end{figure}

\paragraph{A bridge between two carriers.} The disciplines that evolve amounts and those that evolve phases rarely write the two together. Of 2,745 closed models compiled from 1,899 papers in 21 fields, none, as written, couples the mass-action kinetics of amounts to periodic functions of a phase, against 19.7 co-occurrences of the two kinds of term expected if the terms combined at their observed frequencies (Fig.~\ref{fig:record}a,b). The boundaries of the two carriers state which couplings are admissible. Amounts live on the non-negative orthant, whose wall $x_i=0$ requires every loss to vanish as the amount vanishes. Phases live on the circle, which requires every term to be periodic. A forcing $+g\sin\varphi$ that stands alone in a concentration equation points out of the wall during half of every cycle. A periodic term is admissible when it multiplies the amount or stays non-negative at the wall, and the simplest generator that the two boundaries admit is
\begin{equation}
\dot{x} = x(a - bx + g\cos\varphi), \qquad \dot{\varphi} = \omega - hx,
\label{eq:amount-phase}
\end{equation}
in which the phase gates the growth of the amount and the amount slows the phase. When the phase locks, $\dot\varphi=0$ fixes the amount at $x_*=\omega/h$, a value that contains neither the supply $a$ nor the saturation $b$. A change of supply shifts the phase at which the lock holds, $\cos\varphi_*=(b\omega/h-a)/g$, and leaves the amount where the rhythm set it, for as long as $|a-b\omega/h|<g$ and the phase has been captured by the lock. The rhythm, and not the supply, therefore sets the amount (Fig.~\ref{fig:colloid-case}c). Known circuits that maintain an amount exactly at a set point keep the integral of its error in a further controller species~\cite{Briat2016,Aoki2019Integral}; here the phase, $\varphi=\int(\omega-hx)\,\mathrm dt$, is that integral, and no further species is needed. The lock is that of a phase-locked loop, which keeps an oscillator at a reference frequency through the integral of the frequency error~\cite{Gardner2005}; here an amount sets the frequency, and the wall of the amounts restricts the form of the coupling. Couplings of an amount to a phase are known outside the compiled set, in the bosonic Josephson junction and in polariton condensates fed by a reservoir~\cite{Smerzi1997,Wouters2007}; here the coupling is the simplest that the boundaries of two state spaces that the compiled record keeps apart admit, with a frequency that decreases linearly with the amount. Two further couplings of the family are described in Methods.

\paragraph{An operation bound to a material.} Rod-like particles whose interaction depends only on their relative orientation, $-a_{ij}\cos 2(\theta_i-\theta_j)$, cannot keep a written pattern: rotating every particle by the same angle costs nothing, and rotational diffusion erases the pattern along that free direction. Binding the operation to the fixed bond angles $\phi_{ij}$ of particles trapped at an interface derives the missing term,
\begin{equation}
U_{ij} = -a_{ij}\cos 2(\theta_i - \theta_j) - b_{ij}\cos 2(\theta_i + \theta_j - 2\phi_{ij}),
\label{eq:constructed-colloid}
\end{equation}
which quadrupolar capillary deformations of the interface supply. It pins the free rotation and frustrates those loops of the network around which the bond angles are incompatible, as they are on a disordered network (Fig.~\ref{fig:colloid-case}b). In Brownian dynamics of 1,024 particles, a pattern written by a transient field kept a connected overlap of $0.502\pm0.033$ at $D_rt=625$ without macroscopic nematic order; removing the bond term, or replacing it by relative alignment of equal weight, left mean overlaps between $-0.004$ and $0.002$, with standard errors of at most 0.005 (Methods)~\cite{BaulinColloidMemory}.

\paragraph{Against a structural rule.} A network of bistable elements whose couplings and drive can all be satisfied together returns to its previous state whenever the drive returns to a previous turning point. Comparing a material with such a network separates what the material does from what its structure guarantees. Reversing a smallest set of signs that removes the competition from a network of ten elements with competing couplings gives a counterpart, with the same graph, coupling strengths and thresholds; the same elements without couplings reach 31 states. Eighteen sets of eight reversals are smallest, and their counterparts reach 4, 5 or 7 states and return in every excursion; the counterpart used here, which reverses six couplings and the drive on two elements, reaches 5; the network reaches 20 states and fails to return in 65 of 425 excursions~\cite{FieldBridge2026}. The competition between couplings therefore adds between 1.5 and 2.3 bits of states, 2.0 against the counterpart used here, that some history of the drive can write, and with them the network loses the guarantee of return.

\paragraph{Writing and retention.} Nine published models share the supercritical pitchfork, and with it the law of writing of Eq.~\eqref{eq:symmetric-write-law}. Each was swept through the transition with its own noise and the control was then kept at its final value for as long as the sweep had lasted. Six kept the sign they had chosen and three lost it: Lorenz's model to a motion that circulates around both states, the others to their noise. A fourth, a parametric oscillator, kept the sign at the final value and lost it to its noise at its published control, just beyond the transition. A normal form is a reduction to the neighbourhood of a transition: it fixes how a state is written and leaves open how long the state is kept. What decides retention lies far from the transition, and a mechanism must carry it in addition to its normal form. A density with a noise of constant intensity is a reduction of the same kind, of a population of individuals: it holds for what the drift decides and fails for rare switches, whose rate is set by the birth and death rates along the path of escape. In both cases, the last two rows of Table~\ref{tab:moves}, what the reduction omits is an object and not a single term. The record of a mechanism therefore requires one further entry, the lifetime of each state against the noise of the model, which can be calculated from the equations (Methods).

\subsection*{The constructions and the written record}

The constructions start from models that the record contains and add terms that it does not contain. The index of the archive counts equations by the notation of their parts (Fig.~\ref{fig:funnel}), and its empty fine pairings are boundaries between classes of notation rather than combinations that no one has written: the features that characterize the two classes of an empty pairing are written together elsewhere in the index. Notation does not carry over to dynamics either: a model compiled from a paper whose notation states a conservation condition is no more likely than any other to conserve a linear quantity (Methods). What the record lacks appears instead in the terms of closed models, which can be run, and there amounts and phases are never combined although their carriers admit a coupling (Fig.~\ref{fig:record}).

On closed models, recombining the record rarely leaves it. Of the 976 models that stay bounded among 2{,}999 assembled at random from the terms that papers write and swept along one control, 890 follow a sequence of equilibria and attractors that some written model follows, and the other 86 fall in 76 of the 119 sequences that the assembled models show, none of which contains more than four models. Near a transition one motion of a system becomes slow and all the others follow it, so that systems which differ in every detail obey the same few reduced equations~\cite{GuckenheimerHolmes1983}, and this accounts for much of the recurrence. For the coupling of Eq.~\eqref{eq:amount-phase} the sequence can be derived in any network of amounts $\dot x_j=f_j(x)$: a steady state requires $\dot\varphi=0$, so the coupled species stands at $\omega/h$ whatever the kinetics, steady states exist while $|f_i|<g\omega/h$ at that composition, and, where the other species have an isolated steady state at that composition and the control crosses the edge of this window, they appear in pairs at a fold, as in any driven oscillator (Methods). A sequence of equilibria counts states and their stability, whereas what the lock fixes, the amount at $\omega/h$, is a property of the constructed term and of the window in which the phase locks.

\section*{Discussion}

Fick's step from heat to salt and Einstein's from salt to a Brownian particle were made by people who carried a mechanism into a new system and noticed where it did not fit. The constructions of this work make both steps a calculation: carrying is a change of one part of the nested object, and noticing is the defect of the map between the old model and the new. When a correspondence is prescribed, as for the reactor that follows a moving reference, the defect is an equation for the term that the new model lacks, and the term comes with the condition under which it is required. When a model is edited, as for the oscillator that lost two terms or the two carriers joined across their boundaries, the changed model is kept, and its reduction gives a law that the full dynamics confirms or refutes.

The written record is organized in the way that physics organizes its own descriptions. A mechanism is the class of realizations related by maps with zero defect, and the record contains it only through its realizations, as a geometry is known through its charts. Exact transfers between two models differ only by a symmetry of the mechanism, and a difference between routes that changes a response belongs to the mechanism, as the holonomy of a connection belongs to the field. The distribution of the record has a reason of the same kind. Because the reduced equations of a transition do not depend on the details of a system, transitions are shared by fields that share nothing else. What separates fields is the space on which the state lives, a half-line that ends at a wall, a circle or the sites of a graph, and each space admits its own terms; relative to shuffled papers, derivations keep their state space more firmly than their operation. The empty block of amounts with phases is a place where the admissible terms of two spaces had not been worked out together, and not a coupling that physics forbids. Construction is therefore most productive where two state spaces meet, and Table~\ref{tab:carriers} lists such pairs.

The four constructions act in one place: on what a reduction to a local form sets aside. Each restored one such part and obtained a response that the reduced model could not show: a memory written by a weak signal in a reactor that otherwise switches at a threshold, a switch in an oscillator, an amount set by a rhythm, a pattern that a simulated monolayer retains over the simulated time. A reduced law fixes how a state is written; how long the state is kept is decided far from the transition, by its lifetime against the noise, as the models that share one normal form showed.

Where a search by hand follows the published control, keeps a change only when it works alone and stays within one state space, the constructor applies every admissible move, keeps a failed intermediate as the source of the next move and turns each failure whose defect closes into an equation for the missing term. Every changed model becomes a new source, and every term that a failure supplies enlarges the set of operations that later moves can carry. Replicating a unit that has no reflection symmetry of its own around a ring of three, for example, yields a model in which a symmetry forbids the reversal of a real mode and leaves a phase to be stored; such a model is not a variation that a modeller of one field would try, and the same defect decides it. A search of this kind is set by what the equations allow and what the reductions decide, rather than by a target chosen in advance.

The room for such construction is large. The compiled record contains thousands of closed models, each with several parts that can be changed, and the state spaces of Table~\ref{tab:carriers} can be joined in pairs, each of which admits its own set of couplings. A single filled pair gave a regulator that the compiled models of neither field contain, and a single unfolded protocol gave a memory in a reactor known for fifty years. If the other pairs and protocols yield terms at a comparable rate, the mechanisms within reach of calculation outnumber those that have been written, in chemistry, in the design of materials and devices and in biology, where a mechanism is usually sought by experiment alone. The work of the investigator then moves from deriving equations within one discipline to stating the function that a system should perform and the material that should carry it.

On published models the constructor has so far returned known transitions: the onset of convection in Lorenz's model, the fold of the R\"ossler flow and a switch in the Hopf normal form. They calibrate a procedure that runs unchanged on any closed model, and the step from calibration to prediction is one of selection, towards controls that an experiment can set and sources that report a single regime. Chains of three or more changes will show whether construction keeps reaching new outcomes or saturates, as random recombination of the record does. The defect equation is applied here to closed ordinary differential equations swept along one control; extended to fields, delays and populations of individuals, and joined with the lifetimes that decide retention, it reaches correspondingly further.

A term obtained from a defect is calculated from a failure and not selected from the record; in this a constructor differs from a system that produces likely continuations of what has been written~\cite{Bender2021Parrots}. Whether two realizations are views of one core, the question raised for models trained on different data~\cite{Huh2024Platonic,Balestriero2026Platonic}, is calculated as well: they are when their defect vanishes, and otherwise the defect states what differs. What the constructor makes calculable is the term that a new system lacks, the condition under which the term is required and the law of the changed model. The two-feed reactor, the clock that regulates an amount and the interfacial monolayer are each specified closely enough to be built.

\section*{Methods}
\small

\paragraph{Reading equations from the arXiv sources.}
The \LaTeX{} sources of 2,511,195 manuscripts were split into equation blocks, and each symbol was resolved against its definition in the surrounding text. This gives \HyperionCorpusRows{} equation states, each linked to the passage it came from (Supplementary Sections~S1--S4 and S14). The accuracy of this step was checked rule by rule on source examples and has not been measured on a sample annotated by hand (Supplementary Section~S14). Every state was described in the nested form $(((\Omega,\Xi);C,R,P);A)$ by features of mathematical notation: 44 for the operation, 28 for the carrier and 23 for the other parts, 95 in all. The features record the structure of the operation, whether differential, integral, algebraic, variational or stochastic ($\Omega$: \HyperionVTwentyOneOmegaNonNullStates{} non-empty coarse classes and \HyperionOmegaLeaves{} fine classes); the carrier of the states, its topology, dimension and whether it is positive or periodic ($\Xi$: \HyperionVTwentyOneXiNonNullStates{} non-empty coarse classes and \HyperionXiLeaves{} fine classes); and the boundary and bath relations $C$, the observable $R$ and the protocol $P$. The retained equations were joined by typed edges, namely nearest neighbours in operation and in carrier, coupled pairs, candidate transfers and the order of equations within a document, and an edge was given a direction wherever its destination states more roles than its origin, which gives \HyperionVTwentyOneDAGDirectedEdges{} directed links out of \HyperionVTwentyOneDAGInputEdges{} typed edges (Supplementary Sections~S6 and~S7).

\paragraph{What the text index records.}
Because the features are cues of notation, the index groups equations by how they are written. Among the 14{,}518{,}531 rows of the index, the 44 operation features take 69{,}352 distinct values; 40\% of rows carry no operation cue, three values cover half of all rows and 170 cover 90\%. A row with an operation cue shares its exact operation vector with a median of 128{,}665 other rows, so its nearest neighbours in the index come from large groups of identical vectors and are not physically related equations: the time-derivative cue, for instance, is shared by an accretion rate $\dot m$ and the voltage derivative of a neuron model. The 28 carrier features take 110{,}164 values, 30\% of rows carry none, and a row with a carrier cue shares it with a median of 111{,}190 others. The leading principal directions of the operation features are the presence of $\lambda$, which switches on the spectral and eigenvalue features together, and of the string ``$=0$''; those of the carrier features are the presence of $\phi$ or $u(\cdot)$, of $\in$ and of $\epsilon$ or $\ll$. Several operation and carrier features are triggered by the same symbol (angle brackets for an expectation value and for a Hilbert space, $\nabla$ for a spatial derivative and for a coordinate domain, ``stoich'' for a stoichiometric closure and a stoichiometric space), so their co-occurrence is set by the rules. The index serves to retrieve equations and to count notation, and mechanisms are compared on closed models.

\paragraph{Closed models and their interaction shapes.}
Self-contained systems of ordinary differential equations were turned into executable symbolic models from the passages that state them: 2,745 closed models from 1,899 papers in 21 fields (Fig.~\ref{fig:record}). Each model records its state variables, the space they live in ($\mathbb{R}^n$, $\mathbb{R}_+^n$, $(S^1)^n$ or a simplex), its drift $F(q;\mathbf{p})$ and the parameter values or ranges its source declares. After the variables of each equation were renamed in a standard way, each equation calling its own state variable $x$ (Supplementary Section~S30), every drift was split into interaction shapes, such as mass-action products or trigonometric functions of a phase. The count expected for two shapes in one model is the product of their frequencies over the 2,745 models; for amounts and phases the expected counts of the 22 pairs of shapes sum to 19.7 co-occurrences, against none observed (Fig.~\ref{fig:record}b). Exact hypergeometric tests and permutations of the archive confirm the expected counts.

\paragraph{Concentration of the written equations.}
The index of the equations, which covers physics, mathematics, quantitative biology, computer science and nonlinear science (Supplementary Sections~S1--S4 and S15), groups equations by how they are written, and it shows how the record is distributed. Of the \HyperionVTwentyOneMechanismRows{} equations whose operation and carrier are both explicit, half fall into \HyperionVTwentyOneMechanismCellsForHalf{} pairings of an operation with a carrier, which are 0.41\% of the \HyperionVTwentyOneMechanismOccupiedFine{} pairings that occur at all (Fig.~\ref{fig:funnel}). At the coarse level nearly every family of operations meets nearly every family of carriers, but at the fine level \HyperionVTwentyOneMechanismFineEmpty{} of the possible pairings are empty. The expectation under random pairing is $E=Np(\Omega)p(\Xi)$ for the 248 classes of operation and 996 classes of carrier that occur among the $N$ equations, and a pairing is then occupied with probability $1-e^{-E}$. This gives 126,006 occupied pairings against 80,658 observed, 572 pairings for half of the equations against \HyperionVTwentyOneMechanismCellsForHalf{}, and no empty pairing with $E\ge10$ against 2,722. The features of a class are those that at least 60\% of its equations carry. A combination counts as written in single equations when an equation of the index carries every feature of both classes; closed models are those compiled from the papers of such equations. Among families, all but four of the 576 combinations of an operation with a carrier occur; among fine classes, a third of the 247,008 pairings occur where random pairing would fill a half. The 2,722 pairings that are empty although $E\ge10$ are spread over 184 classes of operation and 546 classes of carrier. For 1,965 of them the combination of features of the two classes occurs in another pairing of classes; for the other 757 it occurs in single equations that the index assigns to other classes, in 3 to 2,590 equations for each combination, except one combination whose features are carried by separate equations of the same papers. Of the 1,232 pairs of a feature of the operation with a feature of the carrier, 23 never occur in one equation, and only two of these where ten or more equations are expected. Of the 19,857 papers in which an equation carries one of these combinations, 379 gave a closed model, and among the models from papers whose combination states a conservation condition, 13 of 82 conserve a linear quantity, where one in five of all compiled models does. These are counts of notation: a few ways of writing an operation on a carrier are reused throughout the archive, and most combinations of notations do not occur. Whether equations that are written alike share a mechanism is a question about their dynamics, and it is decided on closed models, which can be run and compared.
Only \HyperionVTwentyOneDAGLayerFiveFraction{} of the rows of the index carry all five parts $(\Omega,\Xi,C,R,P)$: authors state the operation in one equation, the boundary in another and the measurement in a third, so that a mechanism is rarely written in one place and exists only along the derivation that joins its parts. Methods that read sentences or fit single formulas therefore do not see it.

\paragraph{Carriers and the terms they admit.}
Transitions are classified by a count, of the motions that become slow together and of the controls that are tuned, and the count is the same for a concentration, a phase and a magnetization. Mechanisms differ in what the count does not contain, the space in which the state lives, because that space decides which terms an equation may contain, what can be driven and measured, and what form the noise takes (Table~\ref{tab:carriers}). Of the \HyperionCorpusRows{} equation records, 10,136,295 state a carrier in their notation, and the atlas separates these into seventeen classes (Supplementary Section~S4), which Table~\ref{tab:carriers} groups by the state space they describe. Fields on a spatial domain account for four in ten of these records; mechanical coordinates, constrained states, the sites of graphs and lattices, amounts of reacting species, interfaces and spinors follow, and four classes state a condition on a space, its boundary values, initial values, parameter range or domain, and not the space itself.

A field evolves through its gradients and is not determined until values or fluxes are given on its boundary. On a graph the gradients become differences along edges, and the edges close into loops around which couplings can compete. A Hamiltonian flow preserves the volume of phase space, so that no state attracts its neighbours and nothing is stored in an attracting state until friction is added, and a constraint confines the state to a surface of the full space. The classes of the atlas are fitted to the mathematical apparatus that the text of an equation declares, and they do not resolve the distinctions on which the constructions of this work turn, which are read from the equations of each closed model (see above): an amount lives on a half-line that ends at a wall, a phase on a circle around which it can wind, a signed amplitude on a line, and a number of individuals on the integers. The space is not the whole specification: whether the wall of an amount is invariant depends on whether a source acts there, whether a population that has reached zero stays there depends on immigration, and a signed amplitude has a pitchfork only when the operation has a reflection symmetry. The nested object keeps these apart: the space belongs to the carrier, the source to the relations $C$ and the symmetry to the operation, and Table~\ref{tab:carriers} lists them in separate columns. Fields of science are separated by their state spaces, and by notation, and not by their transitions, which they share. A pair of spaces admits a definite set of couplings, and the empty block of amounts with phases is one such pair: a half-line and a circle admit a rate that the phase multiplies and a frequency that the amount shifts. The same reading applies to every pair of spaces in Table~\ref{tab:carriers}: the conditions in its third column state which couplings the pair admits, and each admitted coupling that the record does not contain is a construction of the kind carried out above for amounts and phases.

\begin{table}[!tp]
\centering
\caption{\textbf{State spaces in the written record and the conditions that each imposes on an equation.} The upper rows group the carrier classes of the atlas (Supplementary Section~S4) by the state space they describe, with the number of equation records in each group. Of the \HyperionCorpusRows{} records, 4,382,236 state no carrier, and 3,659,831 fall in four classes (boundary conditions, initial conditions, parameter regimes and membership of a domain) that state a condition on a space, which the nested object places in the relations $C$ and the protocol $P$. The lower rows are state spaces that the classes of the atlas do not separate and that are read from the equations of a closed model. The fourth column names the part of the nested object in which the further condition is stated, and the last gives physical systems whose state lives on the space.}
\label{tab:carriers}
\fontsize{7.4}{8.8}\selectfont
\setlength{\tabcolsep}{4pt}\renewcommand{\arraystretch}{1.25}
\begin{tabularx}{\textwidth}{@{}>{\raggedright\arraybackslash}p{0.11\textwidth}>{\raggedright\arraybackslash}p{0.15\textwidth}>{\raggedright\arraybackslash}X>{\raggedright\arraybackslash}X>{\raggedright\arraybackslash}p{0.11\textwidth}@{}}
\toprule
State space & Atlas classes; records & What the space imposes & What a further condition adds & Examples \\
\midrule
Field on a spatial domain & Scalar fields, vector fields, function spaces, coordinate domains; 4,038,990 & The state is a function of position. Local operations are built from gradients, and the evolution is determined only when values or fluxes are given on the boundary of the domain & Insulated ends (relations $C$) conserve the total amount and absorbing ends drain it; a flux that relaxes in a finite time turns diffusion into a damped wave & Heat conduction, concentration fields, elastic and electromagnetic fields \\
Mechanical coordinates and momenta & Configuration spaces, phase space; 809,689 & A Hamiltonian flow preserves the volume of phase space, so that no state attracts its neighbours & Friction or a bath (relations $C$) contracts the volume and makes attracting states possible, and with them a switch or a memory & Pendulum, rotor, orbits, molecular dynamics \\
State on a constraint surface & Variational constraints, conserved quantities; 621,963 & The state remains on a surface of the full space, and the constraint enters the equation as a multiplier or removes a variable & A conserved total (relations $C$) allows one variable to be eliminated; a constraint on velocities lets a closed cycle of controls displace the state & Compartments of fixed total mass, rolling bodies, incompressible flow \\
Sites of a graph or lattice & Graphs and differential forms, lattices and sites; 499,975 & Gradients become differences along edges, and a reciprocal current conserves the total amount. The edges close into loops, around which the signs or angles of the couplings can be summed & Couplings whose signs multiply to $-1$ around a loop (relations $C$) cannot all be satisfied, and the competition adds states that a drive can write & Diffusion on a network, spin lattices, networks of bistable elements \\
Amounts of species & Reaction species; 445,028 & The state is a set of non-negative amounts. The flow cannot point out of the wall at zero, so every loss vanishes with the amount and a coupling to a loss multiplies the amount & With no source at zero (relations $C$) the wall is invariant and a state can exchange stability with it; a feed or immigration removes this & Chemical reactors, growing populations, gene circuits \\
Interface or membrane & Interfaces and membranes; 33,208 & The domain is itself a variable: the state lives on a surface whose shape responds to it & Particles trapped at the interface (material $A$) deform it, and the deformation gives an interaction that depends on the bond angles & Particle monolayers, lipid membranes, droplets \\
Spinor & Spinors; 27,611 & The state is a complex amplitude whose norm is conserved, and a full rotation reverses its sign & Coupling to an environment (relations $C$) replaces the amplitude by a density operator, whose coherences decay & Spins, qubits, two-level atoms \\
\midrule
\multicolumn{5}{@{}l}{\emph{State spaces read from the equations of closed models}} \\
\midrule
Phase & Circle & Every term is periodic; the state can lock or wind, and the number of turns is counted & A rhythm of fixed frequency (drive $P$) gives a state locked to it & Clocks, coupled oscillators, Josephson junctions \\
Signed amplitude & Line & Both signs are admissible, and so is an additive force & A reflection symmetry of the operation $\Omega$ leaves only odd terms; when the cubic term limits the growth, the transition is a pitchfork and a sign is stored; without the symmetry the pitchfork unfolds into a fold & Sense of convection, magnetization, parametric oscillators \\
Number of individuals & Non-negative integers & Change occurs by discrete events of one or of several individuals, whose rates and sizes fix the noise; for a large population resting at a metastable state, lifetimes grow as $\exp(NS)$ & With no immigration (relations $C$) zero cannot be left & Populations of cells or agents, molecules in a small volume \\
Orientation on a lattice & Angle defined up to a fraction of a turn & Couplings are periodic with that fraction of a turn & A coupling bound to the bond angles (material $A$) makes the sums of angles around closed loops differ from zero, which frustrates the pattern & Particles at an interface, hexatic and nematic order \\
\bottomrule
\end{tabularx}
\end{table}

\paragraph{The record as a language.}
Taken together, these observations describe a language whose vocabulary is small and shared, since the same interaction terms were written independently in distant fields, and whose grammar is set by the carriers, which decide which terms may stand in one equation; the empty block of amounts with phases is a combination that this grammar forbids to the written terms. Models assembled from the written terms on the written carriers pass, in nine cases out of ten, through sequences of equilibria that written models already show, and what remains does not recur. No one designed this language; it formed as authors in separate fields reduced their systems in the same few ways, and the nested form makes it explicit. The recurrence is a property of sequences of equilibria along one control, for the terms and carriers that people have used. A term that the record does not contain is derived rather than selected from it. The defect of a failed move derives such a term, together with the condition under which it is required: the multiplicative coupling of an amount to a phase, which occurs in none of the 2,745 compiled models, the feed that follows a moving reference state, and the coupling of orientations to the bond angles of an interface are terms derived in this way; the last is the quadrupolar capillary interaction known for particles at a fluid interface~\cite{BaulinColloidMemory}, on a lattice that the test of saturation does not represent, and the product of a half-line with a circle is a carrier of this kind. Each adds to the vocabulary a term whose consequence has been calculated, whether or not the completed model passes through a familiar sequence.

\paragraph{A constitutive relation that adds a variable.}
If the heat flux $\mathbf{j}$ relaxes with a finite collision time $\tau$, $\tau\,\partial_t \mathbf{j} + \mathbf{j} = -D\nabla u$, instead of following Fourier's law at once, then together with conservation, $\partial_t u + \nabla\cdot\mathbf{j} = 0$, it gives the telegrapher equation
\begin{equation}
\tau\,\partial_t^2 u + \partial_t u = D\nabla^2 u.
\label{eq:telegrapher-cattaneo}
\end{equation}
A change in the constitutive relation has made the flux a new dynamical variable, turned diffusion into a damped wave with finite speed $\sqrt{D/\tau}$, and changed what must be prepared, the initial rate of change of the field as well as the field itself.

\paragraph{Four levels of correspondence, and transfers that hold.}
A proposed transfer $\tau: I_s \to I_t$ between a source and a target model is checked at four nested levels (Supplementary Section~S17):
\begin{equation}
\mathcal{L}_4 \;\subseteq\; \mathcal{L}_3 \;\subseteq\; \mathcal{L}_2 \;\subseteq\; \mathcal{L}_1 \;\subseteq\; \mathcal{T}.
\label{eq:invariance-ladder}
\end{equation}
At the first level each part $\Omega$, $\Xi$, $C$, $R$, $P$ of the source corresponds to a part of the same kind in the target. At the second the two models share their dimensionless Buckingham groups, $\boldsymbol{\pi}_s = \boldsymbol{\pi}_t$~\cite{Buckingham1914}. At the third the map carries trajectories into trajectories, $\alpha\circ\Phi_s = \Phi_t\circ\alpha$, which is $\Delta = 0$ in Eq.~\eqref{eq:nonlinear-residual-exact}. At the fourth the observables also agree, $R_t\circ\alpha = R_s$. When the defect $\Delta(t,q) = F_t(t,\alpha(t,q)) - D_q\alpha(t,q)F_s(t,q) - \partial_t\alpha(t,q)$ does not vanish, it is expanded over the terms available in the target, and $\mathcal{K}_t\alpha = -\Delta_{\alpha,\beta}$ is solved symbolically for the missing drift, memory kernel or protocol.

Three transfers of the main text hold exactly, the first in its conservation law and dissipation. Diffusion carried onto a graph with reciprocal currents obeys
\begin{equation}
J_{ij} = D w_{ij}(u_i - u_j), \qquad \dot{u}_i = -\sum_j J_{ij} = -D(L_G u)_i, \qquad w_{ij}=w_{ji}\ge 0.
\label{eq:constructed-graph}
\end{equation}
Because the rows and columns of $L_G$ sum to zero, the total $\sum_i u_i$ is conserved (to $2.44\times 10^{-15}$ in the computation) and the energy $\frac{1}{2}u^{\mathsf T}L_G u$ decays at the rate $-D\|L_G u\|^2$. On a $15\times 15$ grid with no flux through its edges ($h=1/14$, $D=0.05$) the graph operator matches the continuum Laplacian to 1.42\% in the interior. Drift and noise carried onto the parameters of a statistical model give Langevin sampling~\cite{Welling2011SGLD},
\begin{equation}
d\vartheta = -\nabla\mathcal{L}\,dt + \sqrt{2T}\,dW_t, \qquad \partial_t p = \nabla\cdot(p\nabla\mathcal{L}) + T\nabla^2 p,
\label{eq:constructed-sampling}
\end{equation}
whose density approaches the Gibbs density $p_\infty = Z^{-1}e^{-\mathcal{L}/T}$ where this density is normalizable and no probability flows through the boundaries: on a 401-point grid at $T=0.45$ the total-variation distance is 0.062, against 0.98 when the drift points uphill and 1.00 without noise. In a protocol of affine stages $\dot{q} = L_j q + b_j$, each stage carries the state to $q_{j+1} = e^{t_j L_j}q_j + \int_0^{t_j}e^{(t_j-s)L_j}b_j\,ds$, and leaving out stage $j$ shifts the final observable by $\delta r_j = R\bigl(\prod_{k>j}e^{t_k L_k}\bigr)(q_{j+1}-q_j)$.

\paragraph{What a construction records.}
A record contains the source equations and their references, the state variables, the carrier and its domain, the constitutive and boundary relations, the preparation and drive, the observable and the material parameters. A move states which of these it changes and keeps the model that results. A move is used in two ways: the changed model is kept and its own law derived, or a correspondence is prescribed and the defect is solved for the missing term. Continuation changes the value of the control; an unfolding adds a direction of control; a reduction supplies a map of coordinates and the regime in which it holds; adding or removing a term changes one stated equation. A fixed expression or a content hash only shows that a stored input has not changed; it is not a conserved quantity. A correspondence is checked by the relevant identity of equations or observables, and an intervention by the response that the changed equations produce. The next move starts from the resulting state and equations, written in the same form. The finite families of moves available are those of the executable records and tutorial of FieldBridge.

\paragraph{The construction as a procedure.}
A construction proceeds in the following steps. Take a written model in its nested form. Change one part to something that the record does not combine with it: a carrier from another field, a boundary, a drive along a second control, an observable, a material. Compute the defect of the change. If it vanishes, the mechanism holds in the new system and its laws are predictions there. If it does not, solve Eq.~\eqref{eq:correction-constraint} under the conditions that the new carrier imposes; the solution is a term that the record does not contain, together with the reason why it is needed. Derive the law that the completed model obeys and test it in the full dynamics. Then add the term and the carrier to the vocabulary and repeat, starting from the completed model. Each pass ends in an equation and a stated consequence, and each enlarges the language available to the next.

\paragraph{A further move.}
The workbench contains one move that the main text does not use. Two generators can be \emph{alternated} quickly, or the system switched between them; the defect is then the difference between the alternated evolution and the evolution under the mean generator.

\paragraph{When an instantaneous term suffices.}
For nonlinear dynamics, the defect follows the change of coordinates along trajectories. With source and target vector fields $F_A(t,q)$ and $F_B(t,q)$ and a map $\alpha(t,q)$ from source to target states, the chain rule gives the defect of Eq.~\eqref{eq:nonlinear-residual-exact}, in which $D_q\alpha$ is the Jacobian of the map. If the defect vanishes on the whole common domain, the two dynamics are intertwined. A correction that depends only on the retained target state exists if and only if all source states that the map sends to one target state require the same target rate. Otherwise the target must keep further coordinates or its history. For linear dynamics with $\beta=\alpha$ the condition reads $\ker\alpha\subseteq\ker(\alpha L_s)$, and it then fixes the reduced generator on $\mathrm{Im}\,\alpha$. The correction is $\mathcal K_t(\alpha q)=-\Delta q$, which is well defined because the defect $\Delta=\Omega_t\alpha-\alpha L_s$ then vanishes on $\ker\alpha$; on a complement of $\mathrm{Im}\,\alpha$ in the target the correction is not determined, and any two completions differ by a term that vanishes on every state the source can reach. The recorded completion is the smallest one that represents the defect, and when the condition fails it is the one with the fewest added coordinates among the candidates searched, which are monomials of the source coordinates (Supplementary Section~S31). For time-independent nonlinear maps the defects of two successive transfers compose as $\Delta_{21}(q)=\Delta_2(\alpha_1(q))+D\alpha_2(\alpha_1(q))\,\Delta_1(q)$, of which Eq.~\eqref{eq:composed-defect} is the linear case: the defect of the first transfer is carried forward by the Jacobian of the second. More generally, for linear generators $\{L_a\}$ and a linear observation $R$, the states that no sequence of operations can tell apart form the unobservable subspace $\mathcal N=\bigcap_{m\ge0}\bigcap_{(a_1,\dots,a_m)}\ker(R L_{a_m}\cdots L_{a_1})$. In the three-compartment exchange network with $x=p_0-1/3$ and $z=p_1-p_2$, symmetric coupling ($W_{01}=W_{02}=1$, $W_{12}=0.7$) puts $z$ inside $\mathcal N$, and $\dot x=-3x$ needs one coordinate. Unequal exchange ($W_{01}=4$, $W_{02}=1$) couples the two, $\dot x=-7.5x+1.5z$ and $\dot z=4.5x-3.9z$; $\mathcal N$ shrinks to $\{0\}$, two coordinates are needed and the defect is $\Delta=-1.5z$. Eliminating $z(t)$ gives the exact Mori--Zwanzig equation $\dot x(t)=-7.5x(t)+1.5e^{-3.9t}z(0)+\int_0^t6.75e^{-3.9(t-s)}x(s)\,\mathrm ds$, which reproduces 100 six-stage protocols to $4.34\times10^{-18}$. For a stochastic change of variables, the generators are compared on observables carried back by the map; there the quadratic variation $(\mathrm dX_t)^2=\mathrm dt$ produces the It\^o drift $\mathcal K_t=\partial_y$ for $y=x^2$.

\paragraph{Cycles of operations, curvature, frustration and memory.}
Horizontal and vertical shears of the plane are generated by
\begin{equation}
X=\begin{pmatrix}0&1\\0&0\end{pmatrix},\quad
Y=\begin{pmatrix}0&0\\1&0\end{pmatrix},\quad
Z=[X,Y]=\begin{pmatrix}1&0\\0&-1\end{pmatrix},
\label{eq:shear-commutator}
\end{equation}
whose brackets close on $\mathfrak{sl}(2,\mathbb R)$~\cite{Murray1993Nonholonomic}. Applying $-Y$, $-X$, $Y$ and $X$ in this order, each with strain $\epsilon$, gives
\begin{equation}
U_\epsilon=e^{\epsilon X}e^{\epsilon Y}e^{-\epsilon X}e^{-\epsilon Y}
=\begin{pmatrix}1+\epsilon^2+\epsilon^4&-\epsilon^3\\\epsilon^3&1-\epsilon^2\end{pmatrix}.
\label{eq:shear-cycle}
\end{equation}
If the material then relaxes under $L$ and is measured by $R$, its response differs from that of a preparation without the cycle by
\begin{equation}
\delta r(t)=Re^{tL}(U_\epsilon-\mathbb I)q_0
=\epsilon^2Re^{tL}Zq_0+O(\epsilon^3).
\label{eq:shear-observed-motion}
\end{equation}
For a symmetric relaxation $L$ with no positive eigenvalue, the quadratic response persists only if $RP_0Zq_0\ne0$, where $P_0$ projects onto the zero modes of $L$. These modes keep a deterministic displacement; in a noisy material an interaction must also confine the stored state. Cycles of both orientations, exact cycles and independent integrations are given in Supplementary Section~S26.

Three further calculations separate transport in space, bond preferences that cannot all be met, and the elimination of dynamical coordinates:
\begin{equation}
\begin{aligned}
H_{\partial S}-\mathbb I&=\frac{iq}{\hbar}\mathcal F_{xy}a^2+O(a^3),\\
\eta_\gamma&=\prod_{\langle ij\rangle\in\gamma}\operatorname{sign}J_{ij},\\
V_{t+s}-V_tV_s&=\Pi U_t(\mathbb I-E\Pi)U_sE.
\end{aligned}
\label{eq:obstruction-representations}
\end{equation}
The first is the holonomy of a smooth connection around an oriented square of side $a$, with the phase convention $H=\mathcal P\exp[(iq/\hbar)\oint A\cdot d\ell]$~\cite{YangMills1954,AharonovBohm1959,Lin2009Synthetic}. In the second, Ising preferences $s_i s_j=\operatorname{sign}J_{ij}$ with $\eta_\gamma=-1$ cannot all be satisfied around the cycle $\gamma$~\cite{Kirkpatrick1977}. The third holds for a full evolution $U_t$, a projection $\Pi$ and a lift $E$ of the prepared state with $\Pi E=\mathbb I$, where $V_t=\Pi U_tE$; it measures how far the reduced evolution fails to compose on its own~\cite{Mori1965,Zwanzig1961}. Curvature, frustration and memory are therefore statements about different relations. The square of the main text has four linear edges, each of which carries its source generator into its target generator with a composite defect of $2.9\times10^{-14}$; yet the maps composed along its two routes between opposite corners differ by a map of norm 8.44, so the difference belongs to the routes and not to any edge. For two routes $p$ and $q$ between the same models, with maps $\alpha_p$, $\alpha_q$ on states and $\beta_p$, $\beta_q$ on rates, subtracting the two relations of the form of Eq.~\eqref{eq:intertwining} gives $\Omega_t(\alpha_p-\alpha_q)-(\beta_p-\beta_q)\Omega_s=\Delta_p-\Delta_q$. When both routes are exact, the difference of the maps intertwines the two evolutions, and for invertible linear maps with $\beta=\alpha$ the relative transport $\mathfrak H=\alpha_q^{-1}\alpha_p$ satisfies $L_s\mathfrak H=\mathfrak HL_s$. The relative transport of exact routes is thus an element of the symmetry group of the source, and it changes a response only if $R_s\mathfrak H\neq R_s$. A frustrated cycle can produce competing local minima, although a single cycle with unequal bond strengths has one minimum up to a global reversal; whether it retains a state depends on its relaxation barriers, which must be calculated separately.

\paragraph{Following states and reducing them at a transition.}
FieldBridge~\cite{FieldBridge2026} follows equilibria by Newton--Raphson continuation as a declared control varies. A signed permutation $g$ of the coordinates is accepted as a symmetry when $F(gq) = gF(q)$ holds at 12 sampled states to a tolerance of $10^{-7}$ relative to the largest component of the drift over the samples; every pattern of signs is tried for up to four variables, and a common sign only for five and six. Where a single real eigenvalue crosses zero, with right and left eigenvectors $v$ and $w$ ($w^{\mathsf T}v = 1$), the transverse coordinates are solved at 41 amplitudes so that their drift vanishes, and a cubic polynomial is fitted to the drift $w^{\mathsf T}F$ along the critical direction. The curve on which the transverse drift vanishes is not an invariant curve of the flow; at the transition it gives the quadratic and cubic coefficients of the centre manifold and departs from it at fourth order in the amplitude. The window is halved up to eight times until the transverse residual is below $10^{-6}$ and the polynomial misfit below 5\%. When a sweep of one parameter meets a fold ($a_2 \neq 0$) or a one-sided write, continuation in a second material parameter locates the cusp ($a_1 = a_2 = 0$, $a_3 < 0$) and gives the direction in which both parameters must change together, tangent to the neutral curve.

\paragraph{Completing a reduction that stops.}
The obstruction at which a reduction stops determines what is kept. If several directions lose stability together, all of them are kept. At a finite sweep rate $r$, the completion also keeps every mode whose relaxation rate at the crossing is no greater than $\sqrt r$, because such a mode cannot relax within the passage time, of order $r^{-1/2}$. If the cubic term vanishes, the leading higher-order term is kept; if it is positive, the distant branch to which the state jumps; and for a one-sided write, a second material parameter together with a sweep through the cusp. A Hopf pair is passed to the phase derivation, and a family of written states that is marginal in one direction at every member is a phase.

The writing law of the completed model is then computed from the protocol. The control is swept so that the critical growth rate changes at the rate $r$, from a time $4r^{-1/2}$ before the crossing to the same time after it. The unwritten branch is followed through the crossing by Newton continuation. The mean $m$ and covariance $C$ of the deviation from this branch obey $\dot m=J(t)m+b(t)-\dot q^{*}(t)$ and $\dot C=JC+CJ^{\mathsf T}+2D\mathbb I$, where $b$ is the bias and $-\dot q^{*}$ the lag of the state behind the moving branch. These moments are integrated until the critical rate, integrated from the crossing, reaches 5. The resulting distribution is restricted to the kept directions and carried back by their linear propagator to one e-fold after the crossing, where its spread is compared with the distance to the emerging stable states. If the spread has already reached that distance, samples are assigned to the nearest state in direction. Otherwise they are evolved under the full nonlinear drift to the end of the sweep and assigned to the nearest stable state. Comparison with the original stochastic dynamics shows whether this split into a linear amplification and a nonlinear selection predicts the probabilities.

For a sweep through a cusp with a saturating cubic term, the law is taken instead from the equation for one coordinate along the moving branch, $\dot s=a_1(t)s+a_2(t)s^2+a_3(t)s^3+\beta(t)+h_s(t)+\sqrt{2D_s(t)}\,\xi$, with coefficients recomputed at 49 points of the sweep and $\beta=-w^{\mathsf T}\dot q^{*}$. This equation keeps the lag finite near the crossing, where the linear stage overestimates it. The sweep is read on a window narrowed until the quartic term of the drift no longer appears as a quadratic one; on the default window, the cusp of a driven Kerr oscillator reads as a transcritical crossing. No law is given when the sweep through the cusp is not a pitchfork, that is, when the second parameter has not removed the asymmetry along the sweep. A straight sweep through a cusp adds a bias, which is removed by curving the sweep, $\mathbf p(t)=\mathbf p_c+t\,\mathbf d+c_2t^2\,\mathbf n$, with $\mathbf n$ normal to the sweep direction $\mathbf d$ in the plane of the two parameters. The bias is read from the full stochastic dynamics at five curvatures, with a bias of one resolvable unit toward either state; its zero is placed between the two curvatures that bracket it and refined by secant steps until the remaining bias is below 0.1 or three standard errors.

Each law is compared with the original stochastic dynamics over the same sweep, with 2,000 trajectories for each bias. The noise is set by $D_s|a_3|/r=0.05$ where the cubic coefficient is resolved, and is the material noise otherwise. The biases are 0.5, 1 and 2 times the smallest bias that the write resolves. A law holds when every state probability agrees with the simulation to within four standard errors or 0.02; a disagreement marks an obstruction in the nonlinear stage. With 2,000 trajectories four standard errors are 0.045 at a probability of one half and 0.009 at a probability of 0.01, so the tolerance is absolute and is a larger fraction of a small probability than of a large one. Keeping the control at its final value for as long as the sweep then separates a choice that is still open after the linear stage from one that noise keeps exchanging between shallow written states. When more than a tenth of the trajectories change state during this interval, the write is a selection between equilibria; when more than a tenth have not yet left the unwritten branch at the end of the sweep, the write has not settled. Over the completed write points of the 31 models of quantum hardware this gives, in the FieldBridge record of 15 November 2026, 29 writes decided in the linear stage, five decided in the nonlinear stage, six selections between equilibria and one unsettled outcome. Diverging trajectories are counted separately and left out of the probabilities, which are therefore probabilities among the trajectories that stay bounded. For the coupled Kerr parametric oscillators with loss, the Lindblad equation is integrated in double precision up to $t=1000/K$, with the density matrix kept Hermitian at every step. Its steady state is compared with a balance equation over the eigenstates of the final Hamiltonian, with transition rates $\kappa\sum_i|\langle m|a_i|n\rangle|^2$; the two agree to within 0.002 in every configuration probability for one to three photons.

\paragraph{Writing laws tested in the original models.}
For the symmetric write, the critical growth rate is swept as $\varepsilon=rt$, and the bias and noise are projected onto the critical coordinate. Linear amplification then predicts the probability of the positive state, Eq.~\eqref{eq:symmetric-write-law}, where $\Phi$ is the standard-normal distribution function and $h_s$ and $D_s$ are the bias and diffusion coefficient in the reduced coordinate before the cubic term is scaled out. In the reduced equation $\dot s=rt\,s-|a_3|s^3+h_s+\sqrt{2D_s}\,\xi$, the substitution $s=D_s^{1/2}r^{-1/4}u$, $t=r^{-1/2}\tau$ gives $\mathrm du/\mathrm d\tau=\tau u-\Gamma u^3+h_s/(D_s^{1/2}r^{1/4})+\sqrt2\,\xi(\tau)$ with $\Gamma=D_s|a_3|/r$. The probability of each state therefore depends on the bias only through $h_s/(D_s^{1/2}r^{1/4})$ and on the material only through $\Gamma$; a term $a_5s^5$ enters as $a_5D_s^2r^{-3/2}$, and a transverse mode of relaxation rate $\lambda$ follows the critical coordinate when $\lambda\gg\sqrt r$. For $\Gamma\to0$ the equation is linear and its solution gives the constant $\pi^{1/4}$. A sweep that starts at $u=0$ and is confined to the window $|t|\le L/\sqrt r$ gives instead $\pi^{1/4}\operatorname{erf}(L/\sqrt2)/\sqrt{\operatorname{erf}L}$, which is 0.990 at $L=1$, 1.274 at $L=2$ and 1.3313 at $L=4$. The reduction is local to the crossing: it gives the probability of the written state and not the time for which that state is afterwards retained, which depends on the barrier far from the crossing. A mode slower than $\sqrt r$ violates the condition and changes the constant. In the dissipative cat qubit the quadrature that the write does not choose relaxes at a rate 0.05 at the write point, and at $\sqrt r=0.14$ the measured constant is $1.217\pm0.017$; when the sweep is slowed until $\sqrt r$ is below 0.05, the constant returns to $1.325\pm0.014$~\cite{FieldBridge2026}. The runs reported here set $D_s|a_3|/r=0.05$, choose the bias for a predicted accuracy of 0.8, and simulate the full nonlinear model. The comparison across realizations uses 2,000 trajectories for each realization, in eight independent streams. The uncertainty of the constant inferred from the measured accuracy is the larger of the scatter between runs and the propagated binomial error. Realizations that reach identical equations are grouped before the weighted comparison. The larger stored record uses 25,600 trajectories per realization in eight streams of 3,200; its eleven realizations represent eight distinct models and give $1.329\pm0.005$. The random streams depend on the identity of the material, so adding a model leaves the others unchanged.

At a fold, the canonical coordinate obeys $\dot x=\mu+x^2$ with $\mu=rt$, and the solution that follows the occupied branch is
\begin{equation}
x=r^{1/3}\frac{\mathrm{Ai}'(-r^{1/3}t)}{\mathrm{Ai}(-r^{1/3}t)}.
\label{eq:fold-delay-solution}
\end{equation}
It passes the static fold at $\mu=1.018793\,r^{2/3}$ and runs away at $\mu=2.33811\,r^{2/3}$~\cite{Haberman1979,Jung1990}. Nine canonical rates from $10^{-3}$ to $10^{-7}$ are tested without noise. Each sweep starts at $\mu=-5r^{2/3}$ on the branch that ends at the located fold; rates whose initial state lies beyond that branch are left out. The scaled crossing is fitted by a fourth-order polynomial in $r^{1/3}$ together with a correction $\delta r^{-2/3}$ for the location of the fold. The uncertainty includes the change with fit order, the change when one rate is left out and a numerical floor of $5\times10^{-7}$. These are deterministic laws of the crossing; escape driven by noise at slower rates is a separate regime.

For phase locking, an isolated limit cycle is reduced through its adjoint phase-response equation and a weak periodic drive is averaged. A resonant harmonic gives $\dot\psi=\Delta\omega-K\sin\psi$ with $K=\epsilon|Z_n|/2$~\cite{Adler1946}. Full driven trajectories at seven detunings and two amplitudes test relaxation inside the locking interval and phase slips outside it, and the inferred width is extrapolated to zero drive. The three laws, their equations and the derivations that stop are documented in the FieldBridge tutorial and Supplementary Section~S27.

\paragraph{The reactor construction and its stochastic test.}
Starting from the one-feed Schl\"ogl model ($a=4.5$, $k_3=5.75$, $b\in[1.075,2.675]$), continuation finds the two folds at $b=1.49009982$ and $2.25990018$. Freeing $a$ locates the cusp at $a_0 = \sqrt{3k_3} = 4.15331193$, $b_0 = (k_3/3)^{3/2} = 2.65350485$ and $c = \sqrt{k_3/3} = 1.38443731$, and gives the joint feed path $a(s)=a_0+s$, $b(s)=b_0-(k_3/3)s$ of Eq.~\eqref{eq:schlogl-path}, the line of Remlein and Seifert~\cite{RemleinSeifert2024}, along which Eq.~\eqref{eq:schlogl-cusp} holds identically. For the moving reference $\alpha(t,x) = x - m(t)$ of Eq.~\eqref{eq:schlogl-moving}, the feed $\dot m(t)$ is the solution of Eq.~\eqref{eq:correction-constraint}: the moving reference produces the term $-\dot m(t)$ in the equation for $u$, and $+\dot m(t)$ in the second feed of the reactor cancels it; it is returned by the symbolic calculation described below and is not supplied to it. With $k_3=5.75$ and $3m(t)^2 - k_3 = 0.02t$ over $t\in[-20,20]$ from $u(-20)=0.04$, a fourth- and fifth-order Runge--Kutta integration (relative tolerance $10^{-10}$, absolute tolerance $10^{-12}$) follows the target normal form to within $1.13\times 10^{-9}$; without the feed $\dot{m}(t)$ the trajectory departs by up to 0.632.

For the stochastic test of Fig.~\ref{fig:constructor}c,d, the full concentration equation $\dot{x} = -x^3 + a(s(t))x^2 - k_3 x + b(s(t)) + h + \sqrt{2D}\,\xi(t)$ was integrated along $s(t) = rt/(2c)$ with $r=0.02$ and $D=0.001$, from $x=c$ at $t=-4/\sqrt{r}$ to $t=+4/\sqrt{r}$. At the end of the sweep $s=0.204$, the two stable states lie at $u=0.861$ and $u=-0.657$, and the upper well is deeper than the lower one by $0.060$, that is $60D$, with a barrier of $0.056$ out of the lower well. The noise is additive, with the same $D$ at every concentration, and no condition is imposed at the wall $x=0$: the passage takes place at $c=1.38$, and the scale of the state during it, $D^{1/2}r^{-1/4}=0.08$, is far smaller. Nine scaled biases $h/(D^{1/2}r^{1/4})$ between $-1.25$ and $1.25$ were run with 4,096 trajectories each (seed 20261027), by the Euler--Maruyama method with steps $\Delta t = 0.005$ and $0.010$ sharing their Brownian increments. A weighted probit regression $\Phi^{-1}(P_+) = b_* + k_* h/(D^{1/2}r^{1/4})$ gives $k_* = 1.3289 \pm 0.0118$ and $b_* = 0.0176$, in agreement with $\pi^{1/4} = 1.3313$; fitting the coefficient of Eq.~\eqref{eq:symmetric-write-law} directly to the simulated probabilities gives $1.329\pm0.012$.

\paragraph{Changes of published models.}
The models are systems of ordinary differential equations taken from the \LaTeX{} sources of arXiv papers, each with the control and the range stated in its source. Of 100 extracted models, 13 were left out because their specification carries no parameter values from the source (three or more numerical parameters, all set to one), and 11 because their equations have the same terms with the same signs as those of a model already present, up to the names of variables and parameters; four of these are records of Lorenz's system. The comparison is structural and does not establish that two models behave alike. Of each such group the record with the widest control range is kept. The 76 remaining models, one representative of each structure, come from 17 archives and from two extractions, of 59 and 17 models (Supplementary Section~S29). Three kinds of change are applied to every model: removing one additive term from one equation; promoting another parameter to the control, swept from a quarter to four times its published value; and extending the range of the control by ten times its width below it, above it or on both sides. An extension stops a decade short of zero where a change of sign would change the meaning of the parameter: where the state variables are amounts, where the parameter stands in a denominator together with a state variable, and where it multiplies a whole equation. A fourth kind, which transplants an additive term from an equation of another model with a new coupling coefficient, is implemented but was not applied, because most of the candidate donor passages do not state a closed system. A changed model is discarded when its trajectories run away from every start: each of 32 starts grows to a hundred times its initial amplitude, or meets a drift that is not finite, at an end of the control range; a bounded model whose attractor lies a hundred times above its starts is discarded by the same rule. Each changed model is examined by the derivations of the symmetric write and of the threshold write, each limited to 60~s. A change is kept as a step when it reaches a write that the model did not reach, or changes the reason why a derivation stops. Pairs are formed from up to eight first changes per model, steps first, each followed by every change available on the changed model; two extensions of the same range are not paired. A pair closes when its outcome, that is, the writes it reaches together with the reasons why the other derivations stop, equals that of the model or of a single change, or combines outcomes that single changes produce; a derivation that fails numerically does not open a pair. Of the 2{,}087 pairs formed across the 76 models, 116 are open, and in 15 open pairs, in five models, the derivations give a bistable write. A write requires two isolated stable equilibria at one value of the control, and the equilibria of each changed model, found from several hundred starts, and the eigenvalues of their Jacobians decide it. Seven pairs have such states, all in the Hopf normal form. In two pairs the promotion alone already gives several stable states. Five have no isolated stable state, because the removals leave a continuum of equilibria or the drift is homogeneous. One has a single stable state at every value of the control. Changes of the protocol were also applied to the 11 repeated models, each with its own control and range, because what a range leaves out is a property of the study and not of the equations. Among the 87 models with their ranges the derivations give a write for five changes of the protocol, and the same calculation of equilibria decides them. Lorenz's model occurs five times among the 87, from four papers: three of its ranges of the Rayleigh parameter include the onset of convection, and two contain the chaotic range alone. Two recover a transition of the system, in Lorenz's model and in the R\"ossler flow; the other three are a control that the specification declares and the equations do not contain, a state that leaves to infinity, and a model that already has several stable states along its published control. The same procedure was applied to the 31 models of quantum-computing hardware. There, attaching to a new carrier strips the carrier and material parameters of the source, attaches the bare operation to the target carrier, and reports which entries of its signature are kept, which change and which belong to the carrier; one example carries the two-part alignment-and-reflection operation from $\pi$-periodic capillary rods, whose coupling decays as $r^{-4}$, to $2\pi$-periodic planar point dipoles, whose coupling decays as $r^{-3}$. A coupling between amounts and a phase on $\mathbb{R}_+^n \times S^1$ is admissible only if $F_i(\mathbf{x},\varphi)|_{x_i=0} \ge 0$ for every $\varphi \in S^1$ and $F$ is $2\pi$-periodic in $\varphi$. This excludes the additive forcing $+g\sin\varphi$ and admits the multiplicative term $g x_i \cos\varphi$.

\paragraph{Screens of feedback and amount--phase couplings.}
A separate automatic screen attached feedback and phase couplings to compiled models and asked whether a specified periodic response appeared. A coupling was attached only if it kept the states physical: on non-negative amounts the drift at a wall $x_i=0$ must not point outward, which a coupling $g\,x_i\cos\varphi$ satisfies wherever the source drift does and an additive term $g\sin\varphi$ violates for half of the phases. None of the 65 additional stable responses that the screen proposed survived matched preparations and two solvers. The responses that survived are locks of models with no oscillation of their own and oscillations produced by slow negative feedback, a known route (Supplementary Section~S33).
\paragraph{Exact checks against statistical judges.}
Every screen in this work that judged models by a statistical fingerprint produced false positives, and each was removed by an exact property of the equations: a part is attached only where the carrier admits it, so that no denominator can vanish and only periodic parts act on angles; a transition with several critical directions is read from the eigenvalues; and the constant of a writing law is measured only after the existence of a choice has been established. For a runaway whose trajectory from just below the threshold passes an edge state, an equilibrium of the diluted drift with a single unstable direction, the threshold that dilution creates is the stable manifold of that state, and close to the state the tangent plane of that manifold gives the threshold without integrating a trajectory; the other thresholds, between a fifth and a third of the pairs in each set, are set by an edge cycle or an edge at infinity. The cases and their counts are given in Supplementary Section~S33. A stationary state is not necessarily a state that can be written. A term such as $1/x$ separates the two sides of $x=0$: a stationary state across the pole cannot be reached from the original preparation by a trajectory of the model. The domain and the drive therefore matter as much as the stationary solutions in deciding what a model can store (Supplementary Section~S33).
\paragraph{Colloidal memory simulations.}
The simulations of ellipsoids at an interface follow the companion study~\cite{BaulinColloidMemory}. Monolayers of $N=1024$ apolar Brownian particles were placed on Voronoi networks with positional disorder $\sigma/a=0.11$ and 0.16. Capillary coupling $b_{ij}$ arises from quadrupolar meniscus deformations with contact-line amplitudes of $2\text{--}4\,$nm at surface tension $\gamma=10\,$mN/m; alignment $a_{ij}$ represents steric repulsion. A transient field writes the target pattern, and the particles then move by Brownian dynamics without field, with rotational diffusivity $D_r$. For the common preparation, in which the written pattern first relaxes, the connected overlap at $D_rt=625$ was $0.502\pm0.033$ and $0.473\pm0.013$ at the two disorders. Uncertainties are standard errors across five graph means (seeds 17, 29, 43, 71 and 97). Controls released from the same target at the same angles, removing capillary coupling or replacing it by equal-weight relative alignment lost this pattern, with mean $|Q_c|<0.004$ at the comparison time and standard errors across the five graph means of at most 0.005. The companion paper gives the derivation of the interactions, the controls of the preparation and the Brownian integration.

\paragraph{Steps executed automatically and steps derived by the authors.}
The following steps run from a specification of the equations, with no further input: the compilation of closed models from the passages that state them and the count of their interaction shapes; the three kinds of change applied to the published models and their pairs; the continuation of states, the detection of symmetries and the reduction at a transition; the completion of a reduction that stops, which for the reactor finds the folds, frees the second feed, locates the cusp and returns the joint path of Eq.~\eqref{eq:schlogl-path}; the stochastic tests of the writing laws; the admissibility conditions of a coupling on a carrier; the comparison of a network with its counterpart; and the test of saturation. The defect and the missing term of Eq.~\eqref{eq:correction-constraint} are computed from a map and a target that are given. For the reactor these are the drift $-x^3+3m x^2-k_3x+k_3m-2m^3$, the map $\alpha=x-m(t)$ and the target of Eq.~\eqref{eq:schlogl-moving}; the symbolic calculation returns the defect $\dot m(t)$ and the term $\mathcal{K}=\dot m(t)$, and the same function applied to the squared radius of a stochastic equation returns the It\^o term (included in the supporting files). Two results were derived by the authors within the same calculation and then tested numerically: the choice of Eq.~\eqref{eq:amount-phase} as the simplest member of the family that the two boundaries admit, made with an AI assistant under the direction of the authors after the analysis of the compiled models had located the empty block; and the term of Eq.~\eqref{eq:constructed-colloid} bound to the bond angles~\cite{BaulinColloidMemory}. The authors chose the source model and the aim of every construction, such as a symmetric write in the reactor. For the colloidal monolayer, an earlier version of the method proposed the system; the interfacial realization was chosen by the authors, and the calculations that followed were automatic.

\paragraph{Saturation of the record at one control.}
The written set is 5,020 compiled models with measured sequences of equilibria, from 3,915 papers. They come from a compilation of every paper of the corpus that writes a time derivative, made after the compilation of the 2,745 models whose interaction shapes are counted above; one model is kept for each distinct specification, and a model enters when its specification passes the screen of the compilation and its sequence of equilibria was measured. The two sets share papers, and neither contains the other. Reference models were drawn with the carrier, parameters, control and range of a written model; in 2,999 of them the terms come from the pool of written terms of the same dimension, excluding the paper of the template, and in 1,499 they are random polynomials of low order. Each equation received a confining term of higher degree, and on amounts a term that would push a variable through the wall had its sign reversed. A model is described by the sequence of its equilibria along the control (their number, the number of stable ones, whether one is marginal and whether a stable one is a focus) and by the attractors reached from random starts. Among the reference models built from written terms, 976 stay bounded and fall in 119 sequences; 86 of them (9\%) fall in 76 sequences that no written model shows, with at most four models in any one. Among the random polynomials, 46 of 564 (8\%) fall in 38 such sequences, a share close to that of the models built from written terms. Reference models therefore reach unwritten sequences, but none that holds more than four models. In the other direction, about two per cent of the written models are not reached at 800 reference models; most are copies of Lorenz's system, and the rest are the damped oscillator at critical damping and the cusp swept across its fold. The test covers closed systems of ordinary differential equations, one control and the terms that occur in the compiled models, and it records attractors by kind; within this description the record is near saturation when models are counted, 890 of 976, and not when distinct sequences are counted, 76 of the 119 being shown by no written model.

\paragraph{States of a network with the constructed coupling.}
For $\dot x_j=f_j(x)$ with $\dot x_i=f_i(x)+g\,x_i\cos\varphi$ and $\dot\varphi=\omega-hx_i$, the row of the Jacobian that belongs to the phase has the single entry $-h$, in the column of $x_i$, and the column of the phase has the single entry $-g\,x_i\sin\varphi$, in the row of $x_i$. Expanding along both gives $\det J=\pm\,g\,h\,x_i\sin\varphi_*\,\det J_{\rm r}$, where $J_{\rm r}$ is the Jacobian of the other species with $x_i$ fixed at $\omega/h$. The two steady states $\pm\varphi_*$ of one composition therefore have determinants of equal size and opposite sign, and they merge where $\sin\varphi_*=0$, that is, at $|f_i(x_*)|=g\,\omega/h$; the merging is a fold when $\det J_{\rm r}\neq0$ there and the control crosses this edge at a non-zero rate. In 52 networks of three species with random quadratic kinetics that admit a lock, the two determinants were opposite in every one. The wall $x_i=0$ contains no steady state, since $\dot\varphi=\omega$ there. The stability of the lock within the range is not fixed by this argument and depends on $J_{\rm r}$. For the single amount of Eq.~\eqref{eq:amount-phase} one lock of the pair is stable throughout the window, and a winding state can coexist with it: at $a=0.2$, $b=0.05$, $g=0.3$ and $\omega=h=1$, 11 of 48 preparations reach the lock and 37 reach a winding state of period 2.11 and mean amount 3.97, whereas at the parameters of Fig.~\ref{fig:colloid-case}c all 48 reach the lock. The condition for a lock was compared, model by model, with a numerical search for equilibria in 1,200 models assembled from written terms with the constructed coupling added: in the 485 models in which the search located equilibria at every control value, at 5,319 of 5,335 control values a steady state exists where one is predicted and is absent where none is (Supplementary Section~S33).
\paragraph{The events behind a noise.} The reduction to a density holds for the location and kind of a transition and fails for rare switches. For a large population of $N$ individuals resting at a metastable state the lifetime grows as $\exp(NS)$, and, when each event adds or removes one individual, the exponent $S=\int\ln(d/b)\,\mathrm{d}x$ is fixed by the birth rate $b$ and the death rate $d$ along the path of escape. For the fall of a budworm population from its low state to extinction the rates give $S=0.58$ along the direct path, where a noise of constant intensity gives $0.085$~\cite{FieldBridge2026}. The low state reaches the outbreak state over the smaller action $0.026$ and returns over $0.20$, so that the exponent of the mean time to extinction is $0.582+0.203-0.026=0.76$, where a noise of constant intensity gives $0.86$. What is missing here is not a single term: a mechanism that includes rare switches must carry the rates of its elementary events.

\paragraph{Single removals and further couplings.}
Removing one term shows what the term does: in 11 of the 76 models, a single removal turns a stable state into an oscillation and identifies the removed term as the one that suppresses the oscillation. Two further members of the amount--phase family of Eq.~\eqref{eq:amount-phase} were also examined: a response of the phase to the amount that depends on the phase, and a direct action of the rhythm on the phase. Departures of a real clock from the predictions of Eq.~(\ref{eq:amount-phase}) can come from these terms, and also from deformation of the cycle, delays and nonlinear sensing.

\paragraph{Retention among models with one normal form.}
The records of the constructor for 134 models, fixed on 30 October 2026, contain eleven with a supercritical pitchfork, of which three are Lorenz's system; all carry the same class of write. Predictions were written from the records before any simulation. Each model was then integrated independently by the Euler--Maruyama method with the noise of its specification, 300 trajectories. The control was swept through the transition to the upper end of its range in the retention time of the record ($T=56.6$; $47.1$ for one model), and the side chosen by each trajectory was measured along the left critical vector. Three intervals of duration $T$ at a fixed control followed: at the end of the range, at the end of the range with a bias against the chosen side of the size that would have decided the write with probability 0.98, and at the published value of the control. A sign counts as kept when fewer than 0.1 of the trajectories end on the other side; the criterion compares the sides at the beginning and at the end of an interval, so that a trajectory that crosses and returns counts as kept. At the end of the range the fraction that changed side was 0.47--0.54 in Lorenz's system, 0.45 in a pitchfork with a quadratic term and 0.29 in an optical parametric oscillator with pump depletion, and at most 0.06 in the six other models; a parametric oscillator with three-wave mixing lost 0.42 at its published control, 3\% beyond the transition, where two stable states exist. The normal form together with the position of the control predicted 19 of the 27 outcomes, and the records, which contain the number of stable states and a test of retention but no lifetime, 20. The entry that the records lack is the lifetime of each state against the declared noise. Three intervals share each model and three of the models are one system, so the outcomes are not independent, and this comparison does not distinguish the records from the normal form; the lifetime is the entry that both lack.

\paragraph{Data and code availability.}
Source \LaTeX{} documents are publicly available in the \href{https://huggingface.co/datasets/synthetix-institute/latex-data}{latex-data} collections, and extracted records, feature codebooks, and simulation outputs are detailed in Supplementary Section~S13. FieldBridge~\cite{FieldBridge2026} is publicly available at \href{https://github.com/synthetix-institute/fieldbridge}{github.com/synthetix-institute/fieldbridge} with an \href{https://synthetix-institute.github.io/fieldbridge/}{interactive demonstration} and \href{https://github.com/synthetix-institute/fieldbridge/blob/main/docs/tutorial/index.md}{executable tutorial} reproducing the reported derivations and calculations; it is free to examine and run under the PolyForm Noncommercial licence. The pipeline that reads the corpus and compiles its models comprises about 3,000 files and is released in part. A companion manuscript developing a domain-specific realization~\cite{BaulinColloidMemory} and the supporting scripts in \path{supporting_files/} accompany this submission.

\paragraph{AI assistance statement.}
OpenAI Codex, Anthropic Claude (Claude Code) and Google Antigravity assisted with coding, simulation execution, figure generation, numerical checks and manuscript drafting. The method described here was also used to create its own examples: the analysis of the compiled models located the gap between amounts and phases, the FieldBridge derivations and checks were applied to the resulting constructions, and the amount--phase model of Eq.~\eqref{eq:amount-phase} was constructed with an AI assistant (Claude Code) under the direction of the authors. The authors directed the scientific investigation, interpreted the results and verified the derivations and final text.

\paragraph{Acknowledgements.}
This work received no specific funding.

\paragraph{Author contributions.}
V.A.B.: conceptualization, methodology, software, formal analysis, investigation, validation, visualization, writing, and supervision. A.C.: corpus parsing software, equation featurization, and data curation. S.R.: manuscript review and editing.

\paragraph{Competing interests.}
The authors declare no competing interests.

\clearpage
\setcounter{figure}{0}\setcounter{table}{0}\setcounter{equation}{0}
\renewcommand{\thefigure}{S\arabic{figure}}
\renewcommand{\thetable}{S\arabic{table}}
\renewcommand{\theequation}{S\arabic{equation}}
\renewcommand{\theHfigure}{S\arabic{figure}}\renewcommand{\theHtable}{S\arabic{table}}\renewcommand{\theHequation}{S\arabic{equation}}
\begin{center}{\Large\bfseries Supplementary Information}\end{center}
\addcontentsline{toc}{section}{Supplementary Information}
\section*{Overview}

This Supplementary Information sets out the full construction behind the main text. It describes the
representation of an equation by its operation and its carrier, the typed graphs built on that representation, the
directed acyclic graph of completions, the atlas, the cost of a transition between representations, the roads (paths
of minimum cost through occupied states) and the constructor. It also reports the predictive tests, the null models,
the ablations, the extended constructions and the statistics of compiled models from which the gap between amounts and
phases was found. The publication atlas is computed from the full-corpus representation, whose descriptive coordinates are fixed before any predictive test is fitted. The
predictive tests refit \(\Om\), \(\Xis\) and \(\Tau\) on training papers alone and are evaluated on test papers that
took no part in the fit. The labels fitted for each split therefore measure how well the description generalizes; they
do not name cells of the publication atlas.

\paragraph{The text index.}
The features of the representation (S2) are deterministic regular expressions applied to the text of an equation and
its context. Each feature records whether a particular cue of notation appears: a dot or $\partial_t$ for a time
derivative, $\nabla$ for a spatial derivative, $\lambda$ for a spectrum, or angle brackets for an expectation value.
The features therefore group equations by the way they are written. Among the 14{,}518{,}531 rows, the 44 operation
features take 69{,}352 distinct values. In 40\% of the rows there is no operation cue at all, three values
account for half of all rows and 170 values account for 90\%, and a row with an operation cue shares its exact
operation vector with a median of 128{,}665 other rows. The 28 carrier features take 110{,}164 values; 30\% of the rows
carry no carrier cue, and a row with a carrier cue shares its carrier vector with a median of 111{,}190 others. The
leading principal directions of the operation features are the presence of $\lambda$, which switches on the spectral
and eigenvalue features together, and the presence of ``$=0$''. For the carrier features they are the presence of
$\phi$ or $u(\cdot)$, of $\in$, and of $\epsilon$ or $\ll$. Several operation and carrier features are triggered by the
same symbol: angle brackets signal both an expectation value and a Hilbert space, $\nabla$ signals both a spatial
derivative and a coordinate domain, and the string ``stoich'' signals both stoichiometric closure and stoichiometric
space. Their co-occurrence is consequently fixed by the rules, and some combinations are empty by rule rather than by
any property of physics. Identical notation also joins unrelated equations; the time-derivative cue, for instance, is
shared by an accretion rate $\dot m$ and by the voltage derivative of a neuron model. The clusters, the layout of the
atlas, the roads, the counts of occupancy and concentration, and the empty pairings reported in this supplement (S3,
S4, S6--S9, S18, S19, S22) therefore describe how physics is written. They form the layer of written
equations shown in main Fig.~\ref{fig:atlas}c,d, and their use is to retrieve source equations. Statistics of
mechanisms, including the gap between amounts and phases, are computed instead on closed models compiled from whole
papers (S30).

The constructions of the main text are deterministic calculations on specified equations. They detect symmetries,
continue solutions in a parameter, reduce models and solve constraints by fixed rules, and wherever numerical sampling
or simulation is needed it is carried out with fixed random seeds. The learned representation organizes the corpus and
proposes transitions, and the compiled models locate gaps. The amount--phase model of the main text was constructed
across one such gap by an AI assistant at the direction of the authors.

\paragraph{Guide to the sections.}
The corpus and the representation are described in S1, S2, S5, S12--S14, S18 and S19. The hierarchical description of
an equation state, the codebooks, the graphs, the atlas and the cost of a transition are given in S3, S4 and S6--S9.
The constructor and the transfers it performs are described in S10, S17 and S25, and the constructions in S26 and S27.
What an unoccupied operation--carrier pairing must acquire before it becomes a physical candidate is discussed in S22.
The null models, the baselines and the transfer controls are in S11, S15 and S16, and the composition of local
remainders and its numerical tests in S20 and S21. Hidden motion and the coordinates that a response requires are
discussed in S23 and S24. Section S28 explains which mechanisms need a critical point, and S29 changes published models
and composes the changes. Section S30 gives the statistics of compiled models, the extended examples and the
measurements of the apparatus, and S31 describes how a construction learned from a failed reduction was reused.
Analyses made with earlier versions of the representation and of the workbench, which the main text does not use, are
listed in S32, and S33 gives the screens of couplings, the exact checks against statistical judges and the comparison
of the lock condition with a numerical search for equilibria.

\section*{S1. Corpus, retained equation states and source records}

The input consists of 2,511,195 arXiv \LaTeX{} documents drawn from two Hugging Face datasets. The
representation retains \HyperionVTwentyOneCorpusRows{} equation states, each linked to its
source. A retained row is a local state of an equation's form, or a transition between two such states, stored together
with the identity of its document and the features detected in the equations at its end points. This row is the unit
that the operation and carrier codebooks (S4) assign to classes. A document, a source equation and a retained
transition row are therefore three distinct units.

Source cards are separate records that keep the original equations and the text around them. Each card stores the
identity of the document, the position of the equation in the local order, the section in which it appears, hints
about its variables, and evidence on the route by which it was reached and on the role the equation plays in building
the model. The graph of source constructions used in the analyses of S32 contains 14,354,227 cards
(its nodes), drawn from 2,151,957 source groups and joined by 12,202,270 edges, each edge lying within a single source.
A separate alignment wrote 2,420,388 links between 850,901 source cards and retained feature rows. These counts describe the sample used in S32 rather than the full-corpus rows of the publication representation. The two rates of pair extraction in
Table~\ref{tab:supp-corpus} also come from that sample.

\begin{table}[htbp]
\centering
\caption{Row counts of the representation and, marked as earlier, measurements of the source
records from the sample used in S32. Those pair-extraction rates are not fractions of completion in the full-corpus representation.}
\begin{tabular}{lr}
\toprule
Quantity & Value \\
\midrule
Documents processed & 2,511,195 \\
Rows retained in the representation & \HyperionVTwentyOneCorpusRows{} \\
Earlier rate of strictly complete pairs & 0.06645 \\
Earlier rate of partial or complete pairs & 0.89349 \\
Earlier source equation cards & 14,354,227 \\
Earlier constructor edges within sources & 12,202,270 \\
Earlier source groups & 2,151,957 \\
Earlier links between feature rows and source cards & 2,420,388 \\
\bottomrule
\end{tabular}
\label{tab:supp-corpus}
\end{table}

The way equation environments are treated, and what is excluded, is described in the Methods paragraph ``Reading equations
from the arXiv sources''.

\section*{S2. Equation features computed before clustering}

Each retained row is first described by a 192-dimensional feature vector computed by fixed rules, namely regular
expressions applied to the text of the equation and its context (their limits are stated in the Overview),
\begin{equation}
x_i\in\Rspace^{192}.
\end{equation}
The vector consists of eight contiguous blocks:
\begin{equation}
\begin{split}
\Rspace^{192}={}&
\Rspace^{24}_{\rm selector}
\oplus\Rspace^{40}_{\rm substrate}
\oplus\Rspace^{32}_{\rm operator}
\oplus\Rspace^{24}_{\rm closure}\\
&\oplus\Rspace^{24}_{\rm observable}
\oplus\Rspace^{16}_{\rm protocol}
\oplus\Rspace^{24}_{\rm equation\ shape}
\oplus\Rspace^{8}_{\rm quality}.
\end{split}
\label{eq:supp-raw-schema}
\end{equation}
Authorship, citations, field labels and named objects are kept with the source records and are excluded from this
vector. The identifiers of the row, of the document and of the source equation are also kept outside the vector, so
that the source can be attached again after a search or a construction has been carried out in the operational space.
Each coordinate that counts pattern matches is bounded separately. A typical count feature is
\begin{equation}
x_{ik}=\min\!\left(1,\frac{\log(1+c_{ik})}{\log 4}\right),
\end{equation}
where \(c_{ik}\) is the number of matches of the corresponding pattern. The coordinates that measure length, balance,
the use of commands and quality are bounded by transforms of their own.

For the graph and variational calculations the raw blocks are collected into three streams:
\begin{equation}
x_i^{\rm op}=(A_i,C_i,R_i,P_i),\qquad
x_i^{\rm sub}=(G_i,S_i),\qquad
x_i^{\rm ctl}=(H_i,Q_i).
\label{eq:supp-streams}
\end{equation}
Here \(A\) denotes the operator apparatus, \(C\) closure or admissibility, \(R\) evidence of an observable, a flux or a
current, \(P\) the order of the protocol, \(G\) the geometry of the carrier, \(S\) the selector or context, \(H\) the
shape of the equation and \(Q\) the quality of the extraction. Coordinates that vanish on every row of the full corpus
are removed before the matrices are fitted. The active coordinates that remain, fixed before fitting, number 44 in the
operator stream, 28 in the substrate stream and 23 in the control stream, 95 in total. The streams regroup the blocks
of Table~\ref{tab:supp-blocks}. The operator stream collects the columns of apparatus (16), closure (10), observable
(10) and protocol (8); the substrate stream collects those of carrier geometry (16) and selector (12); and the control
stream collects those of equation shape (15) and quality (8). Section~S2a lists all 192 coordinates with their block,
definition and type and states whether each is active, so that the regrouping can be checked coordinate by coordinate.

The predictor of the next transition (S5) separates its inputs further. The part of the predictor that represents the
operation receives only the 16 active apparatus columns. The closure, observable and protocol columns are withheld from
both the operation part and the carrier part and are supplied separately as the completion fiber. The carrier part
receives the 28 active carrier coordinates. This separation keeps the mask of completions, which is the quantity to be
predicted, from entering the learned \(\Om\) coordinate.

\begin{table}[htbp]
\centering
\caption{Feature blocks and their use in the present analysis.}
\begin{tabular}{p{0.20\linewidth}p{0.12\linewidth}p{0.56\linewidth}}
\toprule
Block & Raw dimensions & Use \\
\midrule
Selector/context & 24 & Selectors of domain, carrier and context \\
Substrate & 40 & Evidence of geometry, graph, lattice, field and carrier \\
Operator & 32 & Derivative, integral, spectral, algebraic and transport apparatus \\
Closure & 24 & Constraints, admissibility and closure conditions \\
Observable/current & 24 & Evidence of an observable, a flux, a current or a measured output \\
Protocol/order & 16 & Evidence of an ordered realization or an intervention \\
Equation shape & 24 & Controls for presentation and structure \\
Quality & 8 & Controls for extraction, balance and entropy \\
\bottomrule
\end{tabular}
\label{tab:supp-blocks}
\end{table}

The control stream is kept for negative controls and for screening out spurious signals, and it carries zero weight by
default in the action of a mechanism. Dual presentations of the same equation, such as its differential and integral
forms or its strong and weak forms, are represented by the features of route, equation shape and source context. They
are checked separately from changes of substrate.

\section*{S2a. The 192 coordinates of the feature vector}
Each row of the table below is one coordinate of the 192-coordinate deterministic feature layout. A coordinate is marked active when it is nonzero somewhere in the fixed export of the full
corpus. Reserved coordinates are allocated by the layout but receive no feature in this version; they are kept so that
the offsets of the blocks stay the same.
\begingroup\small
\begin{longtable}{p{0.06\linewidth}p{0.06\linewidth}p{0.18\linewidth}p{0.27\linewidth}p{0.14\linewidth}p{0.08\linewidth}}
\toprule
Index & Offset & Block & Coordinate definition & Type & Active \\
\midrule
\endhead
0 & 0 & selector & domain\_membership & pattern count & yes \\
1 & 1 & selector & function\_space & pattern count & yes \\
2 & 2 & selector & normalization & pattern count & yes \\
3 & 3 & selector & positivity & pattern count & yes \\
4 & 4 & selector & boundary\_admissibility & pattern count & yes \\
5 & 5 & selector & initial\_condition & pattern count & yes \\
6 & 6 & selector & variational\_constraint & pattern count & yes \\
7 & 7 & selector & parameter\_regime & pattern count & yes \\
8 & 8 & selector & gauge\_choice & pattern count & yes \\
9 & 9 & selector & admissible\_set & pattern count & yes \\
10 & 10 & selector & conservation\_selector & pattern count & yes \\
11 & 11 & selector & probability\_selector & pattern count & yes \\
12--23 & 12--23 & selector & reserved zero channels (12) & reserved & no \\
24 & 0 & real substrate geometry & coordinate\_domain & pattern count & yes \\
25 & 1 & real substrate geometry & metric\_manifold & pattern count & yes \\
26 & 2 & real substrate geometry & hilbert\_inner\_product & pattern count & yes \\
27 & 3 & real substrate geometry & phase\_symplectic\_space & pattern count & yes \\
28 & 4 & real substrate geometry & probability\_space & pattern count & yes \\
29 & 5 & real substrate geometry & graph\_topology & pattern count & yes \\
30 & 6 & real substrate geometry & lattice\_site\_space & pattern count & yes \\
31 & 7 & real substrate geometry & bundle\_gauge\_space & pattern count & yes \\
32 & 8 & real substrate geometry & stoichiometric\_space & pattern count & yes \\
33 & 9 & real substrate geometry & configuration\_quotient & pattern count & yes \\
34 & 10 & real substrate geometry & boundary\_interface & pattern count & yes \\
35 & 11 & real substrate geometry & scalar\_field\_carrier & pattern count & yes \\
36 & 12 & real substrate geometry & vector\_field\_carrier & pattern count & yes \\
37 & 13 & real substrate geometry & tensor\_field\_carrier & pattern count & yes \\
38 & 14 & real substrate geometry & spinor\_carrier & pattern count & yes \\
39 & 15 & real substrate geometry & differential\_form\_carrier & pattern count & yes \\
40--63 & 16--39 & real substrate geometry & reserved zero channels (24) & reserved & no \\
64 & 0 & operator apparatus & time\_derivative & pattern count & yes \\
65 & 1 & operator apparatus & spatial\_derivative & pattern count & yes \\
66 & 2 & operator apparatus & laplacian\_diffusion & pattern count & yes \\
67 & 3 & operator apparatus & integral\_operator & pattern count & yes \\
68 & 4 & operator apparatus & hamiltonian\_generator & pattern count & yes \\
69 & 5 & operator apparatus & liouvillian\_generator & pattern count & yes \\
70 & 6 & operator apparatus & spectral\_operator & pattern count & yes \\
71 & 7 & operator apparatus & graph\_operator & pattern count & yes \\
72 & 8 & operator apparatus & reaction\_operator & pattern count & yes \\
73 & 9 & operator apparatus & markov\_transition & pattern count & yes \\
74 & 10 & operator apparatus & control\_update & pattern count & yes \\
75 & 11 & operator apparatus & variational\_derivative & pattern count & yes \\
76 & 12 & operator apparatus & commutator\_operator & pattern count & yes \\
77 & 13 & operator apparatus & stochastic\_operator & pattern count & yes \\
78 & 14 & operator apparatus & nonlinear\_operator & pattern count & yes \\
79 & 15 & operator apparatus & nonlocal\_operator & pattern count & yes \\
80--95 & 16--31 & operator apparatus & reserved zero channels (16) & reserved & no \\
96 & 0 & closure constraints & zero\_residual & pattern count & yes \\
97 & 1 & closure constraints & conservation\_balance & pattern count & yes \\
98 & 2 & closure constraints & constitutive\_law & pattern count & yes \\
99 & 3 & closure constraints & algebraic\_constraint & pattern count & yes \\
100 & 4 & closure constraints & normalization\_closure & pattern count & yes \\
101 & 5 & closure constraints & gauge\_constraint & pattern count & yes \\
102 & 6 & closure constraints & stoichiometric\_closure & pattern count & yes \\
103 & 7 & closure constraints & weak\_form\_closure & pattern count & yes \\
104 & 8 & closure constraints & boundary\_condition & pattern count & yes \\
105 & 9 & closure constraints & compatibility\_condition & pattern count & yes \\
106--119 & 10--23 & closure constraints & reserved zero channels (14) & reserved & no \\
120 & 0 & observable current & eigenvalue\_spectrum & pattern count & yes \\
121 & 1 & observable current & observable\_expectation & pattern count & yes \\
122 & 2 & observable current & current\_flux & pattern count & yes \\
123 & 3 & observable current & order\_parameter & pattern count & yes \\
124 & 4 & observable current & free\_energy\_action & pattern count & yes \\
125 & 5 & observable current & residual\_error & pattern count & yes \\
126 & 6 & observable current & probability\_readout & pattern count & yes \\
127 & 7 & observable current & material\_response & pattern count & yes \\
128 & 8 & observable current & biological\_response & pattern count & yes \\
129 & 9 & observable current & classifier\_control\_output & pattern count & yes \\
130--143 & 10--23 & observable current & reserved zero channels (14) & reserved & no \\
144 & 0 & protocol order & recurrence\_step & pattern count & yes \\
145 & 1 & protocol order & algorithmic\_protocol & pattern count & yes \\
146 & 2 & protocol order & measurement\_order & pattern count & yes \\
147 & 3 & protocol order & intervention\_sequence & pattern count & yes \\
148 & 4 & protocol order & training\_recall & pattern count & yes \\
149 & 5 & protocol order & write\_hold\_read & pattern count & yes \\
150 & 6 & protocol order & feedback\_loop & pattern count & yes \\
151 & 7 & protocol order & discrete\_update & pattern count & yes \\
152--159 & 8--15 & protocol order & reserved zero channels (8) & reserved & no \\
160 & 0 & equation shape & latex\_command & pattern count & yes \\
161 & 1 & equation shape & subscript\_superscript & pattern count & yes \\
162 & 2 & equation shape & brace\_structure & pattern count & yes \\
163 & 3 & equation shape & bracket\_structure & pattern count & yes \\
164 & 4 & equation shape & relation\_operator & pattern count & yes \\
165 & 5 & equation shape & display\_environment & pattern count & no \\
166 & 6 & equation shape & matrix\_or\_array & pattern count & yes \\
167 & 7 & equation shape & fraction\_sqrt & pattern count & yes \\
168 & 8 & equation shape & text length & numeric summary & yes \\
169 & 9 & equation shape & word count & numeric summary & yes \\
170 & 10 & equation shape & LaTeX command count & numeric summary & yes \\
171 & 11 & equation shape & relation count & numeric summary & yes \\
172 & 12 & equation shape & bracket count & numeric summary & yes \\
173 & 13 & equation shape & subscript/superscript count & numeric summary & yes \\
174 & 14 & equation shape & identifier count & numeric summary & yes \\
175 & 15 & equation shape & clean endpoint & numeric summary & yes \\
176--183 & 16--23 & equation shape & reserved zero channels (8) & numeric summary & no \\
184 & 0 & quality & source\_clean & quality flag & yes \\
185 & 1 & quality & target\_clean & quality flag & yes \\
186 & 2 & quality & pair\_clean & quality flag & yes \\
187 & 3 & quality & chemical\_formula\_or\_reaction & quality flag & yes \\
188 & 4 & quality & logical\_relation & quality flag & yes \\
189 & 5 & quality & word\_heavy & quality flag & yes \\
190 & 6 & quality & garbage\_or\_fragment & quality flag & yes \\
191 & 7 & quality & identifier\_diversity & quality flag & yes \\
\bottomrule
\end{longtable}
\endgroup

\section*{S3. Hierarchical description of an equation state}

The nested description distinguishes four objects:
\begin{equation}
\begin{aligned}
M_i&=(\Om_i,\Xis_i), & F_i&=(C_i,R_i,P_i),\\
I_{i,\mathrm{op}}&=(M_i;F_i), &
\mathcal I_{i,\mathrm{real}}&=(I_{i,\mathrm{op}};A_i).
\end{aligned}
\label{eq:supp-hierarchy}
\end{equation}
The hierarchy is generative: each object is obtained from the one before it by adding information. The operation
\(\Om\) and the carrier \(\Xis\) together form the mechanism core \(M\). The completion fiber \(F\) then adds the
requirements of closure, observable and protocol, which gives the operational identity \(I_{\mathrm{op}}\). Finally the
realization layer \(A\) assigns the names of objects, the parameters, the units, the boundary data and the source
context, which gives \(\mathcal I_{\mathrm{real}}\). The semicolon in Eq.~\eqref{eq:supp-hierarchy} denotes attachment:
\(F\) is a fiber that depends on the two-factor core, not a third independent factor space.

The hierarchy removes a common ambiguity. Two rows with the same \(M\) may state different information about closure
or about the observable; they then represent the same coarse mechanism core at different stages of completion.
Conversely, the same occupied \(\Om\) can occur together with different occupied values of \(\Xis\), and each such
occurrence supplies a hypothesis of transfer that can be tested. ``Theoretical kernel'' is a status given to a boundary
state \((\Om,0_{\Xis})\) that is found in a source equation. It is not a fifth object, nor another name for every
\(\Om\) label. The dual state \((0_{\Om},\Xis)\) records a carrier for which no operator apparatus was detected.

Every retained row carries explicit flags that mark a missing operation and a missing carrier.
These boundaries distinguish an operation whose carrier was not identified, a carrier whose operation was not
identified, and a mechanism core in which both are present. The counts of concentration are reported separately for
the full archive and for the rows in which both factors are present, so that incomplete records of equations cannot be
read as complete physical mechanisms.

The null boundaries must be analysed, and boundary states linked to their source equations, with one and the same fixed
codebook. Boundary families are linked to source equations only when both calculations used identical codebooks. This
prevents example equations from an earlier fit from being assigned to an \(\Om\) or \(\Xis\) class of the present fit.

A summary of fixed width uses five bits for \(\Om\), five for \(\Xis\), three for the mask that records which
completions are present and three for a summary of the dominant route, which gives an upper bound of 16 bits for the
coarse index shown in the figures. This index labels the state and the evidence of its route. Constants, assumptions,
assignments of variables, boundary terms and the context of the derivation remain in the source cards and the source
records.

The constructor extends a state in the same order,
\begin{equation}
\begin{aligned}
\widehat M&=(\Om_i,\Xis'),\\
\widehat I_{\mathrm{op}}&=(\widehat M;\widehat C,\widehat R,\widehat P),\\
\widehat{\mathcal I}_{\mathrm{real}}
&=(\widehat I_{\mathrm{op}};\widehat A),
\end{aligned}
\label{eq:supp-construct-identity}
\end{equation}
where the source operation keeps a pointer \(p_i\) to its source equation. The candidate carrier is attached first, the
roles of completion are supplied next, and the variables, named objects, references and experimental conditions are
assigned during realization. Detachment is a later step of the constructor, in which a clause supported by the source
is selected for transfer; it plays no part in defining the hierarchy.

\section*{S4. Operation and carrier codebooks and the assignment of all rows}

The operator and substrate codebooks are fitted independently of each other to the standardized active features. The
codebooks of the publication atlas are fitted to at most 500,000 rows sampled from training papers, taken from the end
points of transitions found in source equations, and they then assign all \HyperionVTwentyOneCorpusRows{} retained rows
to their nearest centroids. The numbers of classes are choices of the model, evaluated on test papers by occupancy,
stability and description length. They are not constants of nature inferred from the data.

\begin{table}[htbp]
\centering
\caption{Variables of the representation used by the publication atlas and the predictive
tests.}
\begin{tabular}{llrp{0.43\linewidth}}
\toprule
Layer & Symbol & Count & Function \\
\midrule
Coarse operation & \(\Om\) & \HyperionVTwentyOneOmegaStates{} & \HyperionVTwentyOneOmegaNonNullStates{} populated physical classes and one null boundary \\
Coarse carrier & \(\Xis\) & \HyperionVTwentyOneXiStates{} & \HyperionVTwentyOneXiNonNullStates{} populated physical classes and one null boundary \\
Fine operation & \(\Om_{\mathrm{fine}}\) & \HyperionVTwentyOneOmegaLeaves{} & Nested operator form within one coarse class \\
Fine carrier & \(\Xis_{\mathrm{fine}}\) & \HyperionVTwentyOneXiLeaves{} & Nested domain or geometry within one coarse class \\
Completion & \((C,R,P)\) & 8 masks & Presence of closure, observable and order of operations \\
Source transition & \(\Tau\) & 8 & Directed changes used only for prediction on test papers \\
\bottomrule
\end{tabular}
\label{tab:supp-language}
\end{table}

The publication atlas fixes one assignment of all rows and one nested system of leaves. The predictive tests refit
their codebook of transitions on training papers, but the labels fitted for each split never rename cells of the atlas.
Every retained row of the atlas stays linked to its source equation and to its position in the document, so the
physical meaning of a populated class can be read from its source equations instead of being inferred from a centroid
alone.

\subsection*{Operation and carrier coordinates}

The \HyperionCorpusRows{} equation states, each linked to its source, were described by 44 features on the side of the
operation, 28 features of the state space and 23 contextual controls. The operation atlas uses the 16 active apparatus
coordinates, which separate differential, Hamiltonian, commutator and variational forms from the clauses of completion.
The features of the state space resolve its dimension, its topology and the tensor rank of the carried quantity.

Two-level mini-batch $k$-means clustering in this operational space yields two coupled atlases
(Fig.~\ref{fig:atlas}): \HyperionVTwentyOneOmegaNonNullStates{} populated operation classes and
\HyperionVTwentyOneXiNonNullStates{} populated carrier classes, with one explicit null boundary in each factor. These
classes refine into \HyperionOmegaLeaves{} operation leaves and \HyperionXiLeaves{} carrier leaves.

The two atlases are distinct but coupled through conditional dependence. Part of this dependence is physical, because
an operation can act only where $\Xi$ supplies its domain: a differential operator requires a suitable field domain, a
commutator requires an algebra and a graph Laplacian requires an adjacency. Part of it is fixed by features that are
triggered by the same symbol (Section~S2). The atlas records which combinations of operation and carrier are attested in
the literature; whether their domains and physical conditions permit a transfer is decided by the corresponding
equations.

Tables~\ref{tab:supp-omega-classes} and~\ref{tab:supp-xi-classes} list each class with its code in
Fig.~\ref{fig:atlas}, its number of rows and the features that define it, all computed on the full set of rows.
Two pairs of classes have the same detected features and differ only in their weights: $\Omega_{15}$ weights the
integral more heavily than the time derivative and $\Omega_{17}$ does the reverse, while $\Omega_{08}$ and
$\Omega_{26}$ differ by the time derivative.

{\footnotesize
\begin{longtable}{@{}l p{4.1cm} r p{7.6cm}@{}}
\caption{Operation classes of the text index, with the codes used in
Fig.~\ref{fig:atlas}c. Rows: the number of the 14{,}518{,}531 equation records assigned to the class. Defining
features: the features detected in at least 20\% of the rows of the class, ordered by their share times the logarithm
of their enrichment. The share is the fraction of the class's rows in which the feature is detected, and the enrichment
is that fraction divided by the fraction among all rows. The names summarize these features. They name combinations of
detected mathematical apparatus, not physical mechanisms; features triggered by the same symbol, such as an expectation
value and a Hilbert inner product, occur together by construction (Section~S2).}
\label{tab:supp-omega-classes}\\
\hline
Class & Name & Rows & Defining features (share of the class's rows; enrichment) \\
\hline
\endfirsthead
\multicolumn{4}{l}{\tablename~\thetable\ (continued)}\\
\hline
Class & Name & Rows & Defining features (share of the class's rows; enrichment) \\
\hline
\endhead
$0_{\Omega}$ & no feature detected & 8{,}825{,}812 & --- \\
$\Omega_{01}$ & spectra with commutators & 37{,}548 & spectral operator (1.00; 9.3$\times$); eigenvalue spectrum (0.98; 9.4$\times$); commutator (0.68; 13$\times$); spatial derivative (0.21; 5.2$\times$) \\
$\Omega_{02}$ & spatial derivatives with integrals & 15{,}619 & spatial derivative (0.91; 23$\times$); integral (1.00; 10$\times$) \\
$\Omega_{03}$ & commutators in time evolution & 62{,}891 & commutator (1.00; 19$\times$); time derivative (1.00; 7.8$\times$) \\
$\Omega_{04}$ & nonlinear terms & 169{,}220 & nonlinear term (1.00; 53$\times$) \\
$\Omega_{05}$ & Markov transitions & 17{,}763 & Markov transition (1.00; 762$\times$) \\
$\Omega_{06}$ & Liouvillians and expectations & 398{,}828 & Liouvillian (1.00; 25$\times$); expectation value (0.44; 11$\times$) \\
$\Omega_{07}$ & spatial derivatives & 307{,}403 & spatial derivative (1.00; 25$\times$) \\
$\Omega_{08}$ & spectra and eigenvalues & 1{,}065{,}943 & spectral operator (1.00; 9.3$\times$); eigenvalue spectrum (0.97; 9.2$\times$) \\
$\Omega_{09}$ & control and feedback & 31{,}121 & control update (1.00; 262$\times$) \\
$\Omega_{10}$ & Hamiltonians & 90{,}704 & Hamiltonian (1.00; 98$\times$) \\
$\Omega_{11}$ & noise & 7{,}087 & stochastic term (0.92; 2036$\times$); algebraic constraint (0.21; 2.0$\times$) \\
$\Omega_{12}$ & integrals & 807{,}304 & integral (1.00; 10$\times$) \\
$\Omega_{13}$ & variational derivatives & 11{,}761 & variational derivative (1.00; 577$\times$) \\
$\Omega_{14}$ & diffusion operators & 24{,}906 & Laplacian or diffusion (1.00; 291$\times$); spatial derivative (0.68; 17$\times$); graph operator (0.26; 527$\times$) \\
$\Omega_{15}$ & integrals in time evolution & 161{,}826 & integral (1.00; 10$\times$); time derivative (1.00; 7.8$\times$) \\
$\Omega_{16}$ & commutators with spectra & 32{,}610 & commutator (1.00; 19$\times$); spectral operator (0.83; 7.7$\times$); eigenvalue spectrum (0.77; 7.4$\times$) \\
$\Omega_{17}$ & time evolution with integral terms & 51{,}026 & integral (1.00; 10$\times$); time derivative (1.00; 7.8$\times$) \\
$\Omega_{18}$ & variational dynamics & 12{,}723 & variational derivative (1.00; 577$\times$); time derivative (0.52; 4.0$\times$); spatial derivative (0.24; 6.1$\times$); integral (0.32; 3.2$\times$) \\
$\Omega_{19}$ & Markov spectra & 196 & Markov transition (0.83; 634$\times$); spectral operator (1.00; 9.3$\times$); eigenvalue spectrum (0.94; 9.0$\times$) \\
$\Omega_{20}$ & spectra of integral operators & 130{,}993 & integral (1.00; 10$\times$); spectral operator (1.00; 9.3$\times$); eigenvalue spectrum (0.97; 9.3$\times$) \\
$\Omega_{21}$ & diffusion equations & 24{,}710 & Laplacian or diffusion (1.00; 291$\times$); spatial derivative (0.92; 23$\times$); time derivative (0.61; 4.8$\times$) \\
$\Omega_{22}$ & Liouvillians with integrals & 54{,}822 & Liouvillian (1.00; 25$\times$); integral (1.00; 10$\times$); expectation value (0.51; 13$\times$) \\
$\Omega_{23}$ & commutators & 488{,}569 & commutator (1.00; 19$\times$) \\
$\Omega_{24}$ & time evolution of fields & 165{,}747 & spatial derivative (1.00; 25$\times$); time derivative (1.00; 7.8$\times$); equation set to zero (0.30; 2.3$\times$) \\
$\Omega_{25}$ & nonlinear dynamics & 80{,}078 & nonlinear term (1.00; 53$\times$); time derivative (0.43; 3.4$\times$); integral (0.33; 3.3$\times$); spectral operator (0.33; 3.0$\times$) \\
$\Omega_{26}$ & time evolution of spectra & 135{,}760 & spectral operator (1.00; 9.3$\times$); eigenvalue spectrum (0.97; 9.3$\times$); time derivative (1.00; 7.8$\times$) \\
$\Omega_{27}$ & control dynamics & 22{,}403 & control update (1.00; 262$\times$); spatial derivative (0.23; 5.9$\times$); integral (0.34; 3.4$\times$) \\
$\Omega_{28}$ & commutators with integrals & 66{,}247 & commutator (1.00; 19$\times$); integral (1.00; 10$\times$) \\
$\Omega_{29}$ & Liouvillian spectra & 51{,}862 & Liouvillian (1.00; 25$\times$); spectral operator (1.00; 9.3$\times$); eigenvalue spectrum (0.96; 9.2$\times$); expectation value (0.41; 10$\times$) \\
$\Omega_{30}$ & time derivatives & 1{,}113{,}149 & time derivative (1.00; 7.8$\times$) \\
$\Omega_{31}$ & Hamiltonian dynamics & 51{,}900 & Hamiltonian (1.00; 98$\times$); time derivative (0.35; 2.7$\times$); integral (0.30; 3.1$\times$) \\
\hline
\end{longtable}
}

\begin{table}[!htbp]
\centering
\footnotesize
\caption{Carrier classes of the text index, with the codes used in Fig.~\ref{fig:atlas}d.
Rows: the number of the 14{,}518{,}531 equation records assigned to the class. Defining features: the features
detected in at least 20\% of the rows of the class, ordered by their share times the logarithm of their enrichment. The
share is the fraction of the class's rows in which the feature is detected, and the enrichment is that fraction divided
by the fraction among all rows. The names summarize these features. They name combinations of detected mathematical
apparatus, not physical mechanisms; features triggered by the same symbol, such as an expectation value and a Hilbert
inner product, occur together by construction (Section~S2).}
\label{tab:supp-xi-classes}
\begin{tabular}{@{}l p{4.1cm} r p{7.6cm}@{}}
\hline
Class & Name & Rows & Defining features (share of the class's rows; enrichment) \\
\hline
$0_{\Xi}$ & no feature detected & 4{,}382{,}236 & --- \\
$\Xi_{01}$ & graphs and differential forms & 180{,}373 & graph (1.00; 78$\times$); differential form (0.87; 45$\times$); membership of a domain (0.20; 1.1$\times$) \\
$\Xi_{02}$ & function spaces & 846{,}984 & function space (1.00; 9.8$\times$); scalar field (0.24; 1.2$\times$) \\
$\Xi_{03}$ & boundary conditions & 907{,}714 & boundary admissibility (1.00; 10$\times$); scalar field (0.20; 1.1$\times$) \\
$\Xi_{04}$ & coordinate domains & 751{,}181 & coordinate domain (1.00; 12$\times$) \\
$\Xi_{05}$ & parameter regimes & 1{,}169{,}311 & parameter regime (1.00; 6.8$\times$); Hilbert inner product (0.25; 3.8$\times$) \\
$\Xi_{06}$ & reaction species & 445{,}028 & stoichiometric space (1.00; 24$\times$); membership of a domain (0.23; 1.3$\times$) \\
$\Xi_{07}$ & membership of a domain & 1{,}285{,}590 & membership of a domain (1.00; 5.7$\times$); scalar field (0.27; 1.4$\times$) \\
$\Xi_{08}$ & vector fields & 779{,}665 & vector field (1.00; 14$\times$) \\
$\Xi_{09}$ & scalar fields & 1{,}661{,}160 & scalar field (0.71; 3.7$\times$); Hilbert inner product (0.21; 3.1$\times$) \\
$\Xi_{10}$ & lattices and sites & 319{,}602 & lattice site (1.00; 37$\times$); parameter regime (0.23; 1.6$\times$) \\
$\Xi_{11}$ & initial conditions & 297{,}216 & initial condition (1.00; 39$\times$); scalar field (0.23; 1.2$\times$) \\
$\Xi_{12}$ & conserved quantities & 50{,}669 & conservation condition (1.00; 264$\times$); coordinate domain (0.40; 5.0$\times$); membership of a domain (0.24; 1.4$\times$) \\
$\Xi_{13}$ & interfaces and membranes & 33{,}208 & interface or membrane (1.00; 382$\times$); boundary admissibility (0.24; 2.5$\times$) \\
$\Xi_{14}$ & configuration spaces & 703{,}071 & configuration space (1.00; 15$\times$) \\
$\Xi_{15}$ & variational constraints & 571{,}294 & variational constraint (0.98; 22$\times$); membership of a domain (0.54; 3.1$\times$) \\
$\Xi_{16}$ & spinors & 27{,}611 & spinor (1.00; 500$\times$); boundary admissibility (0.85; 8.9$\times$); parameter regime (0.24; 1.6$\times$) \\
$\Xi_{17}$ & phase space & 106{,}618 & phase space (1.00; 106$\times$); scalar field (0.22; 1.2$\times$) \\
\hline
\hline
\end{tabular}
\end{table}

\section*{S5. Choice of the representation by prediction on test papers}

We chose the structure of the representation by asking how well each
candidate compresses papers that were held out of the fit, measured by their
description length. Every model encoded the same destination tuple,
\begin{equation}
Y_j=(\Om_j,\Xis_j,C_j,R_j,P_j),
\end{equation}
with conditional probabilities estimated with additive smoothing. The models
differed only in what the encoder was allowed to know about the source state.
We compared a code built from global frequencies alone, a code conditioned on
each single factor, a code conditioned on the product of the factors, and a
code conditioned on that product together with the completion fiber.

The structure selected by this comparison is
\begin{equation}
\Om\times\Xis\quad\hbox{with an attached }(C,R,P)\hbox{ fiber}.
\end{equation}
Its predictive information is measured below on two independent divisions of
complete papers into training and test sets, both made in the second version
of the atlas. We report each gain as a fraction of the mean code length of the baseline of
its own split, because raw bit counts from separately fitted transition
codebooks cannot be compared directly.

\subsection*{Control with random partitions of the features}

The comparison above uses codes built from the same blocks of features, so it
cannot tell whether the data prefer the particular assignment of features to
an operation block and a carrier block. To test this, we reassigned the 16
apparatus features and the 28 carrier features at random to two blocks of the
same sizes, 50 times, and fitted and scored each random partition exactly as
we fitted and scored the operation--carrier partition. Every model predicted
the same 44 destination indicators and was given the same completion roles of
the source. The evaluation contained 13{,}943 transitions from 5{,}385 test
papers and used three fitting seeds (271829, 271831 and 271837).

With $8\times6$ codebooks, the completion roles alone required 0.170472 bits
per destination feature. The operation--carrier partition required 0.165327
bits, a reduction of 3.02\%, while the random partitions required 0.164467
bits on average, a reduction of 3.52\%. Forty-eight of the 50 random
partitions predicted better, so that the operation--carrier partition ranked
49th of 51. The 95\% interval for the mean difference, computed by pairing
the scores paper by paper, is $[0.00055,0.00115]$ bits per feature in favour
of the random partitions, and each fitting seed gives the same sign. With
$32\times18$ codebooks the operation--carrier partition ranked 51st of 51.
With $4\times3$ codebooks it ranked 11th, and its advantage over the mean of
the random partitions, $0.00032$ bits per feature with interval
$[-0.00003,0.00064]$, is not resolved. When the values of the features were
used instead of their presence, the operation--carrier partition ranked 46th
of 51.

The comparison therefore supports the coded features and the completion fiber
over global frequencies and over single factors, but it does not support the
particular division of the features into an operation block and a carrier
block. That division rests on a physical presupposition, namely that a
Laplacian requires a geometry and a commutator requires an operator algebra.
We use it as a construction that we impose, not as a finding of the corpus.

\subsection*{Validation on complete test papers}

For the confirmatory predictor, complete papers were divided into training
and test sets before any codebook was fitted. The definitions of the features
and the requested numbers of classes were fixed in advance. The
standardization and the centres of the \(\Om\), \(\Xis\) and \(\Tau\) classes
were fitted on the training papers and kept fixed when the rows of the test
papers were assigned to classes. The predicted quantity was
\begin{equation}
P\!\left(T_2,\Om_{t+1},\Xis_{t+1},F_{t+1}
\mid z_{\Om,t},z_{\Xis,t},F_t,T_1\right),
\label{eq:supp-predictive-target}
\end{equation}
where \(T_1\) is the observed move into the current state and \(T_2\) is the
next move within the same source. The encoder of the operator was given the
apparatus features only.

The two independent evaluations comprised $8{,}383$ chains from $2{,}479$
test papers and $8{,}647$ chains from $2{,}500$ test papers. Knowing the
current mechanism state removed $22.1\%$ and $23.1\%$ of the baseline code length
of the next transition in the two evaluations, and $7.9\%$ and $7.1\%$ of the
code length of the joint destination. The normalized skill in predicting a change
of operation, of carrier or of completion lay between $0.292$ and $0.313$
across both splits.

\subsection*{Recovery of a hidden clause}

In a separate test, also made in the second version of the atlas, one clause
of the current operational identity is hidden in the test papers and has to
be recovered from the others. Given \((\Om,F)\) the test recovers \(\Xis\);
given \((\Xis,F)\) it recovers \(\Om\); and given \((\Om,\Xis)\) it recovers
the completion mask. The exact and top-three accuracies are computed from the
same class assignments that were used for main Fig.~\ref{fig:funnel}.

The strict screen of compositions fitted its probability calibration on a
separate set of calibration papers and then evaluated, in the test papers, 429
cells that were absent from the training papers. Of these, 43 cells (10.02\%)
were realized, and the screen saved 0.0688 bits per event relative to
\(P(T_2\mid T_1)\), with a 95\% interval of 0.0159--0.1216 bits. The Wilson
lower bound on the yield was 7.53\%, above the threshold of 5\% that had been
set in advance for passing cells to the constructor. The calibration test,
however, failed: the mean predicted probability of realization, 6.82\%, lies
below the observed interval [7.53\%, 13.23\%], so the model understates the
very rate it is used to rank. These cells therefore carry predictive structure
and exceed the yield threshold, but they were not passed to the constructor as
targets for scientific construction.

\section*{S6. Four linked graphs}

The analysis keeps two kinds of information apart, in two primary networks: the provenance graph and the operational graph. The source layer builds the provenance graph, which records who or what produced a result and with which data and tools. It indexes documents, citations, named objects and the equations of the sources. The operational graph addresses a harder question: which equations and constraints connect a result to a mechanism, and which observation could disprove it? It indexes transformations, substrates, completion roles and those transitions between them that the sources support. Four linked graphs describe the different relations within and between these layers: the source graph (provenance), the graph of mechanism similarity, the completion graph and the graph of the order of equations within a source. The source constructor graph links the two layers, because it connects the order of equations in a document to changes of their operational roles.
The operational layer is the main index used for discovery. A typed operational edge states which factor is retained, which factor is changed and what evidence of completion supports the change; it therefore imposes a stronger condition than a pairwise similarity of equations or of text. The source layer is used afterwards, once a candidate road or construction has been selected, to audit it retrospectively, to interpret it and to validate it against its sources.

\begin{longtable}{p{0.21\linewidth}p{0.34\linewidth}p{0.35\linewidth}}
\toprule
Graph & Nodes and edges & Question answered \\
\midrule
\endhead
Source graph (provenance) & Papers, source equations, variables, objects and the records
of their extraction & Who or what produced the result, and with which data and tools? (Where did the evidence originate?) \\
Typed mechanism graph & Retained rows joined by edges of six kinds: near in operator,
near in substrate, coupled, transfer, control and sequence & Which formal states are close
under each criterion of similarity? \\
Completion DAG & Typed edges oriented so that the number of roles
or the construction rank increases & Which moves found in the sources add roles that a
complete model requires? \\
Source constructor graph & Equation cards of a source linked by their local order in the document
and by changes of role & How are equations assembled and projected inside a
paper? \\
\bottomrule
\end{longtable}

\ifHyperionVTwentyOneDAGMetrics
In the second version of the atlas, the typed mechanism graph contains the
following numbers of retained edges:
\begin{center}
\begin{tabular}{lr}
\toprule
Edge family & Retained edges \\
\midrule
Operator nearest neighbour & \HyperionVTwentyOneOperatorEdges{} \\
Carrier nearest neighbour & \HyperionVTwentyOneCarrierEdges{} \\
Coupled operation--carrier & \HyperionVTwentyOneCoupledEdges{} \\
Operator-transfer candidate & \HyperionVTwentyOneOperatorTransferEdges{} \\
Carrier-transfer candidate & \HyperionVTwentyOneCarrierTransferEdges{} \\
Control-only candidate & \HyperionVTwentyOneControlEdges{} \\
Within-paper sequence & \HyperionVTwentyOneSequenceEdges{} \\
\bottomrule
\end{tabular}
\end{center}
\fi
The large families of nearest-neighbour edges are capped when the graph is
built. Their counts therefore describe the graph that was used in the
calculations of the second version of the atlas, not the number of all
possible neighbours in the archive.

\section*{S7. Completion levels and the order of equations within a paper}

The completion layer of a row is the number of roles detected in it among
five: carrier or context, operation, closure, observable and order of
operations. The completion DAG (directed acyclic graph) of the second version
of the atlas therefore contains six layers, numbered from zero to five. An
edge is given a direction only when its destination states more of this
physical structure than its origin, that is, a larger number of roles, which need not include the roles of the origin; reformulations within one layer and
candidate transfers remain in the typed mechanism graph.

\ifHyperionVTwentyOneDAGMetrics
\begin{table}[htbp]
\centering
\caption{Occupancy of the completion layers in the DAG of the second version of the atlas.}
\begin{tabular}{rrr}
\toprule
Roles present & Rows & Fraction \\
\midrule
0 & \HyperionVTwentyOneDAGLayerZero{} & \HyperionVTwentyOneDAGLayerZeroFraction{} \\
1 & \HyperionVTwentyOneDAGLayerOne{} & \HyperionVTwentyOneDAGLayerOneFraction{} \\
2 & \HyperionVTwentyOneDAGLayerTwo{} & \HyperionVTwentyOneDAGLayerTwoFraction{} \\
3 & \HyperionVTwentyOneDAGLayerThree{} & \HyperionVTwentyOneDAGLayerThreeFraction{} \\
4 & \HyperionVTwentyOneDAGLayerFour{} & \HyperionVTwentyOneDAGLayerFourFraction{} \\
5 & \HyperionVTwentyOneDAGLayerFive{} & \HyperionVTwentyOneDAGLayerFiveFraction{} \\
\bottomrule
\end{tabular}
\label{tab:supp-dag-layers}
\end{table}

The input of the DAG is the six families of mechanism edges; the
\HyperionVTwentyOneControlEdges{} control-only similarities are not part of it. The DAG retains \HyperionVTwentyOneDAGDirectedEdges{} of these
\HyperionVTwentyOneDAGInputEdges{} typed mechanism edges
(\HyperionVTwentyOneDAGDirectedFraction{}). The candidate transfers are kept outside the DAG, and the other remaining edges either connect
states at the same level of completion or describe a change that does not add
an explicit role.
\fi

The source constructor graph assigns roles directly to the equation cards of
a source, in their order in the document. It therefore supplies the local
order of the equations of one document and their context, whereas the DAG
compares completion across all retained rows of the archive. A complete
source card and an atlas state in layer five are consequently different
measurements and need not coincide.

\section*{S8. Layout of the atlas figures}

The atlas is constructed separately in the space of operations and in the
space of carriers. In each space we first fit a two-dimensional UMAP
projection to the standardized active rows with a fixed random seed; if UMAP
is unavailable, a PCA projection is used instead. Fine leaves are fitted in
the same typed spaces, and each leaf is assigned to the factor class that
contains the plurality of its rows. The atlas shown in the publication then
replaces a single flat projection by a layout on three scales: the factor
states, the occupied leaves within each state, and the residual positions of
individual rows within each leaf.

At the coarsest scale, the end points of the retained typed roads are
converted into pairs of leaves. The counts of these pairs define a weighted
graph of the occupied leaves, and summing them again over the leaves of each
state defines the graph of factor states. A weighted spring layout with a
fixed random seed (\(300\) iterations, \(k=1.35/\sqrt{n}\)) places the factor
states, starting from the normalized medians of their leaves in the initial
projection. The layout is scaled to a spacing of 5.2 display units between
factor states. It is therefore a map of how similar the traffic of roads is
across the whole graph, and not a second fit of the factor codebooks.

At the intermediate scale, the centroids of the leaves in the initial
projection are centred and scaled within each factor state to an island
radius of 0.78. At the finest scale, each displayed row is expressed relative
to the median of its leaf in a local frame of principal components. The
components are divided by their robust absolute scale at the 90th percentile,
transformed coordinate by coordinate with \(\operatorname{asinh}\), normalized
at the 99.5th percentile of the radius and drawn with a residual scale of
0.16. This procedure makes the structure of occupancy inside a leaf visible
while preventing a few outlying rows from stretching the whole panel.

The two detailed panels of the atlas are selected at the level of leaves.
Each occupied leaf is assigned to the factor class that contains the plurality
of its rows. For each state shown, the inset takes the most populated leaf
assigned to that state and shows all displayed rows of that leaf in the
unchanged coordinates of the parent map. Rows with the same class label in
other leaves are not included. Conversely, rows of a minority class that
already lie in the selected leaf remain visible, because they are part of the
measured support of that island. The procedure magnifies one resolved
\(\Omega_{27}\) island and one resolved \(\Xi_{07}\) island; it does not
define a new projection of coordinates.

The Euclidean separation in the resulting figure is the equilibrium between
an attraction weighted by traffic and the repulsion of the graph layout. It
is neither an edge score nor a calibrated geodesic distance, and it cannot be
compared between the operation and carrier panels, which are fitted
independently. A central state has a broadly connected profile of traffic; a
peripheral state has narrower or weaker connections or belongs to a
specialized branch. Rotation, reflection and the absolute origin are
arbitrary. Local branches show the local structure of the features, but their
displayed lengths are not distances calibrated across the whole map.

The centre of the display has no assigned axiomatic meaning. Hierarchy is
measured by the completion DAG, and compression by the description length of
test papers. The local density of points measures how densely the selected
display or feature metric is occupied; it is not used as a curvature or as a
density of axioms. The reconstruction residual of an auxiliary encoder is the
mean squared error between an input row and its reconstruction. Such a
residual can grow because of genuine rare structure, but also because of
sparse features, noise in the extraction, unequal block sizes or the limited
capacity of the model. We therefore exclude it from the symbolic selection,
from the ranking of kernels and voids, from the calculation of the action and
from the selection of candidates for the constructor. A test of an axiomatic
radial hierarchy would require labels of axioms, definitions, theorems and
applications taken from the sources, complete papers withheld for testing,
and controls for the norm of the input, its sparsity, its quality and the
seed of the model.

\begin{figure}[!htbp]
\centering
\includegraphics[width=\linewidth]{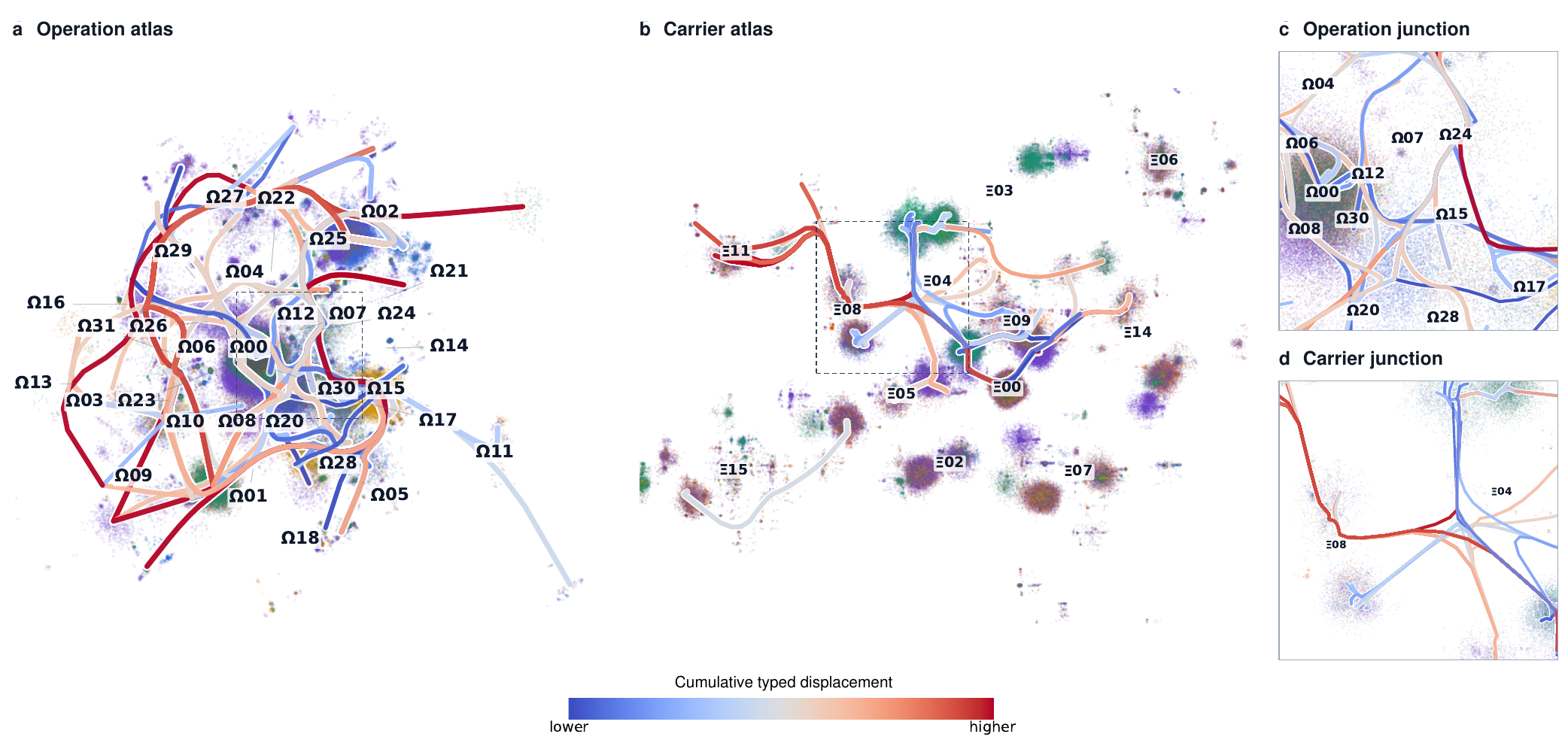}
\caption{\textbf{Roads of least displacement through the operation and carrier features.} (A)~Occupied leaves on the operation side and (B)~on the carrier side, joined by typed edges of the graph. Each curve is the route of least cumulative displacement of the features through the local graph of occupied neighbours.
(C,D)~Enlarged junctions where several routes pass through the same intermediate operation or carrier leaves. The displacement along a road measures the change of the extracted features of the equations. A road locates neighbourhoods of sources in which a transformation can be sought; it does not transport states, observables or conserved quantities, and a physical correspondence is established only by the calculated defect of an explicit transformation (Section~S25).}
\label{fig:supp-roads}
\end{figure}

The atlas itself contains no roads. Supplementary Fig.~\ref{fig:supp-roads} constructs paths of
several steps through occupied leaves on a separate graph, weighted by the
transition cost described in S9.

\section*{S9. Transition cost in the representation and its variation}

For each typed edge \(i\to j\), the cost in the representation used in the
calculations is
\begin{equation}
d_{\rm rep}(i,j)=\sum_b w_b\left\|I_b(j)-I_b(i)\right\|_2^2,
\label{eq:supp-cost}
\end{equation}
where \(b\) runs over the blocks of substrate geometry, selector or context,
operator apparatus, closure, observable and protocol. The control blocks have
zero weight by default. For a family \(F\) of edges, the empirical generator
is the mean displacement
\begin{equation}
G_F(i)=\left\langle I_j-I_i\right\rangle_{i\to j\in F}.
\end{equation}
The first variation of the cost in the direction of this generator,
evaluated on the sampled graph, is
\begin{equation}
\delta_{G_F}d_{\rm rep}
=\sum_{i\to j}\sum_b
2w_b\,[I_b(j)-I_b(i)]\cdot[G_{F,b}(j)-G_{F,b}(i)].
\label{eq:supp-first-variation}
\end{equation}
The cost in the representation orders routes through occupied leaves; it is
not a mechanical action. Because the cost of an edge is the square of its displacement, a displacement $d$ divided into $n$ equal steps costs $d^2/n$, so the route of least cost favours densely occupied regions and depends on the sampling of the leaves and on the number of neighbours. Any physical meaning comes from the clauses that the
equations at the two ends of a route retain. The calculation in S19, made in
the second version of the atlas on rows linked to their source equations,
tests whether operation, carrier, closure, observable and protocol persist
along routes, against permutations of complete papers within strata of papers of similar size. Any physical action,
continuous symmetry, boundary term or conserved quantity that is used in a
subsequent map preserving the mechanism is taken from the source equation.

The roads weighted by transition cost run through occupied leaves. Each leaf
is represented by the row nearest to its median in the full space of
features. A graph joining each representative row to its 10 nearest
neighbours is built separately in each view, with at most 160,000 nodes.
Every edge of this graph receives the exact typed block cost of
Eq.~\eqref{eq:supp-cost} between its two representative rows. The paths shown
in the publication use this cost as their weight, not the distance in the
two-dimensional display. Dijkstra's algorithm returns the paths of least
cumulative cost through several leaves. Direct chords between end points,
forced bridges between disconnected components and straight lines in place of
missing paths are all excluded. The displayed polyline is smoothed only after
the discrete path has been selected. The colour of a path reports its
cumulative cost; the bending of the plotted lines follows the two-dimensional
layout and plays no part in the minimization. It measures neither a gauge
curvature nor a renormalization-group flow.

\section*{S10. Construction of new mechanisms from supported ones}

A construction starts from a core that the sources support and adds a role
that the sources do not state. For an operation \(\Om\), a candidate carrier
\(\Xis'\) and an empty class of the atlas, the proposed identity is
\begin{equation}
\widehat I'=\bigl((\Om,\Xis');\widehat C,\widehat R,\widehat P\bigr).
\end{equation}
The hats mark roles that have not yet been realized from sources or tested.
The construction then proceeds around the cycle
\begin{equation}
\begin{split}
\mathrm{representation\ proposal}\;&\longrightarrow\mathrm{verified\ interface}
\longrightarrow\operatorname{Real}(\widehat I_{\mathrm{op}})\\
&\longrightarrow\mathrm{solver,\ fabrication\ or\ experiment}
\longrightarrow\mathrm{measured\ consequence}\\
&\longrightarrow\mathrm{graph\ update}.
\end{split}
\end{equation}

The atlas ranks possible attachments by typed edges and by distances between
the extracted features of equations. The completion DAG (Section~S7) shows
which role is absent, the source constructor graph gives the order of the
nearby equations, and the source cards (S1) return the original variables and
their context. A candidate expression is then derived from the proposed
identity. Checks of dimensions, closure, residuals, boundaries, negative
controls and experiment decide whether the proposal is
supported or rejected; the ranking by the atlas decides nothing by itself.

A construction can belong to one of several classes. Role completion adds one
of \(C,R,P\) and keeps the supported core. Operator transfer keeps \(\Om\) and
replaces or attaches \(\Xis\); substrate transfer keeps \(\Xis\) and replaces
or attaches \(\Om\). Transition composition proposes a path through fixed
\(T\)-classes. A deformation that preserves an invariant changes a
parameterized realization while keeping an invariant \(Q\) derived from the
source. The design of an observable or of a protocol changes \(R\) or \(P\) at
a fixed core. Finally, a completion that was predicted with high probability
but failed may indicate that a retained clause should be revised. This last
case is treated as a diagnostic proposal unless an alternative model passes
independent tests.

Each class is accepted under its own conditions. Role completion requires
that the context of the source make the added role necessary and that the
completed model be formally closed. Transfer of a factor further requires a
test on data from the target field that were not used in the construction,
and a consequence that is absent from the source realization. Composition
requires every intermediate state and the final composed state to be
admissible. A deformation that preserves an invariant requires the invariant,
calculated independently, to stay fixed, and not merely to keep the same
symbolic label. Proposals of observables and protocols require observability,
identifiability and a feasible intervention. A revision prompted by a failed
prediction requires the revised mechanism to do better than the failed
completion on formal or empirical evidence that was not used to construct it.
All classes keep the controls of dimensions, limiting cases, residuals and
negative cases.

Each of the six construction verbs edits one clause and keeps the others,
under the following conditions:
\begin{longtable}{p{0.14\linewidth}p{0.23\linewidth}p{0.23\linewidth}p{0.29\linewidth}}
\toprule
Verb & Edited clause & Content kept fixed & Condition for acceptance \\
\midrule
\endhead
Complete & Attach one of \(C,R,P\) & \(M=(\Om,\Xis)\) and every completion
clause held fixed & The added role supplies a relation that the model
explicitly lacks, without changing the consequence held fixed. \\
Reattach & Replace or attach exactly one of \(\Om,\Xis\) & The other factor
and the stated consequence & The source and target relations satisfy the
commuting condition of the interface, with a residual below the tolerance
stated for the map. \\
Compose & Concatenate two or more typed moves & The signature held fixed for
each component & Every target interface matches the next source interface,
and the composed model remains closed and computable. \\
Deform & A path in a parameter, scale, boundary or representation & A named
signature \(Q\) and a stated observable consequence & \(Q\) remains fixed
along the stated path, no singularity invalidates the map, and the closed
model remains well posed. \\
Observe & The observable map \(R\), an intervention or the protocol \(P\) &
The mechanism core and its closure & The resulting measurement separates the
candidate from a stated null model or competing mechanism under the same
conditions. \\
Revise & The clause implicated by a consequence that failed & The signature
held fixed and the unchanged clauses & The replacement reduces the same error
that was measured in the failed test and still passes the conditions of
interface and closure. \\
\bottomrule
\end{longtable}

\subsection*{Three independent transfer tests}

The transition cost in the representation (S9) ranks candidate roads but does
not establish a physical invariant. We therefore use three separate
calculations. The first tests dimensional similarity; the second and third
test dynamical quantities.

\paragraph{Dimensionless groups.}
For a dimension matrix \(D\in\mathbb Q^{k\times n}\) we compute the exact
rational null space
\begin{equation}
D\mathbf a=0,
\qquad
\Pi_{\mathbf a}=\prod_{r=1}^{n}q_r^{a_r},
\end{equation}
which gives \(n-\operatorname{rank}D\) independent groups. The elimination is
done in exact rational arithmetic, so that an integer exponent cannot be
altered by a floating-point threshold on singular values. A map of quantities
passes this test only when every group of the source maps to a dimensionless
combination of the quantities declared for the target. When magnitudes are
given, the values of the groups are compared as well, and values within 5\%
are treated as the same regime; the comparison is made on the basis of groups that the elimination returns, and a power of a group changes the relative difference of its values. This is a condition of dimensional
similarity. It is not a Noether theorem, and by itself it does not prove that
two governing equations are equivalent.

\paragraph{Transported quantities.}
For an invertible interface \(\alpha\) satisfying
\(g=\alpha f\alpha^{-1}\), a functional \(Q\) of the source is transported as
\begin{equation}
Q'=Q\circ\alpha^{-1},
\qquad Q'\circ g=Q'.
\end{equation}
The calculation handles linear charges, quadratic functionals and functionals
supplied by the user. It evaluates the trajectory of the target and reports
the relative drift for conserved quantities, or the largest change in the
forbidden direction for monotone quantities. A semiconjugacy or a projection
that cannot be inverted requires a separately specified functional of the
target and is not evaluated by this formula.

\paragraph{Action symmetry.}
For a given first-order Lagrangian and a strict internal generator that acts
on the fields at fixed coordinates, the symbolic calculation gives the
variation of the fields, the Euler--Lagrange equations and the
current. At present only strict symmetries, for which the calculated
variation vanishes, are accepted; symmetries up to a total divergence require
an explicit boundary term and are not treated by this calculation. Generators
with a nonzero coordinate part are rejected, because they require the
prolonged point transformation and its characteristic. The result is
therefore a theorem about the source equations only when the action of the
source, the generator and the boundary assumptions are all given.

In the test cases, two independent groups are recovered from five quantities
whose dimension matrix has rank three; a conserved linear functional and a
dissipative quadratic functional are transported; and the shift current of
the free scalar field is recovered, while a cubic term that breaks the
symmetry is rejected. These tests establish that the algebra of the three
calculations is correct. Physical acceptance still relies on the residuals
and controls specific to each source, described below.

\subsection*{Mechanism-preserving interface}

For the transformations \(\Phi_i\) of the source and \(\Phi_j\) of the
target, the interface map \(\alpha\) is admissible when
\begin{equation}
\alpha\circ\Phi_i=\Phi_j\circ\alpha.
\label{eq:supp-constructor-commuting}
\end{equation}
For dynamics \(\dot x=f(x,e)\) and \(\dot y=g(y,e')\), a map \(\alpha\) of
states and a map \(\alpha_E\) of contexts, the numerical check is
\begin{equation}
r_\alpha(x,e)=d\alpha_x f(x,e)
-g\!\left(\alpha(x),\alpha_E(e)\right),\qquad
\lVert r_\alpha\rVert\leq\varepsilon_\alpha .
\label{eq:supp-constructor-residual}
\end{equation}
Each construction specifies the two relations and both maps. It also assigns
to every state, context, parameter, control and observable a physical type
\begin{equation}
q=\bigl(\text{name},\text{role},\mathbf d_q,u_q,s_q\bigr),
\end{equation}
where \(\mathbf d_q\) is a sparse vector of exponents of the base dimensions,
\(u_q\) is the declared unit and \(s_q\) converts that unit to a canonical
unit. For a map of quantities \(q_i\mapsto q_j\) with numerical scale
\(a_{ij}\), the residuals are
\begin{equation}
r_{\rm dim}=\lVert\mathbf d_i-\mathbf d_j\rVert_1,
\qquad
r_{\rm unit}=\frac{|a_{ij}-s_i/s_j|}{\max(|s_i/s_j|,10^{-12})}.
\end{equation}
The check of dimensions requires \(r_{\rm dim}\leq10^{-12}\), and the check of
units uses the tolerance stated with the map. A yes-or-no declaration that
the dimensions are consistent cannot replace either calculation.

Compatibility is expressed as a set of named conditions
\((k,r_k,\epsilon_k,\triangleleft_k,\lVert\cdot\rVert_k,e_k)\), each of which
gives the type of the condition, the measured value, the tolerance, the
comparison, the norm and the source of the evidence. Complete requires a
residual of closure; Reattach and Compose require a residual of commutation;
Deform also requires the drift of the invariant; Observe requires the margin
by which the measurement separates the alternatives; and Revise requires the
error obtained when the failed test is repeated. A condition without a value
remains incomplete, and a value outside its tolerance fails. Compatibility
estimated from the corpus decides which interfaces are calculated, but it
cannot replace a residual within its tolerance.

Once the interface passes, the operational identity of the target defines a
realization fiber
\begin{equation}
\operatorname{Real}(\widehat I_{\mathrm{op}})
=\pi^{-1}(\widehat I_{\mathrm{op}})
=\{\widehat A_k:\pi(\widehat A_k)=\widehat I_{\mathrm{op}}\}.
\end{equation}
Each \(\widehat A_k\) specifies the variables, the domain, the typed
quantities, the parameter values with their units, the constitutive
relations, the boundary or initial data, the observable and the ordered
protocol. A realization is either a numerical simulation or a fabrication
plan. A numerical realization is specified by the program that computes it,
its parameter values and its expected outputs. A fabrication realization is
specified by a design, the machine to be used and the ordered steps of the
process, and every numerical parameter of the process carries a dimension, a
unit and a tolerance. The output is a written fabrication plan; no machine is
operated. Several members of the fiber may realize the same operational
identity, and agreement among them tests whether the claimed consequence
belongs to the mechanism rather than to one material or to one numerical
implementation.

\subsection*{Pairing of one-sided sectors and realization of the pair}

The exact one-sided sectors give two complementary lists of candidates. The
sector \((\Om,0_{\Xis})\) supplies apparatus detached from any carrier, and
the sector \((0_{\Om},\Xis)\) supplies carriers on which no apparatus was
detected. Pairing a family from one list with a family from the other gives a
proposed reattachment. The interface and its physical realization must then
pass nine separate checks:
\begin{longtable}{p{0.20\linewidth}p{0.70\linewidth}}
\toprule
Check & Required evidence \\
\midrule
\endhead
Type & The identities of source and target declare the operation, the carrier,
the relation and the typed quantities of the state. \\
Symbolic interface & The relations of source and target, the map of states,
the map of contexts, the map of domains and the preservation equation are
written explicitly. \\
Domain & The domain and codomain of the apparatus agree with the state carried
by the target substrate. \\
Dimension and unit & The computed residuals of dimension and of unit
conversion pass for every required map of quantities. \\
Residual & Every residual required by the selected construction verb is present
and within its stated tolerance. \\
Closure & The constitutive assumptions, constraints, boundaries and initial
conditions make the model well posed. \\
Observable & A measurement distinguishes the proposal from a simpler null model
and from its nearest control that does not pair the two families. \\
Protocol & Perturbation, relaxation, sampling and measurement are ordered,
with stopping rules and numbers of replicas. \\
Realization & The realization can be run as a simulation or written as a
fabrication plan, with stated inputs, ordered operations, parameter sweeps,
convergence criteria and outputs. \\
\bottomrule
\end{longtable}

The source examples and the assignments to sectors must come from the same
fit of the codebooks. Occupancy in the corpus is kept separate from physical
compatibility: a pair \((\Om,\Xis)\) that already occurs in the corpus is a
retrospective reattachment and not a candidate for simulation. A pair becomes
a candidate only when all nine checks pass and the catalog of physical models
supplies a program, parameter values, expected outputs, observables and
negative controls.

Before a simulation is run, its specification must contain the parameters, a
nonempty sweep, at least two replicas, convergence criteria, predictions made
before the run, observables and negative controls. A completed run does not
by itself make the mechanism accepted. Acceptance tests separately the
prediction stated before the run, the residuals and the controls, and a
physical claim further requires external or experimental evidence.

At the time of this analysis, the diffusion example of the Methods and the
Langevin example of Section~S17 are controlled reconstructions of established
transfers. They test the formal and numerical conditions, but they were not
selected by a complete automatic loop leading from the one-sided sectors to a
simulation. That missing stage is measured directly: acceptance requires an
attachment supported by the sources that passes the catalog of physical
models, can be run, is run, and passes the tests stated before the run.

Compression and extension have different mathematical forms. Compression is a
many-to-one map from the equations of the sources to a bounded state.
Extension is a set-valued search over the equations compatible with that
state. A construction is consequently assessed by whether it remains valid
under independent constraints, and not by whether it reproduces the source
expression.

\subsection*{Continuum-to-graph diffusion control}

The diffusion control used a regular \(15\times15\) grid (225 nodes and 420
undirected edges), diffusivity 0.05, final time 0.15 and 121 time points. The
construction kept the operation of diffusion and replaced the continuum
carrier by a graph Laplacian. On the interior nodes the relative error of the
operator was 0.01418. The relative drift of the total mass was
\(2.44\times10^{-15}\), the graph energy fell from 3.5650 to 0.4190, and the
state remained nonnegative within numerical tolerance. Diffusion with the
sign reversed increased the energy by a factor of 4216.4, and evolution with
the adjacency matrix alone, in place of the Laplacian, produced a relative
drift of mass of 0.9972.

\subsection*{Brownian-to-optimization control}

The Brownian control solved the one-dimensional Fokker--Planck equation on
401 grid points at temperature 0.45, starting from a density centred at
\(-1.45\) with width 0.1, and evaluated it at times 0, 0.15, 0.8 and 6. The
drift of the total probability was \(3.33\times10^{-16}\). The final density
lay at a total-variation distance of 0.06192 from
\(p_\infty\propto\exp(-L/T)\), and the free energy fell from 1.1305 to
0.00396. With the drift pointing uphill and with zero noise, the final
densities lay at distances of 0.9839 and 0.9951 from the stationary
prediction, and with the uphill drift the free energy rose to 6.3960. A
retrieval test in which the field of the target is hidden has been
specified; its result is required before this control can be described as a
rediscovery from the corpus.

\subsection*{Specifying the target carrier}

The two factors differ in stability, and this asymmetry constrains how a reattachment may be specified. Operation classes recur more consistently across fits, whereas the boundaries of carrier classes move more substantially. Fixing the atlas fixes the numerical labels of both factors, but a physical target must in addition specify its dimension, its discreteness, the type of its boundary and its symmetry, rather than rely on a class label alone. The step of the construction procedure in Methods that leaves the target domain to supply $C_t$, $R_t$ and $P_t$ is therefore required by the construction itself: those relations determine which state can evolve and what can be measured.

\section*{S11. Null models and ablations}

Each result of the analysis was compared with a control that removes or
reverses the ingredient on which the result is supposed to depend. The
controls are the following.

\begin{enumerate}
\item \emph{Control coordinates.} The shape of an equation and the quality of
its extraction are kept as negative-control coordinates. They carry zero
weight by default in the mechanism action, so that a result cannot rest on how
an equation looks or how cleanly it was read.
\item \emph{Single-factor ablations.} Codes evaluated on papers not used for
fitting test whether \(\Om\) adds information once \(\Xis\) is known, and
whether \(\Xis\) adds information once \(\Om\) is known.
\item \emph{Completion ablation.} The gain carried by \((C,P,R)\) is measured
only after the factor product is already known, so that it counts information
beyond the operation and the carrier.
\item \emph{Direction reversal.} The same edges, local to each source, are
encoded in both directions, which tests whether the stated direction of a
derivation carries information.
\item \emph{Composition.} Codes for one step and for two steps measure how much
predictive information is retained when moves are composed.
\item \emph{Composition on separate papers.} Complete papers are split before
\(\Om\), \(\Xis\) and \(T\) are fitted. The papers not used for fitting are
split again into a calibration set and a final test set. The codebooks are
fitted on the training papers and then fixed. Together with
\(P(T_2\mid\Om,\Xis,C/P/R,T_1)\), they must predict better than
\(P(T_2\mid T_1)\) on the test papers, which are not used at any earlier
stage, before the representation is called predictive. A separate yield
threshold decides whether compositions absent from the training papers are
accepted as candidates for construction.
\item \emph{Support in the sources.} A token is accepted only if it has
examples among the original equations, if independent routes agree on it and
if there is evidence for the role it is said to play.
\item \emph{Graph coverage.} The limits on the number of edges and on the
sampled incidence are reported before any pairing is said to be absent, since
an absence may otherwise reflect only incomplete sampling.
\item \emph{Displays and residuals kept separate.} Centrality in the atlas and
the residuals of auxiliary reconstructions are not used for statements about
the representation, the hierarchy or discovery unless they pass an independent
calibration against the sources.
\end{enumerate}

The table lists each statement together with the evidence for it, computed on the full-corpus representation (S1).

\begin{longtable}{p{0.30\linewidth}p{0.16\linewidth}p{0.43\linewidth}}
\toprule
Statement & Result & Evidence \\
\midrule
\endhead
Two-factor symbolic index & Supported & Every row is assigned, and both
factors give independent gains on the test papers \\
Operation--carrier division of the features & Kept for its physical meaning &
In the feature-partition control (S5), 48 of 50 random partitions of the same
features predict better; the division is kept because of its physical
interpretation, not because the data select it \\
Attached completion fiber & Supported & Positive information on the test
papers beyond the factor product, on both splits of the
representation \\
Predictive local representation & Supported & The codebooks are fitted on
complete papers and fixed before evaluation; the code length of the next move
and of the destination is reduced \\
Context-conditioned \(T_2\) prediction in unseen papers & Supported &
$8{,}383$ and $8{,}647$ test chains on two splits by paper of the
representation; the code length of the transition is reduced by $22.1\%$ and $23.1\%$ \\
Conditional module recovery & Measured on the representation & The
hidden operation, carrier and completion clauses are recovered from the
selected assignments \\
Realization of absent compositions in test papers & Measured &
The observed realization is 10.02\%, with a Wilson 95\% lower bound of 7.53\%,
above the 5\% yield threshold set in advance; the predicted mean, 6.82\%,
underestimates it, and recalibration is the next step \\
Directed construction order & Supported & The completion DAG and the edges in
the sources that complete a role \\
Operation--carrier prediction & Next test & Tests at cutoffs by publication year, and on unoccupied pairings supported by the sources \\
Radial axiomatic hierarchy & Next test & Labels of axioms and applications
supported by the sources, with controls for radial order on the test papers \\
Physical conservation law & Next step & Requires the action of the source, its
symmetry and its boundary conditions \\
New mechanism from a gap in the literature & Constructed & No compiled model
joins mass action to a periodic term; the amount--phase construction across
this gap maintains a set point and keeps the integral of its error in the phase, and it keeps the mean
concentration of a published reaction network \\
\bottomrule
\end{longtable}

\section*{S12. How the results were computed}

\subsection*{Configuration of the confirmatory model}

The confirmatory run split the corpus by paper in a deterministic way. Whole
papers were assigned to a training set (80\%), a validation set (10\%, used
for calibration and early stopping) and a test set (10\%); the two splits used
the seeds 314159 and 20260902. The model seed was 271829. The standardization
of the features and the \(\Om\), \(\Xis\) and \(\Tau\) codebooks were fitted
on the training papers only and then held fixed. Each codebook fit used a
sample of at most 500,000 training rows and 500,000 training edges, after
which all aligned rows were assigned in batches of 100,000.

\begin{longtable}{p{0.43\linewidth}p{0.43\linewidth}}
\caption{Configuration of the confirmatory run on complete papers. The values
are the settings with which the predictive model was trained.}
\label{tab:supp-confirmatory-config}\\
\toprule
Quantity & Value \\
\midrule
\endfirsthead
\toprule
Quantity & Value \\
\midrule
\endhead
Requested codebooks & \(32\ \Om,\ 18\ \Xis,\ 8\ \Tau\) \\
Paper partitions & 80\% training, 10\% validation (calibration and early stopping), 10\% test \\
Paper split seed; model seed & 314159 and 20260902; 271829 \\
Training schedule & 1,200 steps per epoch; early stopping after 3 epochs \\
                  & without improvement in calibration (selected epoch 1) \\
Node and sequence batch sizes & 4,096; 4,096 \\
Typed-edge batch size & 8,192 \\
Encoding and hidden widths & 24; 192 \\
Operator and substrate fibers & 4; 3 \\
Optimizer & AdamW \\
Learning rate; weight decay & \(8\times10^{-4}\); \(10^{-4}\) \\
Gradient clipping & 1.0 \\
Reconstruction, node and transition weights & 0.10; 1.0; 1.5 \\
Next-state and factorized-verb weights & 0.50; 1.0 \\
Transition-gradient weight & 0.50 \\
Self-correction weight and noise & 0.20; 0.01 \\
Grammar and self-correction start epochs & 1; 3 (after the selected epoch) \\
Minimum predictive gains & 0.02 bits for transition and destination \\
\bottomrule
\end{longtable}

Training ran under Python 3.9.13 and PyTorch 2.0.1 on four CUDA GPUs. The
model that generates equations is a separate model, and it is not used as
evidence for the statements about the representation made in the main paper.

\subsection*{Processing rate and computational requirements}

The processing run took approximately 15 hours on the four-GPU workstation.
The main bottleneck was the download of the source documents from Hugging
Face; during intervals limited by the download, approximately 20--30 documents
were processed per second. The total duration and these interval rates refer to
the cluster run, and the interval rates are not averages over the whole run.

The processing rate is defined as the number of documents completed since the
start or restart of processing, divided by the elapsed wall-clock time. Rates
reported before the construction of features count only the collection of the
input documents. To produce the full-corpus representation, features are
computed on the CPU from the cleaned source and target equations of
each transition, by pattern matching on the text of the equations and by
array operations. These feature vectors are the inputs to the encoders, trained
on the GPUs, that are described above.

For $N_D$ source documents processed at a mean aggregate rate $v_D$ over one
complete processing pass, the duration of the pass is
\begin{equation}
t_{\mathrm{pass}}=\frac{N_D}{v_D}.
\label{eq:supp-document-processing-time}
\end{equation}
Because the download speed varies during the pass, the rate measured over one
interval cannot replace this mean over the whole pass. The reported duration of
15 hours is the measured duration, not an extrapolation from an interval
limited by the download.

Extraction from the papers and construction of the features use deterministic
structural operations; no generative language model is run at this stage. The
representation has 95 coordinates, and the predictive
encoders use encoding and hidden widths of 24 and 192, with the training
settings given above. The whole analysis fits on a single four-GPU workstation
and does not require pretraining a large text model. A claim of a numerical
advantage in speed, energy or cost over a particular large language model would
require both systems to be evaluated on the same documents, the same extraction
task and the same output quality, with the same accounting of hardware. No such
comparative evaluation is reported here.

\subsection*{Order of the computations}

The analysis was carried out in the following order.
\begin{enumerate}
\item The source documents were processed in full, each retained pair of
equations was converted into a typed feature vector, and the complete list of
features was compiled (S2, S4).
\item The typed graph (S6), the source equation cards (S1) and the source
constructor graph (S7) were built.
\item The symbolic representation was fitted and a codeword was assigned to
every row. The complete index of codewords (S4) and the tables of source
examples for each codeword were produced, and the assignment was checked to
cover every row.
\item The structure of the representation was selected (S5). The confirmatory
predictor was then trained on the training papers and evaluated on the test
papers with the settings of Table~\ref{tab:supp-confirmatory-config}.
\item The one-sided sectors, in which either the operation or the carrier is
null, were checked and source examples were collected for them. Pairings of
the two sectors were then tested and described one by one (S10).
\item The proposed constructions were given typed identities with units,
passed through the checks of S10, ranked, and realized as simulations or as
plans for fabrication. Predictions stated before each simulation were then
compared with its outcome.
\item The completion DAG (S7), the absent operation--carrier products and the
saturation of counterfactual products, the variational and \(J\)-invariant
checks and the counterfactual recovery of clauses (S5) were computed.
\item The diffusion and Brownian controls of S10 were computed. The retrieval
test of the Brownian control with the target field hidden was set up; its
result is not reported here (S16).
\item The figures were generated.
\end{enumerate}

The atlas computed without field labels, the symbolic quotient, the
compression DAG, the figure of typed absent pairings and the roads of exact
action are all drawn from the accepted assignments. The architecture figure
shows the measured prediction, the generation of candidates and the update
after validation as separate parts. The figures draw paths as they were
computed: paths are not replaced by straight lines, components of a graph are
not joined by connections absent from the data, and only the current
assignments are used. All random sampling and all projection steps use fixed
seeds. Assignments, graphs and source records remain linked to one another by
the identifiers of rows and sources.

\section*{S13. Data and Code Availability}

The equations were taken from arXiv \LaTeX{} source documents in the two
\href{https://huggingface.co/datasets/synthetix-institute/latex-data}{latex-data}
collections on Hugging Face; the source files remain subject to the terms of
arXiv and of the dataset. Each extracted equation keeps the identifier of its
document, its position in the paper and a pointer to its location in the
source. The arrays of extracted equations, features, cluster assignments and
transition graphs are too large to distribute with the article and are not
publicly released; the counts and statistics derived from them are given in
the tables of this Supplement. Table~\ref{tab:supp-release-matrix} lists
what is available and where.

FieldBridge, which carries out the constructions and the reductions and checks
them against the original models, is public at
\href{https://github.com/synthetix-institute/fieldbridge}{github.com/synthetix-institute/fieldbridge},
with an \href{https://synthetix-institute.github.io/fieldbridge/}{interactive
demonstration} and a
\href{https://github.com/synthetix-institute/fieldbridge/blob/main/docs/tutorial/index.md}{tutorial}.
The companion projects MorphWiki and Equation X-Ray are public under the same
organization, \url{https://github.com/synthetix-institute/}. The code that
processes the corpus and trains the representation is not publicly available.
Its random steps use fixed seeds, which are listed together with the split of
papers in Table~\ref{tab:supp-confirmatory-config}.

\paragraph{Accompanying research materials.}
The related manuscript \emph{Geometry-Encoded Hidden Orientational Memory}~\cite{BaulinColloidMemory} is available as a preprint. It contains the interfacial model and the Brownian simulations that support the colloidal application, with code and numerical materials in \href{https://github.com/vbaulin/geometry-encoded-orientational-memory}{geometry-encoded-orientational-memory}. FieldBridge~\cite{FieldBridge2026} accompanies the article as an \href{https://synthetix-institute.github.io/fieldbridge/}{interactive demonstration}. The \href{https://github.com/synthetix-institute/fieldbridge/blob/main/docs/tutorial/index.md}{tutorial} supplies the equations, references to the original papers and executable examples of constructions. Section~S27 describes how one mechanism is constructed in models from different fields.

\paragraph{Statistics.}
Equations from the same paper are not independent of one another, so the
intervals for prediction are computed over papers. Proportions are given with
Wilson intervals. Differences in coding cost are computed event by event, and
the method used for the interval is stated next to each result. The stability
of the clusters across random seeds is measured by the adjusted Rand index and
by the overlap of the equations contained in the pairings that together account for half
of all equations (S18). Two-dimensional displays, overlays and subsampled
figures are not used in any statistical test. Sample sizes and exclusions are
given with each table.

\paragraph{Available materials.}

\begin{longtable}{>{\raggedright\arraybackslash}p{0.30\linewidth}>{\raggedright\arraybackslash}p{0.18\linewidth}>{\raggedright\arraybackslash}p{0.42\linewidth}}
\caption{Materials behind the results and where they can be found.}
\label{tab:supp-release-matrix}\\
\toprule
Material & Availability & Where \\
\midrule
\endfirsthead
\toprule
Material & Availability & Where \\
\midrule
\endhead
Source documents & Public & \href{https://huggingface.co/datasets/synthetix-institute/latex-data}{latex-data}
collections on Hugging Face \\
Feature definitions & In this Supplement & The 192-coordinate dictionary and
the features used (S2) \\
Extracted equations, features and cluster assignments & Not public & Counts
and statistics in S1, S4 and S18 \\
Transition graphs & Not public & Edge counts in S6 and S7; statistics of
transitions in S19 \\
Prediction of transitions & Results only & S5 and S18 \\
Constructions and their checks against the original models & Public &
FieldBridge repository and tutorial; S27 \\
Colloid simulations & Public & \href{https://doi.org/10.5281/zenodo.22544787}{doi:10.5281/zenodo.22544787}
and the \href{https://github.com/vbaulin/geometry-encoded-orientational-memory}{geometry-encoded-orientational-memory}
repository \\
Figures & Supplied & With the article \\
\bottomrule
\end{longtable}

Results that rest on the construction calculations can be repeated with
FieldBridge. Results that rest on the processed corpus are reported as counts
and statistics, because the processing code and its outputs are not public.

\section*{S14. Accuracy of equation extraction and classes of error}

The \HyperionVTwentyOneCorpusRows{} retained rows are those that pass fixed
filters for cleanliness and for evidence of their role. The size of the corpus
says nothing about the accuracy of extraction. Measuring that accuracy requires
a separate reference sample annotated by hand. Such a sample has to draw
documents before filtering, keep the numbers of documents and of local
equations as denominators, and score four decisions: the recovery of the
boundaries of an equation, the pairing of source and target, the local order,
and the assignment of the six constructor roles. Precision, recall and \(F_1\)
must be reported for each decision, with confidence intervals clustered by
document.

The classes of error are fixed before annotation. They are split display
environments, merged aligned equations, fragments of prose, chemical formulas,
logical statements, missing targets, reversed local order and incorrect
assignment of roles. The present work checks each cleaning rule and its source
examples, but it contains no independently annotated sample with these
measurements of precision and recall. Conclusions that depend on extraction
therefore remain conditional on this evaluation. The missing evaluation does
not change the row counts reported in S1; it limits what can be said about the
error rate of those rows.

\section*{S15. Baselines for the representation and for the meaning of its classes}

The main predictive comparison asks whether the factorized operational state
codes the next transition better than reduced versions of the same state. This
is an internal ablation, not a comparison with all the available
representations of equations. A complete external comparison must use the same
split of papers, the same chains of equations within each paper and the same
test events for each of the following:
\begin{enumerate}
\item lexical features and character \(n\)-gram features;
\item normalized sequences of \LaTeX{} tokens;
\item representations of mathematical expressions;
\item a scientific-language encoder given the same local context;
\item the 192-coordinate typed vector without compression into factors; and
\item the factorized operational representation used here.
\end{enumerate}
The primary quantities are the code length of the next transition, the joint
code length of the destination and the recovery of a hidden module. Runtime and
memory are secondary quantities. None of the first four external baselines has
been evaluated on the fixed task on complete papers in the present work.

The same restriction applies to the geometry of the atlas. Projections with an
ordinary autoencoder, a two-stream autoencoder, principal component analysis
of the raw coordinates and the GGAE model have been computed, but none of them
was chosen by a common predictive test on papers not used for fitting.

Assigning a meaning to each class requires a second evaluation. For every
accepted \(\Om\), \(\Xis\), \(\Lam\), \(\Tau\), \(\Gam\) and \(\Cur\)
codeword, annotators must judge whether its source equations, shown without
labels, are instances of the proposed operational description. The report for
each token must give its support, the number of independent sources, the
stability of its centroid, the agreement between annotators, the precision
relative to controls in which the tokens are shuffled, and representative cases
of failure. The complete numerical index of codewords is given in S4. A
complete list of meanings checked against the sources has not yet been
compiled, so the numerical addresses remain the names of the classes that can
be reproduced.

\section*{S16. Transfer controls and prospective tests}

The diffusion transfer from a continuum to a graph and the transfer from
Brownian motion to optimization verify the consequences of known transfers
that preserve the mechanism, and they serve as analytic controls. The
complementary test asks whether the model of the corpus finds the target field
when the vocabulary of that field is hidden. In the Brownian test, the
signature of the source is fitted without optimization papers, optimization
equations not used in the fit are ranked with fixed coordinates, and the
labels of the target are revealed only after the ranking. This test is
specified in advance and is the next step.

\section*{S17. Three independent checks of a proposed transfer}

\paragraph{Route 1, dimensional similarity: groups from homogeneity.} Every
valid equation is dimensionally homogeneous. Writing $d(s)\in\mathbb Q^{7}$
for the exponents of the base dimensions of a symbol $s$, each pair of product
terms that are equated or added contributes a linear constraint
$\sum_i a_i\,d(s_i)=\sum_j b_j\,d(s_j)$, and the arguments of transcendental
functions contribute $\sum_i a_i\,d(s_i)=0$. Homogeneity fixes the dimensions
only up to a rescaling of the base units. A small set of constants of known
dimension ($c,\hbar,k_B,G,e,\epsilon_0,\mu_0,\dots$) therefore enters the
linear system as heavily weighted rows, and the residual of each symbol shows
where the corpus is inconsistent with itself. The system is solved separately
for each paper, because a symbol can mean different things in different
papers. The diffusion equation alone then gives $D=L^{2}T^{-1}$ with zero
residual; the chain $v=x/t$, $a=v/t$, $F=ma$ gives the dimensions of force;
and conflicting constraints give a large residual rather than a confident
assignment. The fraction of equations that could be parsed is reported with
every calculation. Equations that could not be parsed contribute nothing,
rather than contributing noise.

\paragraph{Route 2, dynamical transport: invariants carried through a verified map.} If $Q$ is conserved under the
source relation $\Phi_i$ and the map $\alpha$ intertwines the two relations,
$\alpha\circ\Phi_i=\Phi_j\circ\alpha$, then $Q'=Q\circ\alpha^{-1}$ satisfies
$Q'\circ\Phi_j=Q'$ identically. The argument uses no property of the target
dynamics itself. Every \textsc{reattach} step, in which the operation or the
carrier is replaced (Section~S10), therefore reports the invariants it carries
over together with their measured drift. For the diffusion transfer from a
continuum to a graph (Table~\ref{tab:routes}), the transported mass
$\mathbf 1^{\!\top}\mathbf u$ drifts by $5.7\times10^{-16}$ in relative terms in this test; over the full run of the transfer reported in the Methods the drift is $2.44\times10^{-15}$.
The transported Dirichlet energy $\tfrac12\mathbf u^{\!\top}L_G\mathbf u$ does
not increase, and it falls to $0.706$ of its initial value over the evaluation
window ($0.118$ over the full run). The control with reversed sign violates the
energy inequality but conserves mass, whereas the control that uses only the
adjacency conserves neither. The two controls therefore fail different
transported invariants, a distinction that a similarity score cannot make.

\paragraph{Route 3, variational symmetry: symbolic Noether currents.} Where
the source supplies an action
$S[\phi]=\int\mathcal L(\phi,\partial_\mu\phi)\,d^nx$ and a one-parameter
variation $\phi\to\phi+\epsilon\eta$, the current
$j^\mu=\frac{\partial\mathcal L}{\partial(\partial_\mu\phi)}\eta
-\mathcal L\,\xi^\mu$, with $\xi^\mu=0$ for the internal symmetries treated here, is derived symbolically and its divergence
is checked on shell. For the free scalar field with shift symmetry, the
derivation gives $j=(\partial_t\phi,\,-\partial_x\phi)$ and recovers the wave
equation as the Euler--Lagrange condition. When a $\phi^{3}$ term is added, the
shift is identified as no longer a symmetry and is rejected, rather than given
a lower score. Spacetime symmetries and boundary terms are outside the present
implementation.

\begin{table}[htbp]
\centering
\caption{The three routes applied to the diffusion transfer and its controls.}
\label{tab:routes}
\begin{tabular}{lccc}
\toprule
 & Laplacian transfer & sign-reversed & adjacency-only \\
\midrule
$\pi$-groups carried (route 1) & 1/1 & 1/1 & 1/1 \\
transported mass drift (route 2) & $5.7\times10^{-16}$ & pass & \textbf{fail} \\
transported energy monotone (route 2) & pass ($\to0.71$) & \textbf{fail} & \textbf{fail} \\
continuum symmetry (route 3) & continuum only & --- & --- \\
\bottomrule
\end{tabular}
\end{table}

\paragraph{Independence.} The three routes are logically independent.
Dimensional agreement can coexist with broken transport, as in a leaking
discretization at equal Reynolds number, and transport can be conserved
without any continuous symmetry, as in a flow on a graph. The routes are
therefore evaluated separately and are never averaged into one score. The road
displacement plays no part in this decision.

\subsection*{Transfer calculations}

Further constructions test whether a transported operation keeps its dynamics and its observable consequences (Fig.~\ref{extended-fig:validation}). Diffusion carried onto a network and Brownian motion carried onto the parameters of a loss both recover known mechanisms in new state spaces.

\paragraph{Brownian motion to stochastic optimization.}
Keeping the gradient drift together with the diffusion carries Brownian motion in a potential $V(x)$ over to the continuous-time Langevin dynamics that underlies stochastic gradient Langevin sampling of a loss $L(\theta)$~\cite{Welling2011SGLD}. In units in which the mobility and Boltzmann's constant equal one, the Fokker--Planck probability flow $\partial_t p = \nabla \cdot (p \nabla L) + T \nabla^2 p$ has the stationary law $p_\infty \propto e^{-L/T}$, provided the density is normalizable and no probability flows through the boundaries. On a one-dimensional double-well loss, the numerically evolved probability approaches this law to a total-variation distance of $0.062$, while the relative free energy decays monotonically from $1.13$ to $0.004$ (Fig.~\ref{extended-fig:validation}c).

\section*{S18. Occupancy, concentration and reproducibility of the representation on the full corpus}

\subsection*{Occupancy, completion and recovery in detail}

Main Fig.~\ref{fig:funnel}a,b shows the core that carries half of the
equation states. The measurements
below resolve, on the same publication atlas, how the cells of the
representation are occupied, how often the further clauses of a mechanism are
stated, and how well a masked clause can be recovered.

\begin{figure}[!htbp]
\centering
\includegraphics[width=0.82\linewidth]{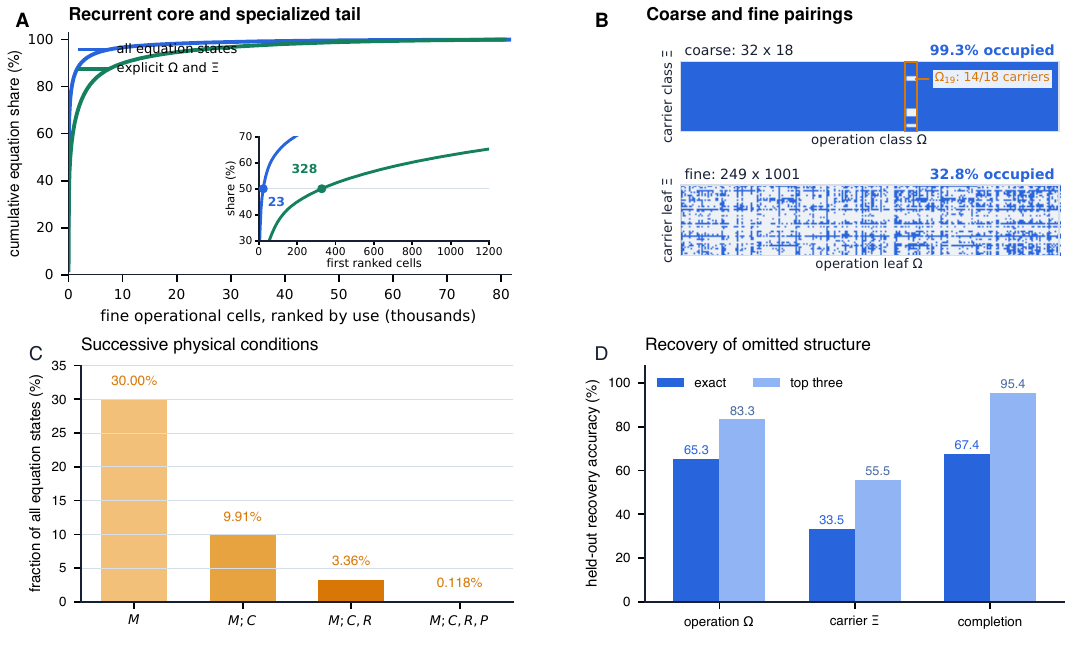}
\vspace{1.5mm}
\includegraphics[width=0.82\linewidth]{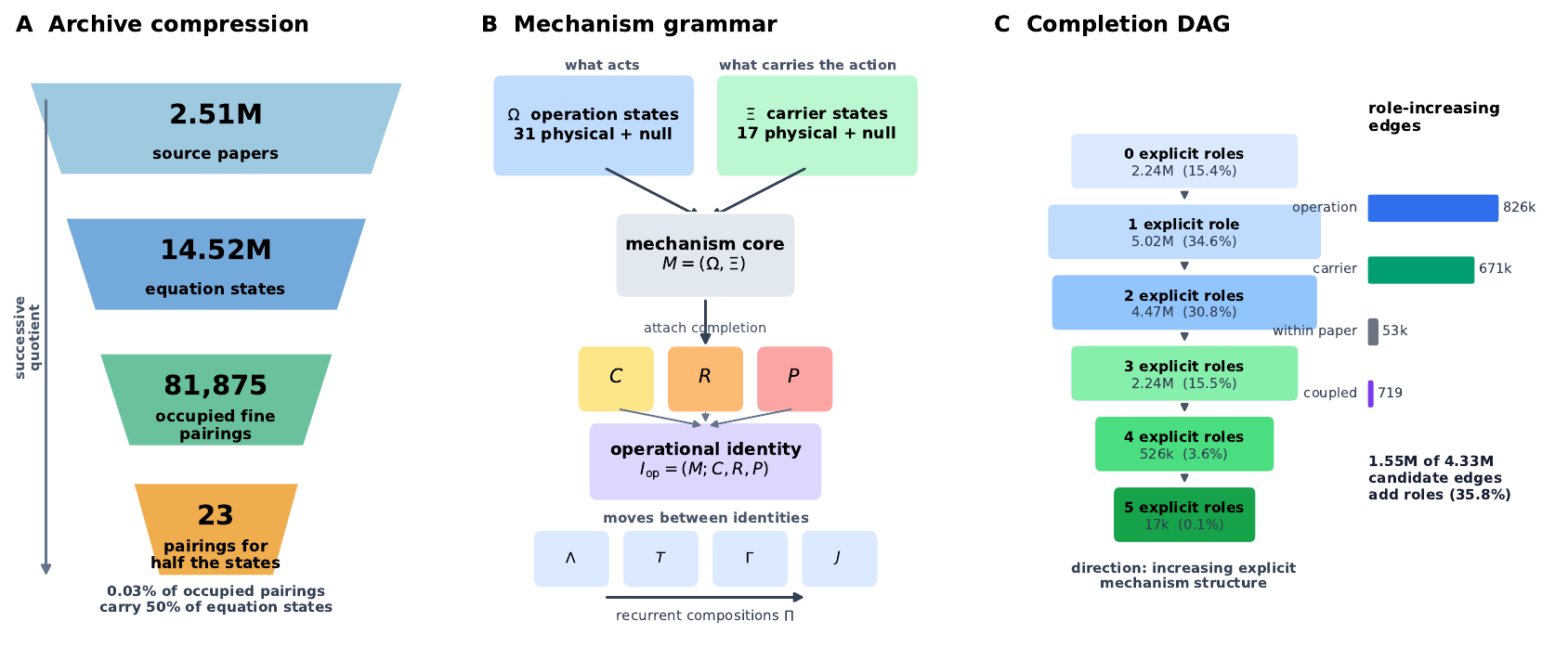}
\caption{\textbf{Compression, occupancy and completion in the nested representation.}
\textbf{Upper:} (\textbf{A})~Cumulative coverage against cell rank on a linear axis, for all equation states (blue) and for the states in which both the operation and the carrier are explicit (green). The inset resolves the two cores that carry half of each population: \HyperionCellsForHalf{} and \HyperionVTwentyOneMechanismCellsForHalf{} fine cells, respectively. (\textbf{B})~Occupancy of the coarse product of populated classes and of the fine product of populated leaves, both computed from the same fixed assignments; columns are operation classes or leaves and rows are carrier classes or leaves. The coarse product is \HyperionCoarseOccupied{} occupied, whereas \HyperionFineEmpty{} of the fine pairings of leaves remain empty. (\textbf{C})~Fraction of equation states that keep a populated mechanism core as the closure, the observable and the order of operations are stated in turn. (\textbf{D})~Accuracy with which a masked operation, carrier or completion clause is recovered in test papers.
\textbf{Lower:} (\textbf{A})~Compression from source papers to equation states, to occupied fine pairings and to the core that carries half of the corpus. (\textbf{B})~The factorized representation joins $\Omega$ and $\Xi$ into $M=(\Omega,\Xi)$, attaches $C,R,P$ to it, and records the transformations between equations. (\textbf{C})~Occupancy of the six completion layers over the full corpus, with the directed edges along which the explicit mechanism structure increases.}
\label{fig:supp-grammar-detail}
\end{figure}

The publication atlas is built from the full-corpus representation. It contains \HyperionVTwentyOneCorpusRows{}
equation states taken from 2{,}511{,}195 arXiv source documents. Each equation
is described by 44 coordinates of its operation and 28 of its carrier,
together with 23 control variables that record how it was extracted and how
it is written. The selected representation has
\HyperionVTwentyOneOmegaNonNullStates{} populated operation classes and
\HyperionVTwentyOneXiNonNullStates{} populated carrier classes, and each of the
two factors has in addition one null class. Fitting the classes in nested form
then gives \HyperionVTwentyOneOmegaLeaves{} operation leaves and
\HyperionVTwentyOneXiLeaves{} carrier leaves.

The coarse product of populated classes is
\HyperionVTwentyOneCoarseOccupied{} occupied, whereas
\HyperionVTwentyOneFineEmpty{} of the fine product is empty. Among the
$249\times1{,}001=249{,}249$ combinations of populated leaves, the fine atlas
has \HyperionVTwentyOneOccupiedFine{} occupied cells. When these cells are
ranked by the number of equation states they contain,
\HyperionVTwentyOneCellsForHalf{} of them
(\HyperionVTwentyOneCellsForHalfFraction{} of the occupied cells) already
contain half of all equation states.

Five independent fits of the representation need 23, 25, 23, 22 and
21 cells, respectively, to reach half of all equation states. For every pair
of fits with different seeds, the sets of rows in these cores overlap with a
Jaccard index of at least \HyperionVTwentyOneHalfCoreJaccard{}. For the
population in which both factors are explicit, the corresponding counts are
328, 336, 348, 310 and 317. We had fixed in advance a tolerance of 10\% on the
pairwise variation of these counts, and the fits exceed it: the largest
relative differences are 16.0\% for all rows and 10.9\% for the explicit
pairs. We therefore report the atlas as a measured partition together with
its range across seeds, and not as an invariant number of mechanisms. The rows
that make up half of the population overlap strongly between fits, even
though the number of cells needed to cover them varies.

The carrier partition changes more from fit to fit than the operation
partition. We compare partitions on the equation rows that two fits share by
the adjusted Rand index (ARI), which equals one for partitions that are
identical up to relabelling. It ranges from 0.8091 to 0.9456 for carriers,
from 0.9976 to 0.9987 for operations, and from 0.9797 to 0.9970 for the fine
pairings. In eight of the ten pairs of seeds the carrier ARI falls below the
minimum of 0.90 that was set in advance. Only the pair of fits with seeds
271829 and 271843 meets all the criteria on coordinates and on coverage. A
small core thus recurs across fits, while the carrier assigned to an
individual equation remains uncertain.

The distribution over all rows includes equation states in which only one of
the two factors is explicit. When both an operation and a carrier are
required, \HyperionVTwentyOneMechanismRows{} states remain, which is
\HyperionVTwentyOneMechanismFraction{} of all equation states. These states
occupy \HyperionVTwentyOneMechanismOccupiedFine{} fine cells. In the selected
atlas, \HyperionVTwentyOneMechanismCellsForHalf{} cells contain half of this
population; the independent fits need
\HyperionVTwentyOneMechanismCellsForHalfMinimum--\HyperionVTwentyOneMechanismCellsForHalfMaximum{}
cells, and the sets of rows overlap with a Jaccard index of at least
\HyperionVTwentyOneMechanismHalfCoreJaccard{}.

For this population with both factors explicit, the product of populated
leaves contains $248\times996=247{,}008$ pairings, of which
\HyperionVTwentyOneMechanismFineEmpty{} are empty. Leaving out the rows with
an unspecified factor therefore changes the empty fraction only from
\HyperionVTwentyOneFineEmpty{} to \HyperionVTwentyOneMechanismFineEmpty{},
while the number of cells that contain half of the population changes from
\HyperionVTwentyOneCellsForHalf{} to \HyperionVTwentyOneMechanismCellsForHalf{}.
In the selected atlas the overall sparsity is thus insensitive to this
exclusion. Whether an individual empty pairing stays empty in other fits, and
whether it is physically admissible at all, can only be decided by comparing
the equations that make up its two factors and the domains of their
operators.

A separate comparison at fixed rank keeps the 540 most populated cells of
each fit; these contain 78.7--79.4\% of all equation states. The
number 540 was chosen in advance, whereas the counts of cells containing half of the
population, given above, are determined separately from each distribution.

Saturation of the coarse product and concentration in the fine product
describe two levels of the same hierarchy. A coarse cell records a broad
combination of an operation and a carrier that the literature has already
used. A fine leaf keeps the detailed form of the operator and the geometry or
algebra of the domain on which it acts. In this description, written theory
reuses a small operational core and distributes specialized constructions
over a long tail. The census also recomputes this rank after removing the
rows in which the operation or the carrier is explicitly null; these
conditioned values are taken from the same assignments of the
representation.

\subsection*{Prediction of the next transition in test papers}

The confirmatory predictor of Section~S12 was evaluated on derivation chains
taken from test papers that had been set aside before any codebook was
fitted. A chain is a run of consecutive equations, linked to their source,
within one document, and one evaluation event is one transition inside such a
chain. The scores describe the test papers that contain such chains, $2{,}479$ and $2{,}500$ in the two divisions, and not the whole of the test set. We report two independent divisions of the papers, with seeds 314159
and 20260902; each assigns whole papers at random, irrespective of their year and of the models they state, so that a model of a test paper can also occur in training papers, and each refits the codebooks on its own training papers. The
probability baselines are the training frequencies conditioned on the
preceding transition. For the destination operation and carrier these are
smoothed categorical frequencies, and for the three completion indicators
they are separate Bernoulli frequencies. The most probable class or indicator
value under these frequencies gives the baseline for destination accuracy.
For the skill of predicting a factorized change, the baseline is instead a
single majority class for each change variable, fitted on the training
papers. All baselines were fixed before the evaluation.

\begin{table}[htbp]
\centering
\caption{Prediction of the next transition on two independent divisions of
the papers into training and test sets. The destination baselines use
probabilities conditioned on the preceding transition; the change skills use
majority-class baselines. All baselines are fitted on the training papers.
Gains in code length are given as fractions of the mean code length of the
baseline, because the vocabulary of transitions is refitted for each split; the minimum gain of 0.02 bits of Section~S12 refers to the absolute gains, which are not listed here.}
\begin{tabular}{lrrrr}
\toprule
 & \multicolumn{2}{c}{Split A (314159)} & \multicolumn{2}{c}{Split B (20260902)} \\
\cmidrule(lr){2-3}\cmidrule(lr){4-5}
Quantity & Model & Baseline & Model & Baseline \\
\midrule
Test papers & 2{,}479 & & 2{,}500 & \\
Derivation chains & 8{,}383 & & 8{,}647 & \\
\midrule
Class of the next transition & 70.3\% & & 60.1\% & \\
Destination operation \(\Om_{t+1}\) & 71.0\% & 72.1\% & 68.8\% & 69.8\% \\
Destination state space \(\Xis_{t+1}\) & 44.0\% & 42.0\% & 43.1\% & 41.8\% \\
Destination completion \(F_{t+1}\) & 59.3\% & 59.7\% & 56.2\% & 57.1\% \\
\midrule
Gain in code length, transition & 22.1\% & & 23.1\% & \\
Gain in code length, joint destination & 7.9\% & & 7.1\% & \\
\midrule
Skill, change of operator & 0.298 & & 0.293 & \\
Skill, change of carrier & 0.313 & & 0.304 & \\
Skill, change of completion & 0.292 & & 0.308 & \\
\bottomrule
\end{tabular}
\label{tab:supp-next-transition}
\end{table}

In both splits the model predicts the most probable destination carrier more
often than its conditional-frequency baseline, but it falls marginally below
the baseline for the destination operation and completion. The corresponding
gains in code length remain positive, and their confidence intervals, which
take the clustering of events within papers into account, exclude zero. Code
length measures the probability that the model assigns to the destination
actually observed, whereas top-one accuracy counts only whether the most
probable destination is the correct one. The model improves the first
quantity without improving the second in every case.

The three factorized changes are predicted with similar normalized skill.
Skill is defined as
\((\text{accuracy}-\text{baseline})/(1-\text{baseline})\), the share of the
room above the baseline that the model captures. All six values lie between
0.292 and 0.313, and their order changes between the splits. If one instead
imposed the same accuracy threshold on every change, the change of carrier
would appear weak, but only because its vocabulary is more evenly balanced:
the change of operator puts 56\% of its mass in a single class, against 25\%
for the change of carrier, so a fixed threshold demands far less of the
former.

In both splits the destination operation is predicted more accurately than
the class of the transition. The class of a transition ranges over the
directed changes between consecutive equations of one source, so it
distinguishes cases that lead to the same destination; a correct destination
can therefore go together with an incorrect class. The destination carrier
and completion do not have this property, and in both splits they are
predicted less accurately than the class of the transition.

The learned representation predicts the transitions between equations that
are recorded in the papers. The physical reductions of Section~S24 rest on a
different basis: they follow from explicitly specified generators and
observables, and the missing concentration coordinate is obtained by closing
the measured variables under the allowed dynamics. These calculations
establish a correction of the model independently of the result on learned
continuation. To measure whether the learned representation helps to
construct such corrections, it must be compared with representations built
from the equations alone and from the text, using the same source equations,
physical conditions and computational tools. That comparison is the next
step.

\subsection*{What the recurrent core contains}

The \HyperionVTwentyOneCellsForHalfMinimum--\HyperionVTwentyOneCellsForHalfMaximum{} cells that carry half of the archive are counted before both factors are required to be explicit. A second census restricts the ranking to the \HyperionVTwentyOneMechanismRows{} rows that have both an operation and a carrier; across the independent fits, the core containing half of these rows contains \HyperionVTwentyOneMechanismCellsForHalfMinimum--\HyperionVTwentyOneMechanismCellsForHalfMaximum{} cells. The nested sectors then add the closure, the observable and the order of operations. The three levels separate equation fragments that recur, cores of explicit mechanisms, and complete physical constructions. The conditioned row counts, the occupied products and the ranks that contain half of each population are all computed from the same assignments of the representation.

\subsection*{Concentration and reproducibility of the two factors}

The analysis covers \HyperionCorpusRows{} equation states linked to their sources in 2{,}511{,}195 arXiv source documents, and resolves each equation into its operation, carrier, closure, observable and ordered operations (Supplementary Section~S2). Many equations map onto few cells because notation is shared: most equations carry one of a small number of combinations of notation cues (Overview). This many-to-one organization does not select the way the features are divided into an operation block and a carrier block (Section~S5). Across five independent fits, \HyperionVTwentyOneCellsForHalfMinimum--\HyperionVTwentyOneCellsForHalfMaximum{} fine cells contain half of all equation states. When both an explicit operation and an explicit carrier are required, \HyperionVTwentyOneMechanismRows{} states remain, and the core that contains half of them spans \HyperionVTwentyOneMechanismCellsForHalfMinimum--\HyperionVTwentyOneMechanismCellsForHalfMaximum{} cells (Supplementary Section~S18). Carrying diffusion from the continuum onto a graph, and carrying Langevin evolution of a probability density onto a new setting, recover known physical consequences.

The operation and carrier factors behave differently under refitting. Across the same five fits, the operation partition is recovered with an adjusted Rand index of at least $0.9976$, whereas the carrier partition goes down to $0.8091$, below the minimum of 0.90 set in advance in eight of the ten pairs of fits. For the joint fine partition the minimum is $0.9797$. Operation classes are thus more reproducible under these fits, while neighbouring carrier classes share boundaries that move when the representation is refitted. The cores that contain half of the equations nevertheless keep largely overlapping sets of rows despite these shifts (Supplementary Section~S18).

Saturation and concentration describe how the equations are distributed inside this quotient. Broad combinations of operation and carrier occupy \HyperionCoarseOccupied{} of the coarse product. Fine forms remain sparse, and they are used unequally. Half of all equation states occupy \HyperionVTwentyOneCellsForHalfMinimum--\HyperionVTwentyOneCellsForHalfMaximum{} fine cells; once both operation and carrier are required, half of the \HyperionVTwentyOneMechanismRows{} explicit mechanism states occupy \HyperionVTwentyOneMechanismCellsForHalfMinimum--\HyperionVTwentyOneMechanismCellsForHalfMaximum{} cells. A small set of notation cues therefore covers most equations, and an extensive tail of rarer cues follows; statistics of mechanisms are computed on closed models (main text).

\subsection*{Structural abstraction compared with physical coarse-graining}

In statistical physics, the renormalization group (RG) integrates out microscopic fluctuations and rescales lengths and fields. Near a continuous transition, systems that flow to the same fixed point share their critical scaling~\cite{Wilson1971}. The operational quotient $Q$ plays an analogous structural role across the mathematical literature, but it does something different: it removes the clauses of realization and completion to expose the shared operational core $M = (\Omega, \Xi)$, rather than integrating out dynamical modes over physical scales. Descriptions that look unrelated can lie in the same fibre $Q^{-1}(M)$ and still differ in their coefficients, boundaries and observables. The quotient thus separates a recurring mathematical form from the material predictions made with it. To identify a core of the atlas with an RG fixed point one would in addition need a physical flow that depends on scale (Supplementary Sections~S10 and S18).

\section*{S19. Persistence of mechanism clauses along transformations between equations}

Let $F$ be a family of transformations $i\to j$ between equation states. We
call the clauses that such a family preserves more often than expected under
permutation its operational stabilizer in the corpus; it is
\begin{equation}
\Sigma_F=\left\{b:\Pr[I_b(j)=I_b(i)\mid i\!\to\!j\in F]
>\Pr_{\mathrm{perm}}[I_b(j)=I_b(i)]\right\},
\end{equation}
with $b\in\{\Omega,\Xi,C,R,P\}$. In the permutation, complete papers were
exchanged as blocks within strata of papers of similar size.
\ifHyperionVTwentyOneRoadMetrics
The calculation on the representation uses
\HyperionVTwentyOneRoadEdges{} sampled transitions between equations from
\HyperionVTwentyOneRoadPapers{} source papers. Consecutive equations keep the
same operation in \HyperionVTwentyOneOmegaRoadRetention{} of the steps and
the same carrier in \HyperionVTwentyOneXiRoadRetention{}, compared with
\HyperionVTwentyOneOmegaRoadNull{} and \HyperionVTwentyOneXiRoadNull{} when
papers are permuted as blocks.
\fi

\begin{figure}[htbp]
\centering
\includegraphics[width=\linewidth]{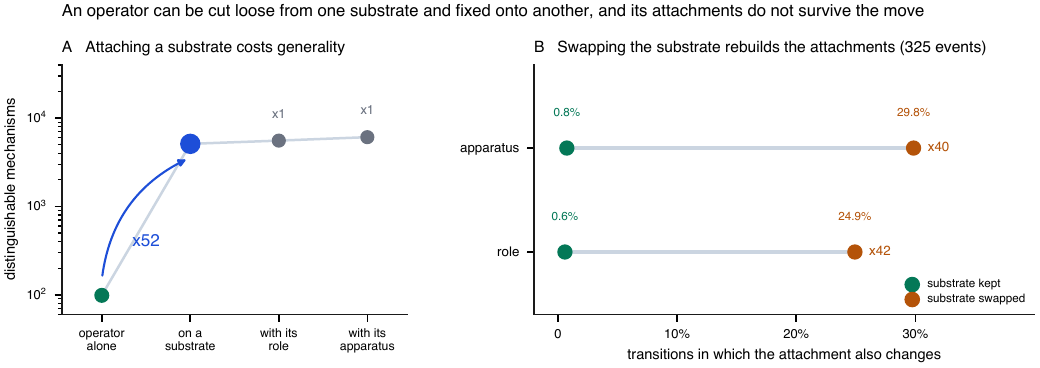}
\caption{\textbf{Specificity and multiple realizability.} Effective diversity
of equations before and after a carrier is attached, and the measured
reconstruction of completion clauses after the carrier is replaced.}
\end{figure}

\subsection*{What consecutive source equations preserve}

\ifHyperionVTwentyOneRoadMetrics
Consecutive source equations keep the same operation in \HyperionVTwentyOneOmegaRoadRetention{} of the sampled steps, compared with \HyperionVTwentyOneOmegaRoadNull{} when papers are permuted as blocks. They keep the same carrier in \HyperionVTwentyOneXiRoadRetention{}, compared with \HyperionVTwentyOneXiRoadNull{} under the same null model (Supplementary Section~S19). Consecutive equations therefore usually keep their notation cues, as one expects when each equation is derived from the one before it.
\fi

\section*{S20. Numerical tests of local compatibility and path dependence}

For each map $f:A\rightarrow B$ we computed the remainder
$\Delta_f=\Omega_B\alpha_f-\beta_f\Omega_A$. Maps that can be composed were
tested with the chain identity of Section~S21. For two paths with common
endpoints we compared both the accumulated remainders and the final maps of
states and of outputs. In the constructed square every edge was compatible on
its own and the remainder of the composite was $2.9\times10^{-14}$, while the
two maps between the endpoints differed by a norm of 8.44. The identity for
the remainders is algebraic, so subtracting its two sides measures
floating-point error and not holonomy. In an independent square, the
difference of the state maps has norm 7.206 while the remainder of the
identity is $2.66\times10^{-15}$. Moreover, when an output map is constructed
so that it satisfies the compatibility condition, exact local compatibility
follows from the construction itself. Physical path dependence is measured by
the actual difference between the path maps and by its action on a specified
state and observable. Sections~S23 and S24 use this distinction.

For the spatially constant $SU(2)$ connections used in the calculation, the
curvature reduces to the matrix commutator, and the covariant Bianchi
identity reduces to the Jacobi identity. The holonomy of a loop scaled with
the loop width with a fitted exponent of 1.999998, equivalent to an exponent
of 0.999999 against the enclosed area; when these connections were replaced
by constant commuting matrices, the residual stayed below
$3.2\times10^{-16}$. This control tests the matrix contribution to the
curvature; for a connection that varies in space, spatial derivatives
contribute as well (Section~S21). When diffusion on a ring was restricted to
a subdomain, the remainder was localized on the two cut edges, and including
it restored mass balance to $4.4\times10^{-16}$. Eliminating one coordinate
from coupled linear dynamics produced an exponential memory kernel. Including
this kernel reduced the error of the visible trajectory from
$3.22\times10^{-2}$ to $6.5\times10^{-18}$; a visible sector that was closed
under the dynamics produced no memory term.

\subsection*{Local composition and the physical consequences calculated from it}

The numerical constructions separate local from global compatibility. In a constructed square of four states, each edge was chosen so that it intertwines the local dynamics, and the algebraic identity for composition was satisfied to $2.9\times10^{-14}$. The two routes around the square nevertheless differed by a map of norm $8.44$ (Supplementary Section~S20).

Each physical representative in Eq.~\ref{eq:obstruction-representations} can be calculated independently. Spatially constant non-Abelian connections generated a curvature that satisfies the Bianchi identity, which for these matrices reduces to the Jacobi identity. The holonomy of a small loop scaled with its width with exponent $1.999998$, equivalent to $0.999999$ against its area; constant commuting connections left a residual below $3.2\times10^{-16}$. Restricting diffusion on a ring localized the missing flux on the two cut edges and restored mass balance to $4.4\times10^{-16}$. Eliminating one variable from a coupled linear evolution produced an exponential memory kernel; in the specified preparation, adding the kernel reduced the error of the visible trajectory from $3.22\times10^{-2}$ to $6.5\times10^{-18}$, whereas a projection that was closed under the dynamics produced no memory (Supplementary Section~S20; derivations in S21 and S23).

\section*{S21. Composition of local remainders and global path dependence}

Consider linear operators and linear maps on their stated domains, and
transformations $1\to2\to3$ that can be composed. Let $\Delta_{12}$ and
$\Delta_{23}$ be the two local compatibility remainders, $\alpha_1$ the first
map of states and $\beta_2$ the second map of outputs. The remainder of the
complete transformation obeys
\begin{equation}
\Delta_{13}=\Delta_{23}\alpha_1+\beta_2\Delta_{12}.
\end{equation}
A mismatch at one step is thus carried into the steps that follow. For two
paths $p$ and $q$ with common endpoints, define
\begin{equation}
H_X(p,q)=\alpha_p-\alpha_q,\qquad
H_Y(p,q)=\beta_p-\beta_q.
\end{equation}
The differences between the paths and the accumulated local remainders then
satisfy
\begin{equation}
\Omega_tH_X-H_Y\Omega_s=\Delta_p-\Delta_q.
\end{equation}
Exact local transport, $\Delta_p=\Delta_q=0$, therefore does not force
$H_X=H_Y=0$: the identity of a mechanism at each step and the independence of
the global result from the path are separate physical conditions. Writing
$H=(H_X,H_Y)$, the three possible cases are
\begin{equation}
\begin{array}{lll}
\Delta=0,\ H=0 &\Longrightarrow& \text{coherent transport},\\
\Delta\ne0 &\Longrightarrow& \text{a local compatibility remainder},\\
\Delta=0,\ H\ne0 &\Longrightarrow& \text{exact local transport with global path dependence}.
\end{array}
\end{equation}

For infinitesimal transformations generated by $X$ and $Y$,
\begin{equation}
e^{\epsilon X}e^{\epsilon Y}-e^{\epsilon Y}e^{\epsilon X}
=\epsilon^2[X,Y]+O(\epsilon^3),
\end{equation}
so the commutator gives the leading difference between the two paths.
Noncommutation is nevertheless not the general condition for path
dependence: Abelian frustrated loops, flat connections with a global
holonomy, and projected linear dynamics can also retain information about
the path.

Let tildes denote the enlarged operations and maps obtained after the missing
physical coordinate has been introduced, and let $\widetilde H_X$ and
$\widetilde H_Y$ be their path differences. Compatible enlarged maps obey
\begin{equation}
\widetilde\Omega_t\widetilde\alpha
-\widetilde\beta\widetilde\Omega_s=0,
\qquad
\widetilde\Omega_t\widetilde H_X-\widetilde H_Y\widetilde\Omega_s=0.
\end{equation}
A connection fixes how states are transported along a specified path, and
its curvature can remain nonzero in a complete theory. An additional internal
coordinate or a memory kernel plays a different role: it restores the
predictive information lost when the state was reduced. Compatibility of the
enlarged dynamics and independence of the path are thus distinct
requirements. The second identity displayed above relates the two path maps;
it does not require them to coincide.

\paragraph{Gauge curvature and transport around a finite loop.}
For a Hermitian connection $\mathcal A_\mu$ with the dimensions of momentum,
define $D_\mu=\partial_\mu-i\mathcal A_\mu/\hbar$. In a coordinate basis the
curvature $\mathcal F_{\mu\nu}$ is fixed by
\begin{equation}
 [D_\mu,D_\nu]=-\frac{i}{\hbar}\mathcal F_{\mu\nu},\qquad
 \mathcal F_{\mu\nu}=\partial_\mu\mathcal A_\nu
 -\partial_\nu\mathcal A_\mu
 -\frac{i}{\hbar}[\mathcal A_\mu,\mathcal A_\nu].
\label{eq:supp-gauge-curvature}
\end{equation}
Equation~\ref{eq:supp-gauge-curvature} contains both the variation of the
connection in space and the noncommutation of its matrices. In an Abelian
example, $\mathcal A_x=0$ and $\mathcal A_y=\mathcal B x$ have commuting
components but a nonzero curvature $\mathcal F_{xy}=\mathcal B$. A square
loop of side $a$, traversed with positive orientation in the $xy$ plane, has
\begin{equation}
 \mathfrak H=\mathcal P\exp\left(\frac{i}{\hbar}
 \oint\mathcal A_\mu\,dx^\mu\right)
 =\mathbb I+\frac{i}{\hbar}\mathcal F_{xy}a^2+O(a^3).
\label{eq:supp-gauge-small-loop}
\end{equation}
Here $\mathcal P$ orders the successive transport operations, with later
operations acting on the left, and the connection is smooth on the enclosed
patch. Under a local unitary change of basis $U$, the curvature transforms as
$\mathcal F'_{\mu\nu}=U\mathcal F_{\mu\nu}U^{-1}$, and a loop based at a
point transforms as $\mathfrak H'=U_0\mathfrak H U_0^{-1}$, where $U_0$ is
evaluated at the base point. The holonomy matrix is covariant, and its trace,
the Wilson loop, is invariant. How the field itself evolves is determined by
a gauge-field action, the coupling to matter and the sources; the connections
specified here determine only how states are transported in that field. The
relations for curvature and covariance follow Yang and Mills (1954), cited in
the main text.

The local expansion in Eq.~\ref{eq:supp-gauge-small-loop} requires a loop
that can be contracted to a point. On a ring of circumference $L$, the
constant Abelian connection $\mathcal A_s=\hbar\phi/L$ has zero local
curvature and holonomy $e^{i\phi}$. Removing it by a change of phase keeps
the wavefunctions periodic only for $\phi\in2\pi\mathbb Z$. For any other
value the phase remains in the boundary condition and changes the spectrum.
This is the Aharonov--Bohm distinction between the local field strength and
the enclosed flux. Gauge-equivalent descriptions preserve both the local
observations and this loop phase, so a nonzero holonomy is consistent with a
complete gauge theory.

\subsection*{Spatial curvature and gauge-invariant transport}

\paragraph{Gauge redundancy, curvature and holonomy along a path.}
In standard field theory, potentials and matter fields related by a gauge transformation describe the same physical predictions. A connection specifies how internal states at neighbouring points are compared, and its curvature determines the leading change of a state transported around a small contractible loop. A nontrivial holonomy can therefore be a prediction of a gauge theory that is already complete. The field equations, the coupling to matter and the sources determine how the connection evolves; the transport calculation determines the response in a given connection.

Gauge curvature includes the variation of the connection in space as well as the noncommutation of its matrices: $\mathcal F_{\mu\nu}=\partial_\mu\mathcal A_\nu-\partial_\nu\mathcal A_\mu-i[\mathcal A_\mu,\mathcal A_\nu]/\hbar$. Commuting components can therefore produce curvature when they vary in space. Conversely, a finite loop can enclose an inaccessible flux and acquire an Aharonov--Bohm phase even where the local curvature vanishes all along the path~\cite{AharonovBohm1959}. Under a change of internal basis the holonomy matrix is conjugated, and its trace and eigenvalues are invariant. These distinctions fix what a response around a loop means physically (Supplementary Section~S21).

\subsection*{Physical distinctions among the effects of composition}

\textbf{Gauge curvature} records how an internal state changes around a small loop; \textbf{geometric frustration} records local preferences of the interactions that cannot all be satisfied at once; \textbf{Mori--Zwanzig memory} records hidden motion that later acts on the retained variables. What the three have in common is that each supplies physical information needed to compose local operations, and each is tested in its own way in Eq.~\ref{eq:obstruction-representations}. They differ in what remains once the description is complete: curvature may stay nonzero after the state and its transport are fully specified, a frustrated system may have a definite equilibrium, and a memory term in a reduced description may be replaced by explicit hidden coordinates. These differences determine what each extension adds to a prediction (Supplementary Sections~S20, S21 and S24).

\section*{S22. Physical meaning of operation--carrier pairings that do not occur}

The cells of the atlas are combinations of notation cues, and some are empty by construction of the codebook (Overview). An empty pairing of an operation and a carrier is read here for its physical meaning, as a guide to construction.
Combinations of operations that are read from the mathematics of closed models,
including common ones that are never made, are reported in the main text
(Fig.~\ref{fig:atlas}c).

\subsection*{What an empty pairing means physically}
In the atlas, \HyperionVTwentyOneFineEmpty{} of the fine operation--carrier
product is unoccupied in the archive. Each empty cell is a detailed pairing of
an operation with a carrier that does not occur in the extracted literature. To
realize such a pairing physically, one needs an operator defined on an
admissible domain of the carrier, a closure of the resulting equations and a
consequence that can be measured. Whether a cell is occupied or empty does not
by itself establish that such a construction exists.

Engineered interactions can change the relations that a material supports.
Table~\ref{tab:retrospective-voids} lists four established examples: active
elastic moduli, localization at the boundary of non-Hermitian lattices,
synthetic magnetic fields and odd viscosity. In each case a specific physical
mechanism was identified, and its consequence was measured or derived. The
examples show what it takes to construct an operation on a carrier that did not
previously support it. 

\begin{table}[htbp]
\centering
\footnotesize
\setlength{\tabcolsep}{3.5pt}
\caption{\textbf{Established examples of engineered relations between an operator and a carrier.} For each construction, the physical assumption that was changed and the response that resulted.}
\label{tab:retrospective-voids}
\begin{tabular}{p{0.18\linewidth}p{0.22\linewidth}p{0.26\linewidth}p{0.28\linewidth}}
\toprule
\textbf{Physical Discovery} & \textbf{Realized Pairing $(\Omega, \Xi)$} & \textbf{Changed Physical Assumption} & \textbf{Mechanism \& Key Reference} \\
\midrule
\textbf{Odd Elasticity} & A nonzero part of $K_{ijkl}-K_{klij}$ relating stress to strain & Elastic forces need not derive from a strain-energy potential when the microscopic elements are actively driven & Active bonds that are not conservative perform work over a quasistatic strain cycle (Scheibner \emph{et al.}, \emph{Nat. Phys.} \textbf{16}, 475--480, 2020). \\
\addlinespace
\textbf{Non-Hermitian Wave Topology} & A non-Hermitian lattice generator whose eigenmodes are sensitive to the boundary & Bulk modes with open boundaries need not be described by real Bloch wavevectors & The non-Hermitian skin effect requires a generalized Brillouin zone and changes the bulk--boundary correspondence (Yao \& Wang, \emph{Phys. Rev. Lett.} \textbf{121}, 086803, 2018). \\
\addlinespace
\textbf{Synthetic Gauge Fields} & An effective vector potential acting on the motion of neutral atoms & Internal states dressed by laser light supply a momentum coupling usually associated with electric charge & An optically synthesized magnetic field produces vortices in a neutral Bose condensate (Lin \emph{et al.}, \emph{Nature} \textbf{462}, 628--632, 2009). \\
\addlinespace
\textbf{Odd Viscosity in Active Fluids} & A viscosity that is odd under exchange of the strain-rate and stress index pairs & Broken parity and time-reversal symmetry permit an additional transverse response & Spinning colloidal magnets produce chiral surface flows, whose spectra support a Hall-viscosity contribution alongside dissipative edge pumping (Soni \emph{et al.}, \emph{Nat. Phys.} \textbf{15}, 1188--1194, 2019). \\
\bottomrule
\end{tabular}
\end{table}

\subsection*{Three kinds of absence}

Three different kinds of absence appear in this work, and they cannot be
interchanged. An empty fine cell of the atlas records a pair of an operation
leaf and a carrier leaf that never occur together. To decide whether such a
pair is admissible one needs the domain of the operator and a physical
closure; the absence itself does not tell what fraction of these pairs is
inadmissible. A typed product void records, for each row, which of the
required roles are missing, and it separates out the case in which both
factors have evidence but no coupled edge joins them. Separate evidence for the
two factors does not establish that they are compatible with each other. A
composition void records an ordered pair of transitions that is never taken
inside an occupied mechanism context. It is therefore a statement about the
sequence of steps in a derivation, not about a mechanism that has not been
realized. The realization rate reported in Section~S11 is a property of the
third kind only and does not carry over to the first two.

\section*{S23. Hidden motion, dynamical closure and stored order}

Eliminating coordinates changes the information needed to predict a motion.
Let $x\in\mathbb R^n$ denote the retained coordinates and $y\in\mathbb R^m$ the
hidden coordinates. Constant matrices $A_0$ and $D$ govern their separate
evolution, and $B$ and $C_0$ couple them:
\begin{equation}
\dot x=A_0x+By,\qquad \dot y=C_0x+Dy.
\label{eq:supp-linear-blocks}
\end{equation}
Solving for the hidden motion and substituting the solution into the first
equation gives
\begin{equation}
\dot x(t)=A_0x(t)+Be^{Dt}y(0)
+\int_0^t Be^{D(t-s)}C_0x(s)\,ds.
\label{eq:supp-linear-memory}
\end{equation}
The first additional term carries the initial hidden state. The integral
carries feedback: $C_0$ excites the hidden motion, $e^{D(t-s)}$ propagates it,
and $B$ returns its influence to $x$. With $C_0=0$, a nonzero $B$ can transmit
an initial disturbance but produces no feedback kernel. When both couplings are
present, the reduced trajectory generally depends on its own earlier values,
even though the full evolution is a linear equation local in time. This is the
finite-dimensional form of the memory induced by projection that was discussed
by Zwanzig (1961) and Mori (1965), cited in the main text.

The number of hidden coordinates needed to reproduce this feedback can be
smaller than $m$. Define the response matrices $H_j=BD^jC_0$ and the block
Hankel matrix $\mathcal H$ with $m$ block rows and $m$ block columns:
\begin{equation}
\mathcal H_{ij}=H_{i+j},\quad 0\le i,j<m,
\qquad r=\operatorname{rank}\mathcal H.
\label{eq:supp-minimal-memory}
\end{equation}
For a finite-dimensional linear model, $r$ is the smallest number of internal
coordinates with which the transfer function $B(sI-D)^{-1}C_0$ can be realized.
This is the classical result of linear realization theory due to Ho and Kalman
(1966), cited in the main text. It counts the independent modes that are
reached from $x$ and return to $x$, with zero initial hidden state. An
arbitrary nonzero $y(0)$ may require additional coordinates to reproduce the
forcing it produces. A change of basis for the hidden coordinates changes the
matrices but preserves the response and the minimal order of its realization.

Three dimensionless systems were evaluated. All start from $x(0)=1$ and
$y(0)=0$. The first has $(A_0,B,C_0,D)=(-1,1,0,-2)$, and the second has
$(-1,0.4,0.6,-2)$. In the third,
\begin{equation}
A_0=-4,\quad B=(1,1,0),\quad C_0=(1,1,5)^{\mathsf T},
\quad D=\operatorname{diag}(-1,-2,-3).
\label{eq:supp-two-returning-modes}
\end{equation}
The third hidden coordinate is excited but has no route back to $x$, so the
feedback contains two independent exponentials rather than three. With
$P=(I,0)$ and the full generator
$L=\left(\begin{smallmatrix}A_0&B\\C_0&D\end{smallmatrix}\right)$,
the instantaneous remainder of the projection is $A_0P-PL=(0,-B)$.
Table~\ref{tab:supp-linear-closure-cases} compares this remainder with the
feedback that is actually generated.

\begin{table}[htbp]
\centering
\small
\caption{\textbf{Remainder of the projection and feedback memory in specified linear systems.}
The last column gives the largest absolute error of Eq.~\ref{eq:supp-linear-memory}
at 13 times in $0\le t\le3$, with the integral evaluated by an independent quadrature.}
\label{tab:supp-linear-closure-cases}
\begin{tabular}{lrrrr}
\toprule
System & $\|A_0P-PL\|$ & $\|BC_0\|$ & $r$ & Equation error\\
\midrule
Hidden influence without feedback & 1 & 0 & 0 & 0\\
One returning mode & 0.4 & 0.24 & 1 & $4.2\times10^{-17}$\\
Two returning modes & $\sqrt2$ & 2 & 2 & $1.9\times10^{-16}$\\
\bottomrule
\end{tabular}
\end{table}

Omitting the memory integral changes the retained trajectory by as much as
$0.0406$ and $0.0655$ in the two cases with feedback. Every eigenvalue of their
full generator has a negative real part, so all initial differences eventually
decay. A memory kernel therefore records the influence of an earlier state
without guaranteeing that the state is stored persistently.

Whether memory persists depends on the preparation and on the observable that
is measured. Consider a state with two coordinates, $q_0=(1,0)^{\mathsf T}$,
and the finite preparations
\begin{equation}
W_A=\begin{pmatrix}1&1\\0&1\end{pmatrix},\qquad
W_B=\begin{pmatrix}1&0\\1&1\end{pmatrix}.
\label{eq:supp-finite-writes}
\end{equation}
Applying them in the two orders produces the difference
$\delta q=(W_BW_A-W_AW_B)q_0=(-1,0)^{\mathsf T}$.
Under a field-free release generated by $L_r=\operatorname{diag}(0,-1)$, this
difference persists in the first coordinate. The observable $R_1=(1,0)$
measures it, whereas $R_2=(0,1)$ gives no signal for the same stored states.
If the release generator is replaced by $L_r=-I$, the difference is erased as
$e^{-t}$. At $t=30$, the magnitudes of the three measured differences are $1$, $0$ and
$9.36\times10^{-14}$, respectively. The preparations are the same in all three
cases. The relaxation spectrum decides which differences persist, and the
observable decides whether those differences can be measured.

These examples use exact matrix exponentials, with adaptive quadrature at an
absolute and relative tolerance of $10^{-12}$. The numerical rank is computed
with a relative threshold of $10^{-10}$ on the singular values. The
constructor can apply this calculation once the source equations provide a
linear generator and a proposed reduction. Distances in the atlas and
statistical labels of transitions do not by themselves specify these physical
matrices.

\clearpage
\section*{S24. Which physical coordinates a response requires}
\label{sec:supp-response-closure}

\subsection*{Reduction under a family of operations}

A reduced description identifies those full states that give the same measured
response under the operations being considered. Let $q$ denote a full state,
$U$ an allowed sequence of operations and $R$ an observable. For deterministic
dynamics the relevant equivalence is
\begin{equation}
q_1\sim q_2\quad\Longleftrightarrow\quad
RU(q_1)=RU(q_2)\quad\hbox{for every allowed }U,R.
\label{eq:supp-response-equivalence}
\end{equation}
For stochastic dynamics, the equality in Eq.~\ref{eq:supp-response-equivalence}
must hold for the conditional distributions of future measurements, not only
for their means. Adding an operation or an observable can distinguish states
that the reduction had identified. The coordinate that distinguishes them must
then be restored, or its effect must be represented by a term that depends on
the history.

For finite linear dynamics, let $L_a$ generate operation $a$, let $P$ map the
full state onto the retained coordinates, and let $R$ specify the linear
observables. Reduced generators $\overline L_a$ and observables $\overline R$
preserve the response to all piecewise-constant sequences of operations when
\begin{equation}
PL_a=\overline L_aP\quad\hbox{for every }a,\qquad
R=\overline RP.
\label{eq:supp-operation-family-closure}
\end{equation}
For $P$ of full row rank, the first identity is equivalent to the invariance of
$\ker P$ under every $L_a$. Necessity follows by differentiating
$Pe^{tL_a}=e^{t\overline L_a}P$ at $t=0$, and sufficiency follows by expanding
the matrix exponential and composing the allowed operations. The second
identity is equivalent to $\ker P\subseteq\ker R$. Autonomous dynamics of the
retained coordinates and preservation of the observable are thus separate
requirements.

For a fixed family $\Gamma$ of sequences of operations that includes the
identity, define the unobservable subspace
\begin{equation}
\mathcal N_\Gamma=\bigcap_{\gamma\in\Gamma}\ker(RU_\gamma).
\label{eq:supp-unobservable-subspace}
\end{equation}
All measured responses are equal for states with the same retained coordinates
exactly when $\ker P\subseteq\mathcal N_\Gamma$. This condition alone need not
close every coordinate retained by $P$. For example, let $\dot q_1=-q_1$,
$\dot q_2=-q_2+q_3$, $\dot q_3=-q_3$, $Pq=(q_1,q_2)$ and $Rq=q_1$. Different
initial values of $q_3$ give the same measured response
$q_1(t)=e^{-t}q_1(0)$ but different retained trajectories
$q_2(t)=e^{-t}[q_2(0)+tq_3(0)]$. Conversely, retained coordinates with
autonomous dynamics cannot predict an observable that depends directly on a
discarded coordinate. The block elimination of Section~S23 keeps both this
dependence on the preparation and the later feedback.

The closed space is built by starting from the row space of $R$, repeatedly
adding its images under each $L_a$, and computing an orthonormal basis until
the dimension stops increasing. The result is the smallest invariant row space
that contains the observables, and Eq.~\ref{eq:supp-operation-family-closure}
then holds for arbitrary initial states. A restricted set of preparations or
of observation times can admit a smaller description. This construction
applies established linear theory of observability and realization; it is used
here to identify which physical coordinates are exposed by a proposed change
of operation. The relative tolerance on the singular values in the rank
calculations is $10^{-10}$.

\subsection*{Selective exchange exposes a hidden population}

Three well-mixed compartments of equal volume contain the normalized
populations $p_0,p_1,p_2$, with $p_0+p_1+p_2=1$. Define the measured deviation
$x=p_0-1/3$ and the hidden imbalance $z=p_1-p_2$. Symmetric exchange between
compartment 0 and each of its neighbours has the rate $1\,\mathrm{s}^{-1}$, and
exchange between compartments 1 and 2 has the rate $0.7\,\mathrm{s}^{-1}$.
Raising only the 0--1 rate to $4\,\mathrm{s}^{-1}$ gives a second, selective
operation. The generators of the deviations, in $\mathrm{s}^{-1}$, are
\begin{equation}
L_{\mathrm{s}}=\begin{pmatrix}-3&0\\0&-2.4\end{pmatrix},\qquad
L_{\mathrm{a}}=\begin{pmatrix}-7.5&1.5\\4.5&-3.9\end{pmatrix},\qquad
\frac{d}{dt}\binom{x}{z}=L\binom{x}{z}.
\label{eq:supp-compartment-generators}
\end{equation}
Both operations conserve the total population and have the same uniform
equilibrium. For $R=(1,0)$ and symmetric exchange alone, the retained space has
dimension one. Including the selective exchange raises it to two, because $z$
now affects $x$. Eliminating $z$ from the selective dynamics instead gives
\begin{equation}
\dot x(t)=-7.5x(t)+1.5e^{-3.9t}z(0)
+\int_0^t6.75e^{-3.9(t-s)}x(s)\,ds,
\label{eq:supp-compartment-kernel}
\end{equation}
where time is measured in seconds and the kernel has units of
$\mathrm{s}^{-2}$. The closed description constructed numerically reproduces
the full model over 100 random sequences of six steps, with a largest
population error of $4.34\times10^{-18}$. What fixes the required extension is
the dimension of the closure, not the number of different materials.

Starting from $p_0=1$, the symmetric and selective operations are applied for
0.2 seconds each, in the two opposite orders. The resulting values of $x$ are
identical, but $z$ differs by 0.0143393. A subsequent selective exchange pulse
of 0.2 seconds converts this into a difference of 0.00146927 in $x$. After ten
seconds of symmetric release, the same probe gives only $5.55\times10^{-14}$,
because the stored imbalance has decayed at the rate $2.4\,\mathrm{s}^{-1}$.
The exchange rates in Eq.~\ref{eq:supp-compartment-generators} determine both
how this population imbalance relaxes and how much it contributes to the
measured response.

\subsection*{Origin of the interfacial-colloid construction}

The coupling to the bond frame used in the colloid example of the main text was
proposed during earlier construction work with the first version of the
representation, which records fragments of equations and the transformations between them. The constructor kept continuous angular Brownian
motion on fixed particle centres and replaced the dependence of nematic
alignment on $\theta_i-\theta_j$ by a dependence on
$\theta_i+\theta_j-2\phi_{ij}$, which the quadrupolar capillary interaction
supplies at a fluid interface. The author specified the physical restrictions
and selected the realizations. The simulations, their controls and their
results are reported in the companion manuscript~\cite{BaulinColloidMemory} and
are not repeated here.

\section*{S25. Mechanism graphs and transformations recovered from source papers}
\label{sec:supp-mechanism-graphs}

\subsection*{Physical dependencies within the nested description}

A constitutive relation determines how a current responds to a field, a boundary condition determines which solutions are admissible, and a measurement selects a quantity calculated from those solutions. Connecting these dependencies gives a graph of the physical model. In Fig.~\ref{fig:atlas}a, the vertices denote physical roles, and an equation or a set of equations is attached to each role. The arrows show which relations determine another part of the model, so they describe dependencies within one mechanism. The atlas connections in Fig.~\ref{fig:supp-roads}, in contrast, join different equation descriptions, and an explicit transformation is needed before they can be taken as a physical correspondence.

The nesting $M=(\Omega,\Xi)$, $I_{\mathrm{op}}=(M;C,R,P)$ and $I_{\mathrm{real}}=(I_{\mathrm{op}};A)$ specifies how much of this model is retained. At the level of $M$, an operation is associated with a carrier on which it can act. Adding $C,R,P$ specifies the constitutive and boundary relations, the observable, and the preparation or imposed operations. The realization $A$ is attached last, to the whole operational description, and lies outside its graph of equation dependencies. It assigns the material coefficients, the geometry of the sample and the experimental values of the initial and boundary data. For example, $C$ specifies a relation between the heat flux and the boundary temperature, whereas $A$ assigns its heat-transfer coefficient for the chosen interface. Removing these assignments makes the construction reusable, and recovering a prediction requires supplying those assignments that enter its equations. A partial mechanism is therefore an unfinished physical model whose missing relations have identifiable consequences.

The equation graphs keep the algebra and the dimensions of the scalar equations they are given, together with the assignment of each equation to a role in the mechanism. An edit by the constructor replaces specified relations and identifies the other relations that are stated to remain unchanged. Three given edits introduce a viscous internal strain into an elastic stress law, a cubic contribution to thermal relaxation, and saturation of a reaction rate. Four response trials for each edit give twelve response calculations, all of which agree with independent analytical and numerical solutions. In these examples the role assignments are given explicitly; the atlas built from the literature provides the equation features and the candidate connections. Figure~\ref{fig:atlas}a shows the general physical organization used to assemble these models, not a graph inferred independently for every equation in the archive. A direct comparison of the stored equations shows that the roles meant to stay unchanged are identical after each edit. Conservation and physical correspondence, however, depend on the equations together with their domains: changing a boundary can leave the bulk expression as it was while changing the domain and the spectrum of the complete generator.

\subsection*{When a correction can act on the retained state}

In a finite-dimensional linear model, let $L_s$ evolve the source state $q$ and let $\alpha q$ be the retained target state. An autonomous target generator $L_t$ exists on the image of $\alpha$ exactly when $\alpha L_s v=0$ for every discarded difference $v\in\ker\alpha$. The condition is necessary, because two states that differ by $v$ have the same target state and must therefore have the same target derivative. It is also sufficient, because it makes the assignment $L_t(\alpha q)=\alpha L_sq$ independent of which source state is chosen to represent $\alpha q$. A trial generator can be corrected on the retained state when its remainder satisfies the same condition. If the discarded difference changes the target derivative, no instantaneous operator on the target state can reproduce both trajectories. The missing dependence is restored by retaining the relevant hidden coordinates, or their history together with the contributions of the initial preparation, as in Sections~S23 and S24. The criterion concerns exact autonomous evolution on the allowed state space; equivalence of the observables alone, or a restricted set of preparations, can require less information.

\subsection*{Changing a current while keeping its balance law}

Let $u(\mathbf x,t)$ be a conserved density and $\mathbf j(\mathbf x,t)$ its current in a fixed domain $V$, with $D>0$ a constant diffusivity and $\tau>0$ a relaxation time of the current. The instantaneous constitutive relation $\mathbf j=-D\nabla u$ gives diffusion. Replacing it by the Cattaneo relation for the current gives\cite{Ferrillo2018Cattaneo}
\begin{equation}
\partial_t u+\nabla\cdot\mathbf j=0,
\qquad
\tau\partial_t\mathbf j+\mathbf j=-D\nabla u.
\label{eq:supp-current-construction}
\end{equation}
Differentiating the first equation in time and taking the divergence of the second eliminates the current:
\begin{equation}
\tau\partial_t^2u+\partial_tu=D\nabla^2u,
\qquad
\frac{\mathrm d}{\mathrm dt}\int_V u\,\mathrm dV
=-\int_{\partial V}\mathbf j\cdot\mathbf n\,\mathrm dS,
\label{eq:supp-current-consequence}
\end{equation}
where $\mathbf n$ is the outward unit normal. The balance law is unchanged: with zero normal current the total quantity is conserved, and an outward flux through the boundary removes it. The finite relaxation time nevertheless changes the dynamics, even with the same balance and the same boundary condition.

For smooth fields the initial density no longer determines its own future, because $\partial_tu(\mathbf x,0)=-\nabla\cdot\mathbf j(\mathbf x,0)$ also depends on the initial current. The state of the first-order system is now the pair $(u,\mathbf j)$. Solving the constitutive equation at each point in space makes the retained history explicit,
\begin{equation}
\mathbf j(\mathbf x,t)=e^{-t/\tau}\mathbf j(\mathbf x,0)
-\frac{D}{\tau}\int_0^t e^{-(t-s)/\tau}\nabla u(\mathbf x,s)\,\mathrm ds.
\label{eq:supp-current-history}
\end{equation}
Both the contribution of the initial current and the integral enter the density equation through their divergence. A divergence-free part of the initial current can therefore remain invisible to the density in the interior, while it still matters for a measurement of the current or for a boundary condition. The example shows exactly how a replacement of a constitutive relation changes the required state and its observable consequences. The graph records a connected change of closure, dynamics and preparation; treating these as independent replacements would miss the additional initial data. Equations~\ref{eq:supp-current-construction}--\ref{eq:supp-current-history} reconstruct an established model and are used here to explain the operation of the constructor. Their symbolic identities were checked.

\subsection*{Terms required by an exact change of variables}

The equations recovered from arXiv:2007.12521 give a different kind of construction: a nonlinear change of variables that preserves a stochastic process. In the Brownian normalization of the source, the modified Bessel process obeys\cite{Hufnagel2020Bessel}
\begin{equation}
\mathrm dX_t=\mathrm dB_t+
\left[\frac{\theta+1/2}{X_t}-aX_t\right]\mathrm dt,
\qquad X_t>0,\quad a>0,\quad\theta>0,
\label{eq:supp-bessel-source}
\end{equation}
where $B_t$ is a standard Brownian motion, $a$ the restoring coefficient and $\theta$ the parameter of the inverse-distance drift. We write $a$ for the source paper's $\alpha$ to distinguish it from a map between states. Setting $Y_t=X_t^2$, the quadratic variation $(\mathrm dX_t)^2=\mathrm dt$ contributes a unit drift, and
\begin{equation}
\mathrm dY_t=2\sqrt{Y_t}\,\mathrm dB_t+
(2\theta+2-2aY_t)\mathrm dt.
\label{eq:supp-bessel-target}
\end{equation}
For a smooth test observable $\phi(y)$ and $h(x)=x^2$, the source and target generators are
\begin{equation}
\begin{aligned}
L_X&=\left[\frac{\theta+1/2}{x}-ax\right]\partial_x
+\tfrac12\partial_x^2,\\
L_Y&=(2\theta+2-2ay)\partial_y+2y\partial_y^2,\\
L_X(\phi\circ h)&=(L_Y\phi)\circ h.
\end{aligned}
\label{eq:supp-bessel-map}
\end{equation}
Direct symbolic differentiation gives a zero remainder for every smooth $\phi$ in the interior of the domain. The correction can also be found without assuming its coefficient. We write $L_Y^{(0)}=(2\theta+1-2ay)\partial_y+2y\partial_y^2$ for the generator without the contribution of the variance and add an unknown constant drift $c\partial_y$. The remainder is then $(1-c)\phi'(x^2)$, which vanishes for all smooth observables exactly when $c=1$, and this recovers the drift in Eq.~\ref{eq:supp-bessel-target}. Both generators annihilate the constant observable, so the test detects a wrong evolution of the observables rather than a failure of the normalization of probability. The extra drift expresses the same fluctuations in the squared coordinate. The observable $\phi(y)=y$ gives the conditional mean
\begin{equation}
\mathbb E[Y_t\mid Y_0=y_0]
=y_0e^{-2at}+\frac{\theta+1}{a}(1-e^{-2at}).
\label{eq:supp-bessel-mean}
\end{equation}
Without the quadratic-variation term this mean is underestimated by $(1-e^{-2at})/(2a)$. An omission of the transformed diffusion is detected instead by the observable $\phi(y)=y^2$. The equations, the coordinate map and the observable together determine which terms must be carried through a construction.

The complete source equations and the stated map were recovered from the derivation that surrounds them in the source paper and were transcribed for the calculation. These equations form the source passage on which the calculation rests; the atlas edge that was first selected only located this passage and is not the verified map. The calculation establishes the local identity between the generators and the equation for the mean shown above; it does not infer global boundary conditions from proximity in the atlas.

\subsection*{Further relations recovered along the selected roads}

The sample recovered in source order contains 41 pairs of equations, which support 32 distinct atlas edges in 29 papers. Inspection and transcription by hand gave six relations that can be calculated: the local normalization of a Kac--Schwarz operator (arXiv:1405.5579), the gauge covariance of the Nahm--Schmid equations (1711.02649), a Lie-group reduction with constrained variations (1702.02911), a conservation law under the Aronsson equation (1907.07436), a finite spectral recurrence (1910.02462), and an invertible map to midpoint and half-difference variables (2501.06792). All six pass their stated symbolic checks. The Bessel transformation above was recovered separately from the surrounding source material. The other pairs of the sample were not classified, so the six relations are examples and give no rate of success for the edges of the graph.

These calculations show why a transformation must include the relations that determine it. The Nahm--Schmid change requires the derivative of the gauge transformation. The Lie-group variables satisfy $\partial_tv-\partial_su+[u,v]=0$, in which the derivative terms cancel the commutator, so variables that do not commute are still compatible. The conserved quantity of the Aronsson example requires a preceding partial differential equation. In each case the two selected displays alone leave out part of the construction. Omitting the scalar amplitude in the Kac--Schwarz normalization leaves a remainder of $-1/48$ for the tested change of coordinates, and omitting the Aronsson equation gives $\mathrm dH/\mathrm dt=-1$ along its control trajectory. Together with the unit remainder that appears when the It\^o drift is omitted, these calculations identify the particular relation on which each correspondence depends. They are checks of known relations in the source papers, obtained by explicit transcription, and not six newly generated physical mechanisms.

\section*{S26. Motions generated by cycles of reversible operations}
\label{sec:supp-lie-composition}

\subsection*{An extension assembled from two shears}

An imposed shear moves a tracer at a speed proportional to its transverse coordinate. When a second shear follows, it therefore acts on positions that the first shear has already changed. If the two shears act in different directions, reversing the applied strains around a closed cycle does not return the tracer to where it started, and a net displacement remains. The leading part of the motion produced by such sequences of controls is described by Lie brackets, as in nonholonomic motion planning~\cite{Murray1993Nonholonomic}. We use two explicit linear flows to calculate this motion and to separate its generation from its retention after the driving stops.

Let $q=(x,y)^{\mathsf T}\in\mathbb R^2$ be the tracer position divided by a reference length, and measure time in units of the inverse imposed shear rate. The dimensionless generators of horizontal and vertical shear, together with their matrix commutator, are
\begin{equation}
X=\begin{pmatrix}0&1\\0&0\end{pmatrix},\qquad
Y=\begin{pmatrix}0&0\\1&0\end{pmatrix},\qquad
Z=[X,Y]=XY-YX=\begin{pmatrix}1&0\\0&-1\end{pmatrix}.
\label{eq:supp-shear-generators}
\end{equation}
The imposed velocities of Eq.~\ref{eq:supp-shear-generators} are $Xq=(y,0)^{\mathsf T}$ and $Yq=(0,x)^{\mathsf T}$, and reversing the sign of the applied shear rate implements $-X$ or $-Y$. These prescribed velocities, linear in position, constitute the whole driving model: inertia, thermal diffusion and the finite boundaries of an apparatus are left out. Because the velocities have zero divergence, area is preserved, and the reference state $q=0$ is a stagnation point common to both flows.

For a positive strain $\epsilon$, the operations $-Y,-X,Y,X$ are applied in that chronological order, each for a time $\epsilon$. Since $X^2=Y^2=0$, each exponential ends after its linear term, and the cycle can be written exactly as
\begin{equation}
\begin{aligned}
U_\epsilon&=e^{\epsilon X}e^{\epsilon Y}e^{-\epsilon X}e^{-\epsilon Y}\\
&=\begin{pmatrix}
1+\epsilon^2+\epsilon^4&-\epsilon^3\\
\epsilon^3&1-\epsilon^2
\end{pmatrix},
\qquad \det U_\epsilon=1.
\end{aligned}
\label{eq:supp-shear-cycle}
\end{equation}
The leading displacement in Eq.~\ref{eq:supp-shear-cycle} is $\epsilon^2Zq$, which is an extension along $x$ and a compression along $y$; the cubic terms add a smaller mixing of the two coordinates. A tracer starting at $q_0=(1,0)^{\mathsf T}$ is displaced exactly by $(\epsilon^2+\epsilon^4,\epsilon^3)^{\mathsf T}$. At $\epsilon=1/4$ the horizontal displacement is $17/256=0.06640625$ and the vertical displacement is $1/64$. Running the whole cycle backwards gives $U_\epsilon^{-1}$, whose horizontal displacement is $-\epsilon^2$. The forward and reversed cycles use the same signed operations for the same durations and differ only in their order.

The two shears supplied as controls span a two-dimensional space of generators, but their repeated brackets generate the three-dimensional algebra $\mathfrak{sl}(2,\mathbb R)$. Its independent directions and defining relations are
\begin{equation}
\operatorname{Lie}\{X,Y\}=\operatorname{span}\{X,Y,Z\},
\qquad [X,Y]=Z,\quad [Z,X]=2X,\quad [Z,Y]=-2Y.
\label{eq:supp-shear-lie-closure}
\end{equation}
Equation~\ref{eq:supp-shear-lie-closure} states which further infinitesimal deformations can be produced from the shears that are available. The generators are tangent vectors at the identity of the group of area-preserving matrices, and their commutator measures, to leading order, how two local orderings of the operations differ, while staying inside the same algebra. The identity of the group is the map that leaves every state unchanged. This should not be confused with the identity of a mechanism in the main text, which concerns whether the dynamics and the observables are preserved between two physical descriptions.

\subsection*{Generated motion and the displacement that is retained}

After the cycle, let the tracer relax deterministically and overdamped according to $\dot q=Lq$, with the dimensionless time $t$ now counted from the end of the driving. We compare the cycled preparation with an uncycled one after the same interval of relaxation. If a row vector $R$ specifies the coordinate that is measured, the difference is
\begin{equation}
\delta r(t)=R e^{tL}(U_\epsilon-\mathbb I)q_0
=\epsilon^2 R e^{tL}Zq_0+O(\epsilon^3).
\label{eq:supp-lie-observed-motion}
\end{equation}
For a symmetric $L$ with no positive eigenvalue, the long-time limit $P_0=\lim_{t\to\infty}e^{tL}$ is the orthogonal projection onto the zero modes of $L$. A quadratic response therefore persists only if $RP_0Zq_0\ne0$. The matrix $RP_0Z$ decides whether any preparation at all can retain such a response, and the vector $q_0$ picks out a particular preparation. When the quadratic coefficient vanishes, higher-order motion can still remain at finite strain, and the exact cycle determines it.

A restoring harmonic force in the transverse direction gives $L=\operatorname{diag}(0,-1)$, so that $y$ decays while $x$ remains free. For $R=(1,0)$ and $q_0=(1,0)^{\mathsf T}$, Eq.~\ref{eq:supp-lie-observed-motion} gives $\delta r(t)=\epsilon^2+\epsilon^4$ at every retention time. Adding the same restoring force along $x$ gives $L=-\mathbb I$ and $\delta r(t)=(\epsilon^2+\epsilon^4)e^{-t}$. In both cases the shears and their bracket are the same, and it is the relaxation that decides whether the generated displacement survives. If the transverse relaxation is kept but the measurement is $R=(0,1)$, the result is $\delta r(t)=\epsilon^3e^{-t}$, and the persistent horizontal displacement is not seen by this measurement at all.

The persistence found here is neutral storage in a coordinate that is deterministic and unconfined; no activation barrier protects it. Thermal diffusion along that coordinate would spread the stored position, so a lasting record in a noisy material also needs a physical interaction that retains the state and a stated precision of measurement. Within the present model, identifying exactly which coordinate survives separates generation, relaxation and observation, without assuming that a nonzero bracket by itself provides memory.

\subsection*{Closure of a reduced description under all allowed operations}

A reduced state that preserves every allowed generator also preserves the motions produced by their brackets. Let $\Pi$ project the full state onto the reduced one, and suppose that $\Pi X=\overline X\Pi$ and $\Pi Y=\overline Y\Pi$. Multiplying out gives
\begin{equation}
\Pi[X,Y]=[\overline X,\overline Y]\Pi.
\label{eq:supp-lie-reduced-closure}
\end{equation}
Repeated use of Eq.~\ref{eq:supp-lie-reduced-closure} extends the relation to the whole generated algebra, so that exact closure under the available generators implies closure under all their Lie compositions. In the shear example, discarding $y$ already violates this condition for $X$: two tracers with the same $x$ but different $y$ acquire different horizontal velocities. Both coordinates are therefore needed to predict the outcome of an arbitrary later sequence of shears. Observing only $x$ after the driving has ended is a different requirement, and it is decided by the subsequent relaxation and measurement in Eq.~\ref{eq:supp-lie-observed-motion}.

\subsection*{How the constructor performs the calculation, and numerical checks}

The constructor's existing operation for composing steps was extended by a step that composes generators. Its inputs are the generator matrices with rational entries, the full linear carrier, the signed controls, the preparation, the observable and the relaxation. Exact rank calculations determine which generators are independent, and repeated commutators give a basis that is closed under the bracket. The leading response of the cycle is obtained by expanding the four exponentials through cubic order. Each finite cycle and its inverse are then passed through the existing calculation of ordered evolution, which propagates the state between every step and on into the relaxation. The matrix exponentials are compared with adaptive eighth-order Runge--Kutta integration (the Dormand--Prince scheme DOP853) at relative tolerances $10^{-7}$ and $10^{-10}$, with absolute tolerances one hundredth as large.

The shear model is computed together with four controls, at strains $1/4,1/8,1/16,1/32$ and relaxation times $0,1,5$. The controls add relaxation along the horizontal direction, measure the transverse coordinate, replace the vertical shear by a second horizontal shear at twice the rate, or start from the common stagnation point. All five constructions agree with the independent integration: the largest componentwise error in the final state, scaled by $\max(1,|q_i|)$, is $3.05\times10^{-12}$. Parallel shears commute, so their cycle is the identity, and at the stagnation point the original noncommuting shears produce no displacement. In a further check, the coordinate basis is rotated together with the generators, the preparation, the relaxation and the observable, and the predicted measured response does not change.

The reversible calculation requires both signs of each control to be admissible on the stated carrier. A backward Markov evolution, or dissipation with its sign reversed, is not accepted as an inverse operation, and general irreversible sequences are still handled by the existing calculation of forward composition. The present implementation is limited to homogeneous linear controls on at most four real coordinates and to symmetric post-drive relaxation with no positive eigenvalue. The physical generators and conditions are supplied explicitly; they are not recovered from learned covariance matrices or from fitted coordinates of the atlas.

\section*{S27. Constructing one mechanism in models from different fields}

\subsection*{Supplied models, transformations and their physical scope}

FieldBridge~\cite{FieldBridge2026} provides the model specifications, the references to the papers in which they were first stated, and the construction algorithms. The model equations are taken as inputs from those original papers, and they are separate from the experiment in which a representation learned from the corpus predicts how equations continue. The requested target, a symmetric write, consists of a single critical coordinate with supercritical cubic saturation, a swept control, a small bias and noise projected onto that coordinate. The transformations act on the named operation, carrier, physical conditions and parameters of the realization. The result of a construction is the reduced equation, its coefficients and the range over which they are calculated, together with a consequence that is then tested in the original model.

The construction detects the symmetry, continues the branch, reduces the dynamics, unfolds the reduced form where this is needed, brings it to canonical scaling and tests the law of the write. Because an unfolding can be applied at a later stage, the target can be reached even when the model has no reversing symmetry to begin with. The comparison across models from different fields can be reproduced with \href{https://github.com/synthetix-institute/fieldbridge/blob/main/docs/tutorial/23_memory_codiscovery.md}{the tutorial on one mechanism in models from different fields}.

A laser shows how the calculation proceeds~\cite{DeGiorgio1970}. With real field amplitude $E$, population inversion $N$ and positive parameters,
\begin{equation}
\dot E=\tfrac12(gN-\kappa)E,\qquad
\dot N=P-\gamma N-gNE^2.
\end{equation}
The reflection $E\to-E$ reverses the critical amplitude. At the pump $P=\gamma\kappa/g$ the amplitude becomes critical while the inversion still relaxes at a finite rate. Setting the drift of the inversion to zero gives $N=P/(\gamma+gE^2)$, and expanding about the threshold gives
\begin{equation}
\dot E=\tfrac12\left(\frac{gP}{\gamma}-\kappa\right)E
-\frac{g^2P}{2\gamma^2}E^3+O(E^5).
\end{equation}
The cubic coefficient at threshold is $-g\kappa/(2\gamma)$, and the numerical construction reproduces it. In the truncated model of convection, a reflection reverses the sense of the critical roll, and the reduction reproduces the cubic coefficient $-\sigma/[2(\sigma+1)b]$~\cite{Lorenz1963}. Both are local reductions near onset; they do not eliminate the fast variables exactly at every amplitude. Simulations of the original models test the consequence that the reductions predict.

In the genetic toggle, exchanging the two repressors reverses the antisymmetric critical direction~\cite{Gardner2000}. In a ring of four repressors, a shift around the ring reverses its alternating mode, whereas an odd ring reaches an oscillatory pair of modes instead. In the Schl\"ogl reactor~\cite{Schlogl1972}, unfolding the drift $-x^3+ax^2-k_3x+b$ reaches its triple root at $x_0=\sqrt{k_3/3}$, $a=3x_0$, $b=x_0^3$; tuning a second parameter in this way removes the quadratic term. A model with the symmetry of a square keeps two critical directions, and reducing it to a single coordinate would remove the direction that the write selects. The symmetry, the number of critical modes and the sign of the nonlinear saturation are thus separate physical requirements.

\subsection*{The probability of a write, calculated before simulation}

For the canonical swept write $\mathrm dx=(rtx-x^3+h)\mathrm dt+\sqrt{2D}\,\mathrm dW_t$, the early linear amplification gives
\begin{equation}
e^{-rt^2/2}x(t)\simeq
h\int_{-\infty}^{t}e^{-rs^2/2}\,\mathrm ds
+\sqrt{2D}\int_{-\infty}^{t}e^{-rs^2/2}\,\mathrm dW_s.
\end{equation}
The amplified variable tends to a Gaussian with mean $h\sqrt{2\pi/r}$ and variance $2D\sqrt{\pi/r}$. The cubic drift later stabilizes its sign, and the probability of the positive state is
\begin{equation}
P_+=\Phi\left(\frac{\pi^{1/4}h}{D^{1/2}r^{1/4}}\right).
\end{equation}
For a full model, the bias and the noise are projected onto its critical coordinate. The simulations integrate the original nonlinear equations, not the reduced drift. Because their bias, sweep rate and noise are finite, they depart from the limit of linear amplification, and the cubic coefficient together with the spectral gap of the transverse modes sets the extent of the regime in which the limit applies.

\subsection*{Other targets and an inverse construction}

The target of a threshold write ends an occupied stable branch at a fold. Models that cannot be reduced to the symmetric write may still admit this target. The canonical fold equation and the two constants of its Airy solution are given in the Methods. Crossing the static position of the fold, at $1.018793\,r^{2/3}$, is a different event from the later divergence of the solution at $2.33811\,r^{2/3}$. Sweeps of the full models approach the crossing law, with corrections that depend on the finite rate. The input equations, the selection of the branch and the extrapolation in rate are given in \href{https://github.com/synthetix-institute/fieldbridge/blob/main/docs/tutorial/25_memory_threshold_write.md}{the tutorial on the threshold write}. When a resolved oscillatory pair of modes is present, a description by phase applies instead; the construction of locking under weak forcing and its tests on the original models are given in \href{https://github.com/synthetix-institute/fieldbridge/blob/main/docs/tutorial/26_memory_phase_locking.md}{the tutorial on phase locking}.

The spin example solves an inverse problem, finding the interactions that satisfy a requirement, rather than reducing a supplied model to a target. For open XY chains of three to five sites with pairwise ZZ interactions, requiring exact commutation with every exchange bond that can be varied independently leaves only a uniform coupling between all pairs. The response then factorizes into an exchange amplitude and $\cos^{N-1}(2\lambda t/\hbar)$. The solver and the checks on the full Hamiltonian, including the restriction to nearest neighbours and a perturbed Ising pair, can be reproduced with \href{https://github.com/synthetix-institute/fieldbridge/blob/main/docs/tutorial/14_inverse_construction.md}{the inverse-construction tutorial}. This is a design calculated within a stated family of interactions, and the mechanism it uses is the established one of collective dephasing.

\section*{S28. Which mechanisms need a critical point}

\subsection*{Reduction near a loss of stability}

When a stable state loses stability as a control passes a critical value, the relaxation rates of a few modes
approach zero while the rates of the others stay finite. On time scales long compared with the finite relaxation times, the fast
coordinates follow the slow ones on an invariant centre manifold, and the dynamics reduce to the amplitudes of the
critical modes~\cite{GuckenheimerHolmes1983}. A near-identity change of variables then removes those terms of the
reduced drift that do not change the local dynamics. Which terms remain is fixed by the number of critical
directions, by how the symmetry acts on them and by whether the critical eigenvalues are real or
complex~\cite{GolubitskyStewartSchaeffer1988}. Table~\ref{tab:supp-normal-forms} lists the forms that occur in this
work.

\begin{table}[htbp]
\centering
\caption{Reduced forms at a loss of stability. The form is set by the critical directions, the symmetry and the
type of eigenvalue, and not by the material.}
\label{tab:supp-normal-forms}
\small
\begin{tabular}{p{0.24\linewidth}p{0.13\linewidth}p{0.27\linewidth}p{0.26\linewidth}}
\toprule
Critical directions & Symmetry & Reduced form & Mechanism \\
\midrule
One, real & none & $\dot x=\mu+x^2$ & threshold write at a fold \\
One, real; a state exists at every control value & none & $\dot x=\mu x-x^2$ & exchange of stability \\
One, real, two controls & none & $\dot x=\mu+\lambda x-x^3$ & cusp; unfolding of a fold into a two-sided write \\
One, real & $x\to-x$ & $\dot x=\varepsilon x-x^3+h$ & symmetric write at a supercritical pitchfork \\
One, real, positive cubic coefficient & $x\to-x$ & $\dot x=\varepsilon x+ax^3-x^5$ & write to a distant state at a
subcritical pitchfork \\
Two, real, related by a symmetry & square & $\dot z=\varepsilon z-\lvert z\rvert^2z+g\bar z^3$ & write into a
degenerate mode, four states for $0<\lvert g\rvert<1$ \\
Two, complex pair & phase rotation & $\dot A=(\varepsilon+i\omega)A-(1+ic)\lvert A\rvert^2A$ & onset of a limit
cycle (Hopf) \\
\bottomrule
\end{tabular}
\end{table}

\subsection*{Scaling and the ratio of noise to sweep rate}

Before rescaling, the swept symmetric write reads
\begin{equation}
\mathrm dx=(rt\,x-a_3x^3+h)\,\mathrm dt+\sqrt{2D}\,\mathrm dW_t .
\end{equation}
With $t=r^{-1/2}\tau$ and $x=r^{1/4}a_3^{-1/2}y$, it becomes
\begin{equation}
\mathrm dy=(\tau y-y^3+\tilde h)\,\mathrm d\tau+\sqrt{2\Gamma}\,\mathrm dW_\tau,\qquad
\tilde h=\frac{a_3^{1/2}h}{r^{3/4}},\qquad \Gamma=\frac{Da_3}{r}.
\label{eq:supp-scaled-write}
\end{equation}
The probability of the positive state is therefore a function of $\tilde h$ and $\Gamma$ alone. After the rescaling, a quintic term
carries the factor $r^{1/2}$, and an eliminated mode with relaxation rate $\kappa$ relaxes at
the rate $\kappa r^{-1/2}$ in the scaled time; both contributions disappear as the sweep becomes slower. In the language of the
renormalization group, the control and the bias are relevant, while higher powers and fast modes are
irrelevant~\cite{ChenGoldenfeldOono1996}. At a fixed control $\varepsilon$, the rescaling $x\propto\varepsilon^{1/2}$,
$t\propto\varepsilon^{-1}$ gives an effective noise $Da_3/\varepsilon^2$, which grows as the critical point is approached.
Noise therefore dominates where $|\varepsilon|\lesssim(Da_3)^{1/2}$, which is the analogue of the Ginzburg criterion for a system without
spatial extent. In the limit $\Gamma\to0$, linear amplification gives $P_+=\Phi(\pi^{1/4}\tilde h/\Gamma^{1/2})$
(S27)~\cite{KondepudiNelson1985}.

\subsection*{Locality, first-order transitions and several order parameters}

The reduction is valid while the control and the amplitude remain small compared with the relaxation rates of the
eliminated modes and with the scale at which higher terms become important. A slow sweep decides the sign of the amplitude while
the amplitude is still growing linearly, and hence inside this region. How long the written state is retained depends on the barrier between
the stored states, which lies outside the region; FieldBridge computes it for gradient systems from the full equations along a band relaxed from the straight
path between the states~\cite{FieldBridge2026}.

In a first-order transition the state is lost locally, at a fold or at a subcritical pitchfork, and both of these have local forms.
Where the state then jumps to, and the control at which distant states are equally stable, are properties of the full
equations. Near a tricritical point, where the cubic coefficient changes sign, the quintic form
$\dot x=\varepsilon x+ax^3-x^5$ with small positive $a$ still contains the distant branches, so a weakly first-order
transition remains local in this extended sense.

A second order parameter that remains stable is eliminated, but it modifies the cubic coefficient. For
\begin{equation}
\dot x=\varepsilon x-a_3x^3+cxy,\qquad \dot y=-\kappa y+gx^2,
\end{equation}
the slaved value $y\simeq gx^2/\kappa$ gives the effective coefficient $a_3-cg/\kappa$, which changes sign when
$cg>\kappa a_3$. A saturable absorber acts in this way at the threshold of a laser and makes the threshold first
order. When two order parameters lose stability together, both must be kept. For
\begin{equation}
\dot x=\varepsilon x-x(ax^2+by^2),\qquad \dot y=\varepsilon y-y(cx^2+dy^2),
\end{equation}
the pure states $(x^2=\varepsilon/a,\,y=0)$ and $(x=0,\,y^2=\varepsilon/d)$ are both stable, so that one mode wins, when
$c>a$ and $b>d$; when $c<a$ and $b<d$ the two modes coexist~\cite{Lamb1964}. If a symmetry relates the two
directions, the symmetry fixes the form. For three equivalent states a term $\bar z^2$ in $\dot z$ is allowed, and it makes the
transition first order~\cite{Landau1937}; the symmetry of a square forbids this term and leaves the four-state form of
Table~\ref{tab:supp-normal-forms}.

In a spatially extended system, a band of wavevectors near the critical one becomes soft, and the amplitude becomes a
field obeying an amplitude equation of Ginzburg--Landau type. Fluctuations of this field then renormalize its
coefficients; for a shell of critical wavevectors, as at the onset of convection, they make the transition weakly
first order~\cite{Brazovskii1975,SwiftHohenberg1977}. The models in this work have one or a few critical modes, and for them
the reduction with noise holds without this renormalization.

\subsection*{Mechanisms that do not need a critical point}

Writing by a sweep needs a critical point because a weak bias can select a state within the time of the sweep only where the landscape is flat; away from a critical point a persistent bias writes by activated switching. Other
mechanisms rest on properties that hold at any distance from a loss of stability, as follows.

\textit{Phase of a limit cycle.} A periodic solution shifted in time is again a solution, so the phase is an
exactly neutral direction of every stable cycle. The response of the phase to a perturbation is given by the periodic solution $Z$ of the
adjoint equation $\dot Z=-J^{\mathsf T}Z$, normalized by $Z\cdot F=\omega_0$. When a weak drive near $n$
times the natural frequency is averaged over the cycle, the phase difference obeys the Adler equation $\dot\psi=\Delta\omega-K\sin\psi$, with
$K$ set by the $n$th harmonic of the phase response along the drive~\cite{Adler1946}. This reduction requires a weak drive,
not proximity to the onset of oscillation.

\textit{Rotation and conservation.} The commutators of the terms of the Hamiltonian and of the observable generate a Lie
algebra. When this algebra is su(2), the observable rotates at a single frequency; a larger algebra gives several frequencies, and an
observable that commutes with the Hamiltonian is conserved.

\textit{Return-point memory.} Consider a network of hysteretic elements whose couplings, the drive included, all become
cooperative once the signs of some elements are reversed. In such a network the state after any closed sub-loop of the drive equals
the state before it~\cite{Sethna1993}. The condition is sufficient at every amplitude of the drive.

\textit{Return to a set point.} After a step in the input, the measured output returns to its previous steady value when the
dynamics contain an integrator of the error~\cite{Francis1976}. If the integrator leaks at a rate $\delta$, a fraction
$1/(1+kg/\delta)$ of the step remains, where $k$ is the rate of integration and $g$ the gain of the output.

\textit{Linear memory and parity.} The memory that a driven body keeps of its past inputs is set by its linear response, and a
drift that is odd in the state and the input produces no response at even degree.

Retention, by contrast, depends on the whole landscape. A record can be lost in three ways: by relaxation in a curved minimum, where the information about
the write decreases as $e^{-2\kappa t}$; by diffusion along a neutral direction, where it decreases as $1/t$; or by
activation over the barrier $\Delta V$ between the stored states, at the Kramers rate proportional to
$e^{-\Delta V/k_{\mathrm B}T}$~\cite{Kramers1940}.
\section*{S29. Changes of published models and compositions of two changes}

\subsection*{Models and changes}

The changes act on systems of ordinary differential equations extracted from the \LaTeX{} sources of arXiv papers,
each with the control parameter and the range stated in its source. Of 100 extracted models, 24 were left out. In 13
the specification carries no parameter values from the source: it has three or more numerical parameters and all of
them are set to one, so that a transition found in it is not located at values that the paper reports. Eleven have
equations with the same terms and the same signs as a model that is already present, up to the names of the
variables and of the parameters; the comparison is structural and does not establish that two models behave alike:
four copies of Lorenz's system, a second Hopf normal form, two epidemic models with three compartments, a second pair
of coupled Lorenz systems, a forest model, an Abel equation and a pendulum. Of each such group the record with the
widest control range is kept. The 76 remaining models, one representative of each structure, 59 from the construction pilot and 17 from a later extraction,
are drawn from 17 archives, namely nlin (12 models), physics (8), q-bio (7), math-ph (7), chao-dyn (6), math (6),
cs (5), cond-mat (5), stat (5), and eight other archives with one to three models each.

In the corpus, the transition from one equation to the next is labelled by the part of the equation it acts on (the
operator apparatus, the carrier geometry, the closure, the observable or the protocol order) and by whether it adds,
removes, modifies or preserves that part (S3, S9). Three of these labels can be carried out as changes of a model.
Removing operator apparatus means deleting one additive term from one equation. Adding protocol order means promoting
another parameter to the role of the control and sweeping it from a quarter to four times its published value.
Modifying protocol order means extending the range of the control by ten times its width, below the published range,
above it or on both sides. An extension stops a decade short of zero where a change of sign would change the
meaning of the parameter: where the state variables are amounts, where the parameter stands in a denominator together
with a state variable, and where it multiplies a whole equation. A fourth change, which adds a term taken from another model joined to the given one by a
road through the atlas, was implemented but could not be applied: of the 78 passages found at the far ends of the
roads, 73 do not write a closed system of ordinary differential equations.

A changed model whose trajectories run away from every initial condition is discarded: each of 32 starts grows to a
hundred times its initial amplitude, or meets a drift that is not finite, at an end of the control range; a bounded model whose attractor lies a hundred times above its starts is discarded by the same rule. Every other changed model is
analysed by the symmetric-write and threshold-write derivations of S27, each limited to 60~s of computation. A change
counts as a step when it lets the model reach a write that the original model does not reach, or when it changes the
reason why a derivation stops. FieldBridge implements the three changes, their compositions in pairs and the rule that
decides whether a pair closes; its tutorial on changes of a model works through the Hopf normal form, the Lorenz
model, the disk dynamo and the two-population epidemic model discussed in this section~\cite{FieldBridge2026}.

\subsection*{Single changes}

Of 707 changes, made up of 336 removals, 187 promotions and 184 extensions, 219 are steps. By the record of the derivations, before the checks by hand described below, nine models reach a
write through a single change (Table~\ref{tab:supp-single-changes}). A change of the protocol reaches a write in four
models and a removal does so in six; one model is reached in both ways.

\begin{table}[htbp]
\centering
\caption{Writes recorded by the derivations for a single change of a published model, before the checks by hand described in the text. Each model is counted once for each kind of change
and each kind of write.}
\label{tab:supp-single-changes}
\small
\begin{tabular}{llr}
\toprule
Change & Write reached & Models \\
\midrule
Removal of a term & symmetric & 3 \\
Removal of a term & threshold & 3 \\
Promotion of a parameter to the control & symmetric & 2 \\
Extension of the range of the control & threshold & 2 \\
Promotion of a parameter to the control & threshold & 1 \\
\bottomrule
\end{tabular}
\end{table}

The 11 repeated records were given the same changes of the protocol, each with its own control and range, because
what a protocol leaves out is a property of the protocol; among the 87 extracted records a change of the protocol is recorded
as reaching a write in five, the four above and one record of the Lorenz model. Checked by hand, two of the five
recover a transition of the system, the Lorenz record with two stable states and the R\"ossler flow with one. In the R\"ossler flow
$\dot x=-(y+z)$, $\dot y=x+ey$, $\dot z=f+xz-\mu z$, the source sweeps $\mu$ from 0.8 to 0.9; the two equilibria of
the flow exist for $\mu^2>4ef$ and appear at a fold, which lies at $\mu=0.52$ for the published $e$ and $f$ and is met
when the range is extended downwards or when $f$ is promoted to the control. Of the two equilibria one is stable
near the fold and the other is a saddle, so that the extension recovers a known transition and not a choice between
two stable states. In a one-variable model of a
cosmological attractor, $\dot z=(1+z)(6-\lambda(2+z))$, widening the range of the slope $\lambda$ of the potential is
recorded as a threshold write, but the record marks no bifurcation: the stable state $z=6/\lambda-2$ leaves to infinity
as $\lambda$ passes through zero, so that a state is lost and none is chosen. The five extracted records of the
Lorenz model come from four papers. Three declare a range of the Rayleigh parameter that contains the pitchfork at
$r=1$ where convection sets in, and the record kept for the statistics of models runs from 0 to 100. Two declare
only the chaotic range, from 27 to 29 and from 24.74 to 60; extending the second downwards meets the pitchfork and the
two stable states of steady convection. This is the example of the main text. The modified Bullard dynamo needs more care. Its specification declares $Z_0$ as the
control, although this symbol does not occur in the drift; at the zero-field equilibrium the corresponding quantity is
instead fixed as $Z_0=g/f$ by the torque parameter $g$ and the friction $f$. Sweeping an independent $Z_0$ that the
drift never uses therefore cannot probe the velocity-controlled instability of the source. Varying $g$ or $f$ does
cross the known threshold $g/f=1$ and recovers the magnetic branches associated with field reversals, so this result
concerns how the parameters are linked in the supplied specification rather than new behaviour of the dynamo. In a
symmetric two-population epidemic model, either promoting a rate to the control or removing the quadratic infection
term is recorded as giving a symmetric write. Checked by hand, the published model already has four stable states at its
published parameters, two of them asymmetric between the populations, so that the promotion adds no state that the
published control lacks. The five models that a removal alone reaches were examined with the same calculation of equilibria. In a model of a pest and a predator with additional food, removing the loss $-\delta y$ of the predator leaves two stable states for $0.05\le c\le0.2$ in the published range of the competition $c$, but the published model already has two stable states at $c=0.025$ and $0.05$, so that the removal displaces a coexistence that the model has. In the other four the changed model has no two isolated stable states: the Lotka--Volterra pair without the growth of the prey has a line of equilibria, the density of molecules in a trap without its linear loss decays to its only state, the mean-field Jaynes--Cummings model keeps families of equilibria with a zero eigenvalue and no stable one, and in the dimer the amplitude decays to zero. Of the nine models of Table~\ref{tab:supp-single-changes}, the checks by hand therefore keep one, the R\"ossler flow, as a recovered fold. The most frequent change of outcome is of a different kind: removing a term turns a stable state or a
fold into an oscillation. This happens for 26 removals in 11 models, and in each case it identifies the removed term
as the one that suppresses the oscillation, by damping it or by displacing the state.

A write reached after a change is a property of the changed model, not of the published system. For this reason the
record of each step names the term that was removed or the parameter that was changed.

\subsection*{Compositions of two changes}

Pairs are built from up to eight first changes per model, with steps taken first, and each first change is followed
by every change that is available on the changed model. Two successive extensions of the same range are excluded,
since otherwise the range would grow without limit. The outcome of a pair consists of the writes it reaches together
with the reasons why the remaining derivations stop. A pair closes when this outcome equals the outcome of the original
model or of one of its single changes, or when it combines outcomes that single changes produce. A pair is open when
it reaches a write that no single change reaches, or a reason for stopping that no single change produces. A
derivation that fails numerically does not make a pair open.

Of 2,087 pairs, 116 are open; they lie in 19 of the 76 models, and 15 of the open pairs, in five models, are recorded as reaching a write
that no single change reaches (Table~\ref{tab:supp-compositions}). The remaining open pairs end in a reason for
stopping that no single change produces: a two-dimensional critical space, a fold or an oscillation. Most open pairs
close again after a third change. Promoting a parameter closes 273 of 279 such triples, extending the range closes
114 of 124, and removing a term closes 70 of 100.

\begin{table}[htbp]
\centering
\caption{Pairs of changes whose outcome does not close on the outcomes of single changes.}
\label{tab:supp-compositions}
\small
\begin{tabular}{llrrr}
\toprule
First change & Second change & Pairs & Open & Open share \\
\midrule
Extension & removal & 367 & 40 & 0.11 \\
Removal & removal & 273 & 28 & 0.10 \\
Promotion & removal & 302 & 26 & 0.09 \\
Promotion & extension & 343 & 14 & 0.04 \\
Removal & extension & 241 & 5 & 0.02 \\
Removal & promotion & 300 & 3 & 0.01 \\
Extension & promotion & 261 & 0 & 0.00 \\
\bottomrule
\end{tabular}
\end{table}

Promoting a parameter after an extension never opens a pair, because the new control replaces the range that was
extended. The Hopf normal form shows how two removals can open a pair. Take $\dot x=\omega(\mu x-y-x\rho^2)$,
$\dot y=\omega(\mu y+x-y\rho^2)$ with $\rho^2=x^2+y^2$ and $\omega>0$. Removing only the rotation term $-\omega y$
gives two zero eigenvalues at $\mu=0$, whereas removing only the growth term $\omega\mu x$ keeps the oscillatory pair
$\pm i\omega$. Removing both gives $\dot x=-\omega x\rho^2$, $\dot y=\omega(\mu y+x-y\rho^2)$. On the invariant line
$x=0$ this system reduces exactly to $\dot y=\omega(\mu y-y^3)$. For $\mu>0$ the two written states
$(0,\pm\sqrt\mu)$ attract in the full plane, with eigenvalues $-\omega\mu$ and $-2\omega\mu$. At $\mu=0$, however,
both directions are critical, so the write class returned by the search requires a two-coordinate derivation of its
stochastic law; the deterministic bistability follows directly from the changed equations. The two deletions are fixed
and commute, so this example measures their joint effect on the dynamics and not a dependence on their order. Each of the 15 pairs was checked independently of the constructor, by finding the equilibria of the changed model
from several hundred starts and the eigenvalues of their Jacobians; a write requires two isolated stable equilibria
at one value of the control. Seven pairs have such states, all in the Hopf normal form: four pairs of removals, each of which
leaves two stable equilibria for $\mu>0$ where no single removal leaves any for $\mu<2$, and three in which a removal of a growth
term follows an extension of the range of $\mu$ upwards. The other eight do not. In a four-variable model with a discrete
symmetry (two pairs), the promotion of the self-saturation coefficient alone already gives eight stable states on the
axes, which its derivation had missed, and the removal adds none. In a two-component signalling network (three pairs)
the removals break the conservation of the two total amounts and leave a continuum of equilibria, each with a zero
eigenvalue. In a homogeneous cubic normal form (two pairs) the Jacobian annihilates every equilibrium,
$J(\mathbf{x})\mathbf{x}=3\mathbf{F}(\mathbf{x})=0$, so that no isolated stable state exists at any value of the
control. In a reduced model of a forced phase oscillator, with the detuning promoted and its range extended, there is
one stable equilibrium at most at every detuning. The scripts of this check are in the supporting files.

\subsection*{Models of quantum-computing hardware}

The same changes were applied to the 31 models of quantum-computing hardware described in the Methods. Of 415 changes, 144 are steps, and 11
models reach a write through a single change within the 60~s screen (in the five-minute survey, 10 of the 31 models
reach a write that the device itself does not have); almost all of these writes come from removing a term. The
frustrated triangles of coherent Ising machine nodes, of Kerr parametric oscillators and of rf-SQUID annealer qubits
reach the symmetric write when one coupling is removed: breaking the symmetry of the triangle lifts the
two-dimensional critical mode to the one-dimensional pitchfork. Removing one coupling term of the non-degenerate
parametric converter breaks the continuous phase symmetry of its written states and leaves two discrete states. The
parametric amplifier below threshold reaches the threshold write when its loss term is removed, which places it above
threshold. A pair of Kerr parametric oscillators reaches the symmetric write when its Kerr rotation, its coupling or
its loss is removed. FieldBridge contains the 31 hardware models, the survey of their derivations, laws and changes,
and the report of that survey, in its tutorial on the hardware of quantum computing~\cite{FieldBridge2026}.

\subsection*{What the form of a model predicts, and where the prediction fails}

Features that can be read from the equations without solving them, such as the polynomial degree, an odd symmetry, a
conserved sum or the sign of the linear term, predict the outcome of the derivations for the corpus models. The
predictions are weakest in two cases. Phase locking is predicted poorly when the drift is odd in the state
(precision 0.76), and a drift with a conserved sum needs a second parameter before it can show a two-sided write
(precision 0.75). On the hardware models, which were selected for having a classical image with a write, these rules fail. A symmetric write is reached by 58\% of the hardware models
but by only 6\% of the corpus models, and a threshold write by 74\% and 10\%, respectively. The corpus rule that a
polynomial drift has no write along its control is correct for 32\% of the hardware models, whereas simply predicting
the most frequent outcome would be correct for 94\%. Only one rule, based on the odd symmetry of the drift, keeps a
useful precision of 0.71. The form of a model therefore orders the outcomes of the derivations within one population
of models, but across populations the outcome is decided only by carrying out the derivation.

\section*{S30. Further constructions and measurements on the mathematical record}
\label{extended-sec:extended-constructions}
\subsection*{Mathematical apparatus of the equations and their neighbourhoods in the sources}

Displayed equations were extracted from 2{,}511{,}195 arXiv \LaTeX{} documents in physics, mathematics, quantitative biology and computational science, giving \HyperionCorpusRows{} equation records in all (Methods). Each record keeps the paper it comes from, its position among the equations of that paper and the text around it, and the definitions of its symbols and the values of its parameters are read from that text. A location in the index therefore leads back to equations that can be calculated. The complete models used below are assembled from these records across whole papers, and each model records which lines it uses and whether each line occurs verbatim in its source (Methods).

The index describes each record by the mathematical apparatus found in it: derivatives in time and space, integrals, spectra and eigenvalues, commutators, stochastic and control terms, and the domains, fields, lattices, graphs and phase spaces on which these act (Methods). Clustering these features gives \HyperionVTwentyOneOmegaNonNullStates{} operation classes and \HyperionVTwentyOneXiNonNullStates{} carrier classes, each with a null class for the records in which no feature was found (main Fig.~\ref{fig:atlas}c,d). A point of the atlas is a record, a cluster is a combination of features that recurs, and points placed close together within a cluster have similar features; the separations across the map are set by the layout of Supplementary Section~S8 and are not calibrated distances, and no separation measures whether two equations have the same dynamics. Within this representation the literature reuses a small number of combinations. Half of the \HyperionVTwentyOneMechanismRows{} records that carry both an operation and a carrier feature fall into \HyperionVTwentyOneMechanismCellsForHalf{} of the \HyperionVTwentyOneMechanismOccupiedFine{} occupied fine pairings, or 0.4\%, and across five fits \HyperionVTwentyOneMechanismCellsForHalfMinimum--\HyperionVTwentyOneMechanismCellsForHalfMaximum{} pairings contain half of them (Supplementary Section~S18). This concentration measures how often the same mathematical apparatus is reused, not how many distinct physical mechanisms exist: one operator acting on two domains, or under two boundary conditions, falls into a single pairing.

Because the features are read from the notation, the index groups equations by the way they are written. The abstraction from the names of variables is intended, so that an accretion rate $\dot m$ in astrophysics and the voltage derivative of a neuron model receive the same time-derivative feature. Two equations that share an index code can therefore govern different physical processes. Features triggered by the same notation, such as an expectation value and a Hilbert space written with the same brackets, occur together by construction, and a record shares its exact feature code with more than $10^5$ other records (Methods). The index therefore serves for retrieval and for the statistics of mathematical expression, while the statistics of interactions are computed on complete models.

\paragraph{Order of equations within papers.} The order in which equations appear gives a further measurement. It is made on the text index and is therefore a statement about how papers are written. All five extracted parts ($\Xi,\Omega,C,R,P$) are present in \HyperionVTwentyOneDAGLayerFive{} rows of the index (\HyperionVTwentyOneDAGLayerFiveFraction{}; Supplementary Table~\ref{tab:supp-dag-layers}). A paper can introduce its boundary conditions, observables and preparation separately from its governing equation, so recovering a model requires assembling these relations from the surrounding source material.  How a source is used physically in a construction is decided by the equations and conditions recovered for the chosen transfer, not by this statistic.

The paths of least feature change between occupied clusters, which we call the roads of the atlas, are given in Supplementary Section~S9. A road is a route through the representation; it does not transport states, observables or conserved quantities. Its use is to locate neighbourhoods in the sources where explicit transformations between equations can be looked for. In a sample of such neighbourhoods, six calculable relations were recovered from the source papers, and all six pass their symbolic checks (Supplementary Section~S25).

\subsection*{Which interaction shapes occur together in the closed models}
The compiled collection contains 2{,}745 models from 1{,}899 papers in 21 fields.
Of these, 1{,}720 contain a nonlinear term, and 1{,}432 contain a nonlinear term other than
a power of the equation's own variable. With each equation brought to the canonical form
described in Methods, the mass-action shape $xy$ occurs in 378 models from 16 fields,
$x(1-y)$ in 53 models from 12 fields, $xy^2$ in 10 fields and $\sin y$ in 9 fields.

The shapes $x(1-y)$ and $y(1-x)$ occur together in 20 models, against 0.44 expected
if the shapes were placed independently, or in 14 papers against 0.35. The shapes $xy^2$
and $x^2y$ occur together in 8 models against 0.21, or in 4 papers against 0.13.
These counts concern local expressions that appear together. Each equation is canonicalized
separately and calls its own state variable $x$, so the shapes do not keep a common
assignment of variables across the equations of a model. A genuine pair of reciprocally competing species,
$\dot x=x(1-y)$, $\dot y=y(1-x)$, gives the same local shape twice. Conversely,
$\dot x=x(1-y)$, $\dot y=z(1-y)$, $\dot z=-z$ gives both reported shapes without any
reciprocal competition. Counting physical modules would therefore require a consistent
assignment of variables across equations, together with their signs and the relations
between their parameters.

Mass action and $\sin y$ occur together in no compiled model, whereas independent placement predicts 5.6 models
or 3.9 papers. The absence extends to the whole block: none of the 22 pairs formed by the
11 mass-action shapes with the periodic shapes $\sin y$ and $\sin x$ occurs, against 19.7 co-occurrences expected. The count concerns the equations as their sources write them: with $u=\cos\varphi$ and $v=\sin\varphi$ a phase becomes two polynomial variables on the unit circle, and a model written in that form is counted among the polynomial shapes.
Periodic shapes, by contrast, combine with powers of the variables as often as chance predicts (7 against 8.6), and with each
other more often (3 against 0.33; main Fig.~\ref{fig:record}). Amounts and phases do meet in established physics,
in the bosonic Josephson junction~\cite{Smerzi1997} and in polariton condensates fed by a reservoir~\cite{Wouters2007};
the gap marks a combination that the written equations of most fields leave out. It also specifies a construction: a
product carrier $\mathbb R_+^n\times S^1$ with couplings admissible on both factors, from which the amount--phase
model of the main text follows.

\subsection*{Completing an incomplete specification}
To go back from an abstract core $M = (\Omega, \Xi)$ to a physical model, the projection $Q$ has to be reversed: one member of its fibre $Q^{-1}(M)$ must be chosen and given compatible boundary closures $C$, measured observables $R$ and protocols $P$. Constructing a physical theory from a partial description then poses three different mathematical problems:
\begin{equation}
\begin{aligned}
(\Omega,\emptyset)&\longrightarrow(\Omega,\Xi_t;C_t,R_t,P_t),\\
(\emptyset,\Xi)&\longrightarrow(\Omega_t,\Xi;C_t,R_t,P_t),\\
((\Omega,\Xi);\emptyset,\emptyset,\emptyset)&\longrightarrow((\Omega,\Xi);C,R,P).
\end{aligned}
\label{extended-eq:constructor-boundaries}
\end{equation}
In the first, $(\Omega,\emptyset) \to (\Omega,\Xi_t; C_t,R_t,P_t)$, a known dynamical operation $\Omega$, such as continuum diffusion $\nabla^2$, is carried onto a new physical carrier $\Xi_t$, such as a discrete network; this requires deriving the boundary constraints $C_t$, the observables $R_t$ and the preparation protocols $P_t$ that belong to the new carrier. In the second, $(\emptyset,\Xi) \to (\Omega_t,\Xi; C_t,R_t,P_t)$, the physical medium $\Xi$ is given, and the constructor has to derive an admissible dynamical generator $\Omega_t$ that produces a desired physical response. In the third, $((\Omega,\Xi);\emptyset,\emptyset,\emptyset) \to ((\Omega,\Xi);C,R,P)$, a dynamical core that is not yet closed receives the closures and measurement couplings needed to make its evolution determinate.

\subsection*{Projection, preparation and memory}
\begin{figure}[!htbp]
\centering
\includegraphics[width=\linewidth]{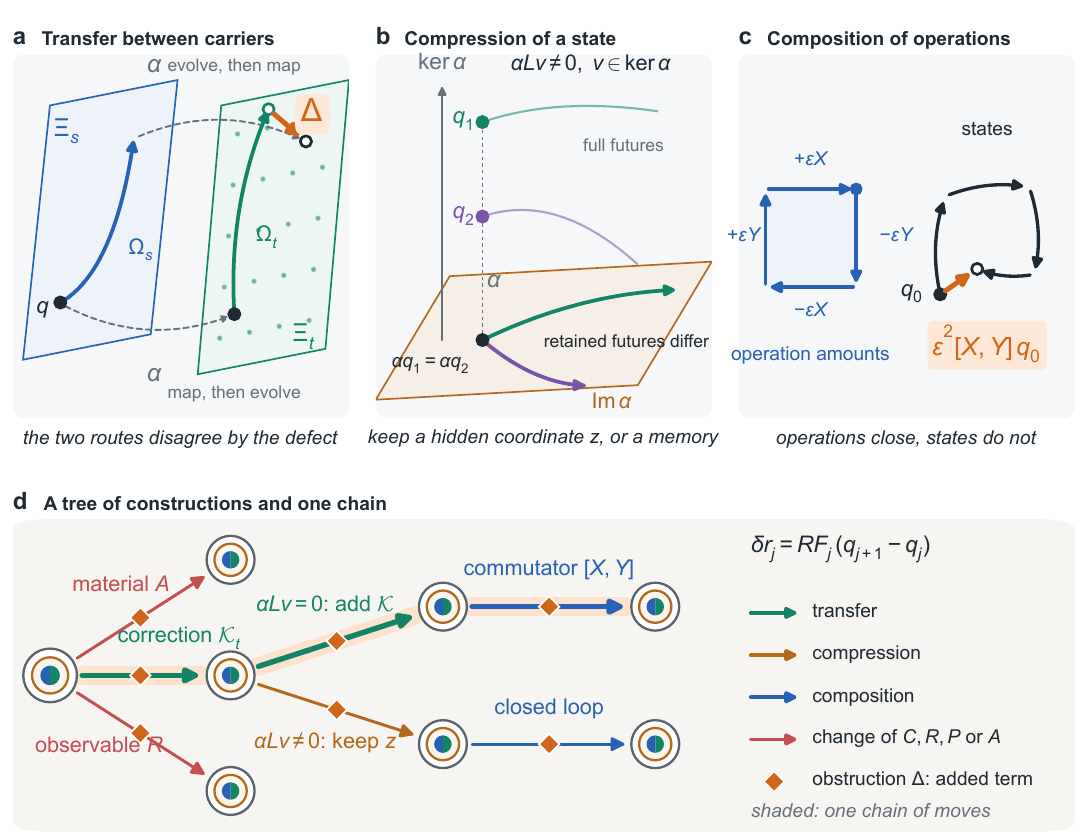}
\caption{\textbf{Compression, obstruction and composition in construction.} (a)~Transfer between carriers compares mapping then evolving with evolving then mapping. Their defect constrains a correction through $\mathcal{K}_t\alpha=-\Delta$ (Eq.~\eqref{eq:correction-constraint}). (b)~States differing along $\ker\alpha$ have equal retained images but different retained futures when $\alpha Lv\neq0$ for some discarded $v$; an additional coordinate or memory is then required. (c)~A cycle of horizontal and vertical shear returns their accumulated amounts to zero but leaves a leading displacement $\epsilon^2[X,Y]q_0$ (Methods, Eq.~\eqref{eq:shear-cycle}). (d)~Nodes represent nested identities and edges represent construction steps. An obstruction can require an additional term or retained coordinate; branches also arise from alternative completions of one core. The thick path is a chain, whose stage contribution to the final observable is $\delta r_j=RF_j(q_{j+1}-q_j)$ for fixed subsequent linear dynamics (Methods).}
\label{extended-fig:constructor}
\end{figure}

How discarding degrees of freedom produces memory, and with it an obstruction to compression, can be seen in three coupled compartments between which particles or probability diffuse (Fig.~\ref{extended-fig:constructor}b). The state is given by the occupation probabilities $(p_0, p_1, p_2)$, which evolve as $\dot{\mathbf{p}} = W \mathbf{p}$ under a symmetric rate matrix $W = W^{\mathsf T}$ with zero row sums. We use two coordinates: the departure of the central compartment from equilibrium, $x = p_0 - 1/3$, and the imbalance between the outer compartments, $z = p_1 - p_2$.

When the central compartment is coupled equally to both outer ones ($W_{01} = W_{02} = 1$, with direct exchange $W_{12} = 0.7$ between the outer compartments), the central coordinate does not depend on the outer imbalance at all: $\dot{x} = -3x$. Because of this reflection symmetry, every preparation with the same $x(0)$ gives the same central relaxation, whatever the value of $z(0)$. The outer coordinate $z$ can therefore be discarded without any loss in predicting the central observable.

Suppose now that the symmetry is broken by strengthening the coupling between the central and the first compartment ($W_{01} = 4$), while the rates stay reciprocal ($W_{ij} = W_{ji}$). The central departure $x$ is then coupled to the outer imbalance $z$ through the generator
$L = \begin{pmatrix}-7.5 & 1.5 \\ 4.5 & -3.9\end{pmatrix}$, so that $\dot{x} = -7.5x + 1.5z$ and $\dot{z} = 4.5x - 3.9z$.
Because the unobserved coordinate $z$ now feeds back into the rate of change of $x$, the dynamics projected onto $x$ alone is no longer autonomous. Integrating out $z(t)$ exactly gives an integro-differential equation for $x(t)$:
\begin{equation}
\dot x(t)=-a_xx(t)+be^{-a_zt}z(0)+\int_0^tbc\,e^{-a_z(t-s)}x(s)\,\mathrm ds,
\label{extended-eq:compartment-memory}
\end{equation}
where $a_x = 7.5$, $a_z = 3.9$, $b = 1.5$, and $c = 4.5$.

Equation~\eqref{extended-eq:compartment-memory} shows where the memory comes from. The second term is a fading memory of the unmeasured initial preparation $z(0)$. The integral is an exponential memory kernel: past values $x(s)$ act on the present through excursions of the probability into the outer compartments. The obstruction to compressing the dynamics without memory thus appears physically as a Mori--Zwanzig memory kernel~\cite{Mori1965,Zwanzig1961}.

In general, a projected state $\Pi q$ evolves autonomously under a reduced generator $\overline L_a$, and the measured observable is preserved as $R = \overline R \Pi$, if and only if the \textbf{intertwining condition} holds:
\begin{equation}
\Pi L_a=\overline L_a\Pi,\qquad R=\overline R\Pi.
\label{extended-eq:intertwining-condition}
\end{equation}
The first equality, $\Pi L_a = \overline L_a \Pi$, requires that time evolution and projection commute, which means that the discarded modes must not feed back on the retained coordinates. Unequal coupling between the compartments breaks this commutation and creates a compatibility defect, which has to be removed either by an explicit memory kernel or by restoring $z$ as a retained coordinate. The second equality, $R = \overline R \Pi$, ensures that the projected coordinate contains everything needed to predict the measurement. When a later measurement couples to degrees of freedom outside the projected subspace, additional state coordinates must be restored for that measurement. The two conditions, one on unobserved feedback and one on what is measured, together express the equivalence of predictive states for stochastic processes~\cite{Shalizi2001} and the realization theory of state spaces for dynamical systems~\cite{HoKalman1966,Petreczky2011}.

\subsection*{A change of variables in a stochastic process, and a cycle of shears}
A classic case in which the defect is removed by an instantaneous correction is a nonlinear change of variables in a stochastic process. Take a modified Bessel process, $\mathrm{d}X_t=\mathrm{d}B_t+[(\theta+1/2)/X_t-aX_t]\,\mathrm{d}t$, and transform it by $y = h(x) = x^2$~\cite{Hufnagel2020Bessel}. Pulling back the generator naively leaves the compatibility defect $\mathcal{D} = \mathcal{L}_X h^* - h^* \mathcal{L}_{Y,\mathrm{naive}} = h^* \partial_y$, which comes from the nonvanishing quadratic variation $(\mathrm{d}X_t)^2 = \mathrm{d}t$. Requiring exact intertwining with $\mathcal{L}_Y = \mathcal{L}_{Y,\mathrm{naive}} + \mathcal{K}_t$ shows that the correction is precisely the It\^o drift term, $\mathcal{K}_t = \partial_y$, which gives $\mathrm{d}Y_t = 2\sqrt{Y_t}\,\mathrm{d}B_t + (2\theta+2-2aY_t)\,\mathrm{d}t$. Leaving this term out underestimates the mean trajectory systematically by $(1-e^{-2at})/(2a)$. The correction follows from the algebraic structure of the transformation alone, without fitting any trajectory.

Beyond projections and changes of variables, applying operations that do not commute one after another in time produces a net transport that none of the operations produces by itself (Fig.~\ref{extended-fig:constructor}c). Consider a continuous medium subjected to a closed cycle of alternating horizontal and vertical shears, generated by $X = \begin{pmatrix}0&1\\0&0\end{pmatrix}$ and $Y = \begin{pmatrix}0&0\\1&0\end{pmatrix}$. The signed shear increments sum to zero over the cycle, but because the flow maps do not commute the cycle leaves a net deformation, whose leading part is a diagonal extension given by the Lie bracket $[X, Y] = \begin{pmatrix}1&0\\0&-1\end{pmatrix}$ (Methods, Eq.~\eqref{eq:shear-commutator}).
The same geometry underlies nonholonomic locomotion. A falling cat turns over with zero total angular momentum by cycling its shape, and a microorganism swims in Stokes flow at low Reynolds number by making nonreciprocal strokes, as Purcell's scallop theorem requires~\cite{Purcell1977}. In the same way, a sequence of noncommuting operations in a graph of operations moves the state along the directions of their Lie brackets. Whether this displacement survives as an observable state depends on the relaxation that follows: it persists if the generated mode projects onto the persistent zero-mode eigenspace of the subsequent relaxation operator $L$, and otherwise it decays back to thermal equilibrium.

\subsection*{Known relations recovered by the constructor}

The canonical symmetric-write equation used in these comparisons is
\begin{equation}
\mathrm dx=(\varepsilon x-x^3+h)\,\mathrm dt+\sqrt{2D}\,\mathrm dW_t.
\label{extended-eq:canonical-write}
\end{equation}

Equation~(\ref{extended-eq:canonical-write}) recurs because, near a loss of stability, the form of the reduced dynamics no
longer depends on the material. As the relaxation rate of one mode approaches zero, all other modes relax much faster
and follow it, so the model reduces to the amplitude of this slow mode, and a change of variables removes the higher
terms that do not alter the motion near the critical point~\cite{GuckenheimerHolmes1983}. What remains is fixed by the
number of critical directions, by how a symmetry acts on them, and by whether the instability is oscillatory. One real
direction gives a fold, a transcritical form if one state exists at every value of the control, or a pitchfork if a
symmetry reverses the direction. An oscillatory pair gives the Hopf form, and several directions related by a symmetry
give a form built from the invariants of that symmetry~\cite{GolubitskyStewartSchaeffer1988}. The argument has the
structure of the renormalization group. After the rescaling $x\propto\varepsilon^{1/2}$, $t\propto\varepsilon^{-1}$
and $h\propto\varepsilon^{3/2}$, the control and the bias are relevant, while the higher powers and the fast modes are
irrelevant~\cite{ChenGoldenfeldOono1996}. Noise is relevant too: in a sweep at rate $r$ it enters only through the
ratio $D|a_3|/r$, where $a_3$ is the cubic coefficient before rescaling. Equation~(\ref{extended-eq:canonical-write}) is
therefore Landau's equation for one order parameter, with noise~\cite{Landau1937}. A single mode has no spatial
fluctuations that could change its exponents, so this mean-field form holds near the critical point. In an extended
system a band of modes softens together, the amplitude becomes a field, and fluctuations can change the exponents or
even the order of the transition~\cite{SwiftHohenberg1977}.

Eight physical realizations from six scientific disciplines arrive at this symmetric write, through four different symmetry classes:
\begin{itemize}
\item In the laser model with a fixed phase and a real amplitude, the critical amplitude changes sign, which a measurement against a phase reference distinguishes and the intensity does not; in the model of convection rolls, the symmetry reverses the sense of circulation.
\item In a synthetic genetic toggle switch, the critical mode is set by the symmetry that exchanges the two repressors.
\item In a cyclic ring oscillator of four genes, the bifurcation comes from a discrete cyclic permutation that reverses the alternating pattern of expression.
\end{itemize}
When a system lacks the exact symmetry, a change of the control parameter generally meets an asymmetric saddle-node bifurcation (a fold) rather than a pitchfork. In a genetic toggle with unequal promoters, as in the reactor above, the constructor identifies the defect that breaks the symmetry and unfolds the system through a cusp catastrophe by sweeping a second physical parameter, such as the strength of a promoter or the feed rate of a chemical, which restores the symmetric bistable write. Stochastic simulations of these models give the constant of the write law, $\pi^{1/4} \approx 1.331335$: they yield $1.350 \pm 0.019$ across seven models and, in longer runs, $1.332 \pm 0.005$ across eight different models (Methods).

Like the renormalization group, the reduction is local: it holds while the control is small compared with the
relaxation rates of the eliminated modes. Locality is enough to decide which state is written, because a weak bias
can act within the time of a sweep only where the landscape is flat. How long the state is kept depends on the barrier between distant states,
which lies outside the reduction. A first-order transition is local only where the occupied state loses stability, at
a fold or a subcritical pitchfork. Where the jump lands, and where distant states are equally stable, are global
properties. The exception is the neighbourhood of the point at which the cubic coefficient changes sign, where the
quintic form $\dot x=\varepsilon x+ax^3-x^5$ still contains the distant states. Several order parameters enter in two
ways. If they become critical together, all of them are kept and their symmetry fixes the reduced form; if the
symmetry allows a quadratic term, as for three equivalent states, the transition is first
order~\cite{Landau1937}. If one of them remains stable, it is eliminated, but its coupling to the critical mode shifts
the cubic coefficient and can reverse its sign, as a saturable absorber does at the threshold of a laser.

\begin{figure}[!htbp]
\centering
\includegraphics[width=\linewidth]{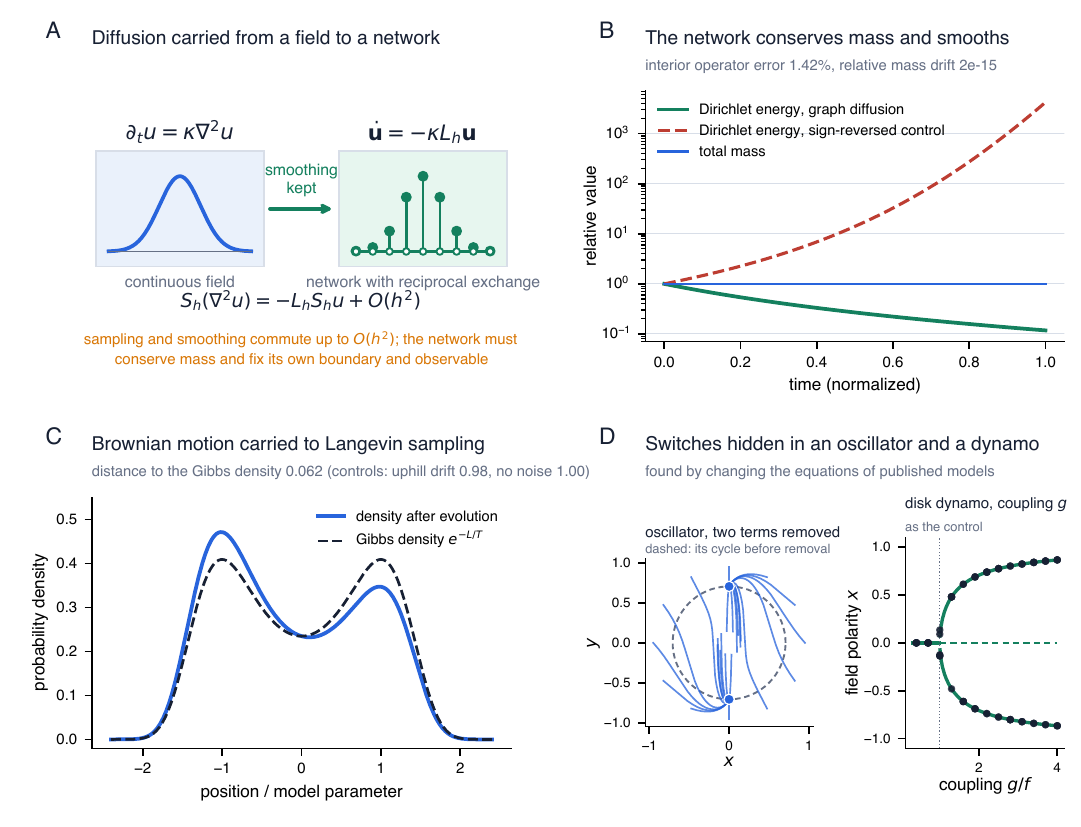}
\caption{\textbf{Transfers of the diffusion core and switches found by changing published models.} (A)~Diffusion carried from a continuous field to a network: reciprocal exchange keeps the smoothing, and sampling intertwines $\nabla^2$ with the graph Laplacian up to $O(h^2)$. (B)~The network conserves mass and its Dirichlet energy decreases; on the finite grid the error of the interior operator is 1.42\%, and a control with reversed sign makes the energy grow. (C)~Drift and fluctuations carried onto the space of parameters of a loss recover Langevin sampling, whose density approaches the Gibbs density. (D)~Changes of published models: removing two terms of an oscillator leaves two stable states in place of its cycle (dashed). In the compiled specification of Bullard's disk dynamo, promoting the driving torque to the control crosses the known threshold at which the torque balances the friction and recovers the two polarities of its field; the control that this specification declares does not occur in its equations, so the result concerns the specification and not the dynamo (S29). Equations and numerical conditions are given in Methods.}
\label{extended-fig:validation}
\end{figure}

A proposed link between two fields is accepted only by calculation. The constructor uses the equations and physical assumptions of the source to test whether the link allows an operational transfer: it calculates the defect of the operator, the error of the truncation and the boundary closures that the target medium requires. Carrying continuum diffusion onto networks and onto stochastic laws of probability tests this calculation, and it recovers the intertwining of the operators, the conservation of mass and the convergence to the equilibrium Gibbs density (Fig.~\ref{extended-fig:validation}; Methods).

In the memory mechanisms based on bifurcations that are evaluated here, writing uses the marginal stability at a critical point, where a weak signal can select among the emerging states. In other physical settings writing works instead by biased activated switching or by controlled rotations. For a weakly driven stable limit cycle, phase reduction generally gives a periodic phase-coupling function, which under resonant harmonic forcing reduces to the classical Adler equation of phase locking~\cite{Adler1946}. Other mechanisms rest on exact structural properties of the equations and need no perturbation expansion. A spin precesses at a single frequency when the commutators of its Hamiltonian terms and of its observable close into the Lie algebra $\mathfrak{su}(2)$, and an observable that commutes with the Hamiltonian is strictly conserved. In athermal disordered media with ferromagnetic interactions and no-passing dynamics, closed subloops of the drive give exact return-point memory~\cite{Sethna1993}, and a regulated output returns to its set point when internal feedback integrates the error of the control~\cite{Francis1976}. How long a stored state persists is governed by the wider potential landscape: the state decays by relaxation within a local basin, diffuses along a neutral direction that is not pinned, or escapes over barriers by thermal activation (Supplementary Section~S28).

One route to a new mechanism is to transfer a known one to another physical substrate and then complete analytically what an adiabatic reduction leaves out. The transitions recorded in the literature point to other routes. Consecutive equations in a paper differ by a few elementary transformations: an operator term is added, removed or modified, an internal coupling becomes the control parameter, the range or the sweep rate of a protocol changes, or a closure constraint is altered (Supplementary Sections~S3 and~S9). When these transformations are applied systematically to a model extracted from a paper, each gives a modified system, and the reduction determines the mechanism that emerges in it without a target chosen in advance by a person.

\subsection*{New physical hypotheses for the hardware of quantum computing}

The same procedure was applied to the hardware of quantum computing. It starts from a census of the hardware rather
than from chosen examples. Reviews of superconducting, semiconductor, trapped-ion, neutral-atom, photonic and
optomechanical platforms describe 60 hardware
elements~\cite{Krantz2019,Blais2021,Burkard2023,Bruzewicz2019,Browaeys2020,Slussarenko2019,Aspelmeyer2014}, and a
search of 53,726 abstracts on quantum computing among 2.5 million papers adds two classes that the reviews omit:
cavity magnonics and spin ensembles coupled to a cavity. Thirty-eight of these elements have a classical image with a
write (21 in full and 15 in part from the review census, together with the two classes from the abstract search), and they are represented by 31 models. The other elements, such as gates, error-correcting codes and linear
amplifiers, store or process information without a write, and the reason for excluding each one is recorded. Where an
element reaches a known mechanism, the comparison with the full model tests the transferred law and its constant. Across the completed write points of the 31 hardware records, the completed write law classifies 29 writes as made in the linear stage, five in the nonlinear stage and six as selections between stable states that the noise keeps exchanging, and one outcome is not settled at the end of the sweep: in the bistable cavity-optomechanical element, three slow modes leave 13--14\% of trajectories on the unwritten branch when the sweep ends. The 31 hardware models, their derivations and the reported comparisons can be
explored with FieldBridge~\cite{FieldBridge2026}, in its tutorial on the hardware of quantum computing.

Three of the results concern devices in current use. First, a linear sweep of a parameter through the bifurcation cusp of a driven Kerr resonator that measures a qubit shows a dynamic delay offset, $\approx 0.8$ times the resolution threshold set by the noise, at sweep rate $0.02$ and noise ratio $0.05$. Because the dispersive shift of the measured qubit enters as a bias, this delay, which depends on the protocol, introduces a systematic error in the measurement. The constructor gives a nonlinear sweep whose curvature, tuned on the full stochastic dynamics (Methods), compensates the delay for the specified protocol and noise level. Second, frustrated triangles of annealer qubits, of parametric oscillators and of the nodes of
coherent Ising machines keep a two-dimensional critical plane. The law on this plane predicts the probabilities of the six frustrated configurations to within $0.026$ of the full stochastic dynamics on the triangle of the coherent Ising machine (certified), to within $0.039$ and $0.044$ at the two frustrated write points of the triangle of Kerr parametric oscillators, and to within $0.024$ across the eight states of the rf-SQUID triangle; these deviations measure the correction from the nonlinear stage that remains. Third, where the vacuum noise of
a device with only a few photons still acts after the linear stage, as in Kerr-cat qubits at zero detuning, the
completed law fails, and the next completion is the quantum dynamics itself. For a lossy network of Kerr parametric
oscillators with one to three photons, the ground states of an antiferromagnetic triangle in a field are degenerate
but not related by any symmetry. They are sampled almost fairly. Tunnelling splits them into separate levels, and the
loss fills these levels almost as a thermal bath would, as in quantum heating~\cite{Goto2018Boltzmann}. A slow anneal
without loss therefore favours the state that tunnelling stabilizes, while the lossy network does not.

\section*{S31. Reusing a construction learned from a failed reduction}
\label{sec:constructor-learning}

The polynomial extension of the constructor takes a deterministic source vector field $F$, an
observable map $h$ and proposed equations $G$ for that observable. It calculates
$DhF-G(h)$ and looks for polynomial corrections in the retained coordinates
by solving the identities between coefficients. When a derivative cannot be expressed in
those coordinates, the search adds observable functions suggested by the monomials of the residual
or by the source coordinates, and derives their equations as well. The added functions are monomials of the source coordinates, so that two hidden variables $y$ and $z$ that enter only through $y+z$ and relax at one rate are returned as two coordinates, although the single coordinate $y+z$ also satisfies the identity.
A separate verifier, called independently, parses the expressions again and recalculates the
identity between vector fields. The method applies exactly to autonomous polynomial equations on
real Euclidean coordinates, with at most four source coordinates and a bounded
polynomial search. Stochastic conventions, periodic coordinates, boundary
constraints and non-polynomial dynamics require separate methods of calculation.

\subsection*{Selection of a construction and its withheld consequences}

A candidate is selected on development preparations and parameter settings. Its equations
and its coordinate map are then fixed before any withheld intervention is evaluated.
The full source is integrated with an eighth-order Dormand--Prince method (DOP853) and the constructed model with a fifth-order Runge--Kutta method (RK45);
refining the step size and the tolerance checks numerical convergence. The trajectory
error of each coordinate is divided by one plus the largest absolute value of the
reference. Removing the derived correction must produce an error that is resolved under the
same comparison. This normalization depends on the origin of the observable: for the hidden preparation of the table below, $\dot u=-au+by$, $\dot y=-cy$, removing the correction gives an error of $0.100$ on the development preparations with $u=x$ and of $1.25\times10^{-6}$, which is not resolved, with $u=x+10^5$ and the same dynamics. A failure on a withheld intervention ends that campaign. A persistent record
marks interventions as used before their outcomes are evaluated, and within that record it refuses
to use them again to select another construction of the same source vector field.

What is stored is the additional observable maps and the supports of the polynomials,
not numerical trajectories or generated code. On a new source model the coefficients are solved
again, and the identity of the equations and its numerical consequences
are tested again. Reuse therefore supplies a candidate construction; it does not supply its acceptance.

\subsection*{Calculations on specified equations}

The source equations in these tests were specified directly. A cubic amplitude
description of $\dot x=ax-bx^5$ acquires the missing fifth-order saturation;
its nonzero steady amplitude then scales as $(a/b)^{1/4}$ rather than as the cubic
prediction $(a/b)^{1/2}$. A related source with both cubic and fifth-order
terms tests whether the support of the correction can be reused. Other sources test an unobserved
initial preparation, a sufficient observable $y^2$ in place of the signed
coordinate $y$, and the drift of a moving origin of coordinates.

\begin{table}[htbp]
\centering
\small
\caption{Exact completions and their normalized trajectory errors on withheld interventions.
The equations are supplied cases for verification, not new physical laws. The errors were computed with Python 3.11, NumPy 1.26.0 and SciPy 1.14.1.}
\begin{tabular}{p{0.31\linewidth}p{0.45\linewidth}r}
\toprule
What the source requires & Constructed dynamics or coordinate & Error \\
\midrule
Fifth-order saturation & $\dot u=au-bu^5$ & $3.64\times10^{-12}$ \\
Transferred support & $\dot u=au-2bu^5-cu^3$ & $4.01\times10^{-12}$ \\
Hidden preparation & $\dot u=-au+by$, $\dot y=-cy$ & $5.29\times10^{-13}$ \\
Squared hidden influence & $v=y^2$, $\dot v=-2cv$ & $6.51\times10^{-13}$ \\
Moving origin & $u=x-m$, $\dot m=v$; retain $-v$ & $2.44\times10^{-13}$ \\
\bottomrule
\end{tabular}
\end{table}

With the previously calculated support available, the transferred amplitude equation needed one
candidate support, against five when the library of supports was empty. Both runs gave identical equations and the same error on the withheld interventions.
This is a count of candidates on one named transfer, not a measured general
gain in running time. The errors when the correction was omitted ranged from 0.086 to 0.730. The squared
observable shows why a completion need not recover every source coordinate:
when the retained derivative depends on $y^2$ and $\dot y=-cy$, the pair $(u,y^2)$
closes without retaining the sign of $y$.

The input specifications are supplied in the supporting files, and the records are produced from them by the comparison and learning module of the FieldBridge workbench.

\section*{S32. Analyses made with earlier versions}

Five analyses of earlier drafts used versions of the representation or of the workbench that have since been replaced,
and the main text does not rely on them; they are listed here so that references to them in earlier versions of this work can be followed. The first predicted which unoccupied operation--carrier
pairings would be filled later, using the classes of an earlier representation; it must be repeated on the current
classes before its numbers can be used. The second measured how often commutators and transformation differences are
reused in later equations of the same derivation, on an earlier sample of equations linked to their sources. The third
proposed an optical relaxation law for a hydrogel as a prospective test; it has not been measured. The fourth compared
equation proposals made by the first and second versions of the representation. The fifth searched compiled models for
dynamics that the language of the representation lacked, with an earlier version of FieldBridge; its persistent groups
were chaotic, quasi-periodic and unbounded motion, which are known dynamics.

\section*{S33. Screens of couplings, exact checks and the numerical comparison of the lock}

\subsection*{Screens of feedback and amount--phase couplings}

A separate automatic screen attached feedback and phase couplings to compiled models and asked one question: whether a specified periodic response appeared. It did not look for additional stable states, and its outcome says nothing about whether a coupling is new. A coupling was admitted only if it kept the states physical. On non-negative amounts the drift at a boundary $x_i=0$ must not point outward, $F_i(\mathbf x)|_{x_i=0}\ge0$; an added coupling $g\,x_i\cos\varphi$ vanishes there and keeps this condition wherever the source drift meets it, whereas an additive term $g\sin\varphi$ violates it for half of the phases. The multiplicative form is one admissible choice; a non-negative periodic inflow is another. The amount in turn shifted the frequency of the oscillator, and only periodic functions of the phase were used. Of the 2{,}745 compiled records, 650 had distinct source displays, 514 had connected equations and 407 could be integrated; after duplicates were removed, 292 configurations were screened, 133 on non-negative amounts and 159 on signed coordinates. Each configuration was integrated from eight preparations, together with controls in which the coupling acted in one direction only, the oscillator ran uncoupled, or only its frequency entered: 885{,}120 trajectories on graphics processors. Of 271 proposed configurations, 15 were passed to independent verification, twelve candidates for phase locking and three for oscillations produced by relaxing feedback. A candidate had to show the periodic response in at least seven of eight preparations with full feedback, in a refined integration and at a neighbouring coupling. Four verifications completed without reproducing the response, and six stopped at their time limit after results that already broke one of these conditions. Five stopped, at the time limit or at a failed integration of a control, before the decisive test; these were then evaluated under the same conditions with no time limit. Three did not reproduce the response. A cascaded Brusselator with the phase on its third species reproduced it in all eight preparations and lost it at the stronger neighbouring coupling. A linear compartment model met every condition, but the pass comes from the control and is not a lock: the model has a stable equilibrium and nothing that oscillates, and its gain-only response was still relaxing when the integration ended and closes on a periodic orbit in a run five times longer. The failed integration came from a model compiled without the simplex domain of its source, and that record was removed.

A wider re-evaluation of 7{,}376 records, which are the 2{,}745 compiled records, 3{,}996 further systems written in the same papers and 635 specifications from later batches of the same compilation, and which include configurations the first screen had left out, admitted 798 models and tested the proposed responses over matched preparations and with two solvers. Of these models, 745 were screened and 320 proposed responses tested. None of the 65 proposed additional stable responses was supported. Seven other responses survived on the sampled preparations: three are locks on models with no oscillation of their own, and four are oscillations produced by slow negative feedback on Lorenz and cubic models, a known route. The control in which the coupling acts in one direction only keeps the bare frequency of the phase, so a lock that this control misses can come from the shift of the mean frequency rather than from feedback.

\subsection*{Exact checks against statistical judges}

Every screen in this work that judged models by a statistical fingerprint produced false positives, and each was found by an exact check, which is part of the method. In a screen of single attached parts, 954 of 1{,}038 changes recorded as gaining a stored state came from attachments that the carrier does not admit: a term $1/x$ on an unconstrained coordinate, where a state beyond the pole counts as stored, or a polynomial on an angle, which unwraps the angle onto a line. A part is attached only where a denominator cannot vanish and only periodic parts act on angles; the admissible screen leaves 84 stored-state gains among 25{,}893 bounded changes. Of 27 points first labelled as transitions with several critical directions, 6 are genuine. Most of the others sample the planes of equilibria of a dissipationless top at the ends of its control range, and three are folds, read as degenerate because the tolerance on an eigenvalue grew with the slope of the branch, which diverges at a fold; the tolerance is set by the eigenvalues. Six apparent deviations from the writing law of Eq.~\eqref{eq:symmetric-write-law} came from one-sided offsets, from points where no choice exists and from a write point outside the control range; a constant is measured only after the existence of a choice has been established. The feedback screen above proposed 65 additional stable responses, none of which survived matched preparations and two solvers, and its one pass on a chemical network was detuning rather than feedback. Which bound limits a runaway follows from its degree of escape $q$, $\dot x\sim|x|^q$, because the growth supplies $\tfrac12\,\mathrm d|x|^2/\mathrm dt\sim|x|^{q+1}$: dilution at rate $\mu$ removes $\mu|x|^2$ and crowding $|x|^3/K$, so dilution holds only an exponential runaway slower than itself and turns a faster-than-linear one into a threshold, crowding limits runaways below quadratic degree and, at quadratic degree, those whose coefficient is smaller than $1/K$, and division, a jump $x\to x/2$ at size $X_c$, limits every degree with a period proportional to $X_c^{1-q}$. A threshold rule for runaway growth, found on escapes along a fixed direction, was confirmed on unseen models of that kind (median error 0.018 decades) and failed on escapes that turn or grow exponentially, and it is stated for escapes along a fixed direction. Its failure named the general statement: the threshold that dilution creates is the stable manifold of an edge state, an equilibrium of the diluted drift with a single unstable direction, of which a point on an invariant ray is the special case. Where the trajectory from just below the threshold passes such a state, the tangent plane of its stable manifold, computed without integrating any trajectory, gives the threshold with a median error below 0.01 decades in all three unseen sets. The thresholds whose trajectory passes no edge state, between a fifth and a third of the pairs in each set, are set by other invariant sets, an edge cycle or an edge at infinity. The completion of the writing law at an obstruction went through nine versions; several failed a set of models fixed before they ran, and each failure named the next correction. These checks are exact properties of the equations, so they do not depend on the judge that proposed the candidate.

\subsection*{The lock condition compared with a numerical search for equilibria}

The condition under which the constructed coupling of an amount to a phase locks is derived in the Methods for any network of amounts. The statement was also compared with the calculation used for the written terms. The comparison was specified on 10 November 2026, before any model was drawn. Templates were 1,200 written models on amount carriers with at most eight variables, drawn without replacement. The equations of a template were drawn from the written terms as in the test of saturation (Methods). One amount $x_i$, chosen at random, received the term $p\,x_i\cos\varphi$, and a phase was added with $\dot\varphi=p_1-p_2x_i$, where $p$, $p_1$ and $p_2$ are parameters of the template, signed so that the phase can lock at a positive amount; no parameter was introduced. Of the three constructed terms of this work only this one enters: the feed that follows a moving reference depends on time, and the term bound to bond angles acts on a lattice that the generator does not represent. The written set, the description and the code are those of the test of saturation. Because that code does not identify $\varphi$ with $\varphi+2\pi$, each model was computed twice: with the phase embedded as a point of the unit circle, to which the radius relaxes, so that each lock is counted once and a running phase is a cycle; and with the phase unwrapped onto a line, as the phases of written models are computed, where a lock repeats at every turn. By the rule of that specification, a sequence is absent from the record only if the same model leaves the written sequences in both representations; a sequence absent in the embedded representation alone, whose unwrapped model falls among written phase models, is the lock of a written phase model on a new carrier.

The calculation was stopped at 300~s per model, as in the test of saturation. Of the 1,200 models in the embedded representation, 244 gave a complete record, 877 were stopped by the limit (641 before the sequence of equilibria was obtained), and 79 failed for other reasons; in the unwrapped representation 193 were complete and 883 were stopped. Both representations are complete for 173 templates. The analysed models are those that the calculation finishes quickly, and a model with a fast relaxation and a slow lock is the kind that it does not; the counts below describe the analysed subset. Sequences of equilibria that no written model shows were found in 50 of the 244 embedded models and in 7 of the 193 unwrapped ones, against 184 of 2,345 models assembled from written terms alone in the same population. Among the 173 templates, none leaves the written sequences in both representations, 24 leave them only when embedded, each with its unwrapped model among written phase models, and 2 only when unwrapped. With the attractors added to the description, 2 templates leave in both representations, 36 only when embedded (34 with the unwrapped model among written phase models) and 4 only when unwrapped. The two were examined by hand. The first is $\dot T=-AT^3-mT^2+AT\cos\varphi$, $\dot\varphi=F-mT$ with $A=300$, $F=1$ and the control $m$ between 1 and 3. Its locks lie at $T_*=F/m$ with $\cos\varphi_*=T_*^2+mT_*/A$; they appear at $m=1.0017$ as a stable node and a saddle, with eigenvalues $-149$ and $-1.95$ for the node and $-153$ and $+1.90$ for the saddle at $m=2$. The code of the comparison read the node as marginal, since its slow rate is small beside the fast one, and in the unwrapped representation it located only the saddles; the sequence is the fold at which a lock appears. The second has four amounts and a phase, a drift term with a pole in one amount and no stable equilibrium, and 33--42\% of its trajectories escape. The statement of the main text is the derivation of the Methods, which holds for every model, including those stopped by the limit. The window condition was then checked model by model. For each of the 1,200 embedded models and each of the 11 values of its control, the coupled amount was set at the root of the frequency law, the other equations were solved with that amount fixed (400 starts), and a pair of locks was predicted for every solution with $|F_i|<g\,x_i$, with no integration. The derivation classifies every model: 575 have no steady state at any of the 11 values of the control, so that no lock with constant amounts exists in them, 242 have a lock at every one of these values, and in 382 an edge of the window lies inside the range; one model has a pole at the predicted amount. The code of the comparison located equilibria at each control value in 485 models, counting those completed with a limit of 7,200~s on a subsample of the stopped ones. At 5,319 of these 5,335 control values a steady state exists where one is predicted and is absent where none is; the 16 exceptions belong to two models, one with a set point of $10^{-20}$ and one whose drift contains the square root of a variable. Where at most four steady states are predicted and none lies on the edge of the window, their number agrees at 4,267 of 4,363 control values. At 725 control values the drawn coefficients place a lock on the edge of the window, $|\cos\varphi_*|=1$, where the two locks coincide, and the code records it as marginal. Of the 107 models that the code finished without a complete sequence, 101 have equilibria at the number of control values predicted, 79 of them at none. Of the 824 models that the limit stopped, 475 are predicted to have no lock at any value of the control.


\begin{thebibliography}{999}
\bibitem{Anderson1972} P. W. Anderson, More is different, \emph{Science} \textbf{177}, 393--396 (1972).

\bibitem{YangMills1954} C. N. Yang and R. L. Mills, Conservation of isotopic spin and isotopic gauge invariance, \emph{Phys. Rev.} \textbf{96}, 191--195 (1954).

\bibitem{AharonovBohm1959} Y. Aharonov and D. Bohm, Significance of electromagnetic potentials in the quantum theory, \emph{Phys. Rev.} \textbf{115}, 485--491 (1959).

\bibitem{Wigner1960} E. P. Wigner, The unreasonable effectiveness of mathematics in the natural sciences, \emph{Commun. Pure Appl. Math.} \textbf{13}, 1--14 (1960).

\bibitem{Fourier1822} J. Fourier, \emph{Th\'eorie analytique de la chaleur} (Firmin Didot, Paris, 1822).

\bibitem{Fick1855} A. Fick, Ueber Diffusion, \emph{Ann. Phys.} 170, 59--86 (1855).

\bibitem{Einstein1905} A. Einstein, \"Uber die von der molekularkinetischen Theorie der W\"arme geforderte Bewegung von in ruhenden Fl\"ussigkeiten suspendierten Teilchen, \emph{Ann. Phys.} 322, 549--560 (1905).
\setlength{\itemsep}{1.5pt}

\bibitem{Kadanoff1966} L. P. Kadanoff, Scaling laws for Ising models near $T_c$, \emph{Physics} \textbf{2}, 263--272 (1966).

\bibitem{Wilson1971} K. G. Wilson, Renormalization group and critical phenomena. I. Renormalization group and the Kadanoff scaling picture, \emph{Phys. Rev. B} \textbf{4}, 3174--3183 (1971).

\bibitem{WilsonKogut1974} K. G. Wilson and J. Kogut, The renormalization group and the epsilon expansion, \emph{Phys. Rep.} \textbf{12}, 75--199 (1974).

\bibitem{Fortunato2018Science} S. Fortunato \emph{et al.}, Science of science, \emph{Science} \textbf{359}, eaao0185 (2018).

\bibitem{Lo2020S2ORC} K. Lo, L. L. Wang, M. Neumann, R. Kinney and D. S. Weld, S2ORC: The Semantic Scholar Open Research Corpus, \emph{Proc. ACL} 4969--4983 (2020).

\bibitem{Huh2024Platonic} M. Huh, B. Cheung, T. Wang and P. Isola, The platonic representation hypothesis, in \emph{Proc. 41st Int. Conf. Machine Learning}, PMLR \textbf{235} (2024); arXiv:2405.07987.

\bibitem{Balestriero2026Platonic} R. Balestriero and Y. LeCun, Proving and improving the platonic representation hypothesis with JEPAs, arXiv:2610.39302 (2026).

\bibitem{Schmidt2009SR} M. Schmidt and H. Lipson, Distilling free-form natural laws from experimental data, \emph{Science} \textbf{324}, 81--85 (2009).

\bibitem{Brunton2016SINDy} S. L. Brunton, J. L. Proctor and J. N. Kutz, Discovering governing equations from data by sparse identification of nonlinear dynamical systems, \emph{Proc. Natl. Acad. Sci. U.S.A.} \textbf{113}, 3932--3937 (2016).

\bibitem{Raissi2019PINN} M. Raissi, P. Perdikaris and G. E. Karniadakis, Physics-informed neural networks: A deep learning framework for solving forward and inverse problems involving nonlinear partial differential equations, \emph{J. Comput. Phys.} \textbf{378}, 686--707 (2019).

\bibitem{Cornelio2023Descartes} C. Cornelio \emph{et al.}, Combining data and theory for derivable scientific discovery with AI-Descartes, \emph{Nat. Commun.} \textbf{14}, 1777 (2023).

\bibitem{RomeraParedes2024Nature} B. Romera-Paredes \emph{et al.}, Mathematical discoveries from program search with large language models, \emph{Nature} \textbf{625}, 468--475 (2024).

\bibitem{Lorenz1963} E. N. Lorenz, Deterministic nonperiodic flow, \emph{J. Atmos. Sci.} \textbf{20}, 130--141 (1963).

\bibitem{Schlogl1972} F. Schl\"ogl, Chemical reaction models for non-equilibrium phase transitions, \emph{Z. Phys.} \textbf{253}, 147--161 (1972).

\bibitem{Gardner2000Toggle} T. S. Gardner, C. R. Cantor and J. J. Collins, Construction of a genetic toggle switch in \emph{Escherichia coli}, \emph{Nature} \textbf{403}, 339--342 (2000).

\bibitem{Briat2016} C. Briat, A. Gupta and M. Khammash, Antithetic integral feedback ensures robust perfect adaptation in noisy biomolecular networks, \emph{Cell Syst.} \textbf{2}, 15--26 (2016).

\bibitem{Aoki2019Integral} S. K. Aoki, G. Lillacci, A. Gupta, A. Baumschlager, D. Schweingruber and M. Khammash, A universal biomolecular integral feedback controller for robust perfect adaptation, \emph{Nature} \textbf{570}, 533--537 (2019).

\bibitem{Adler1946} R. Adler, A study of locking phenomena in oscillators, \emph{Proc. IRE} \textbf{34}, 351--357 (1946).

\bibitem{FieldBridge2026} FieldBridge, Computational workbench for mechanism construction across scientific fields (2026). \href{https://github.com/synthetix-institute/fieldbridge}{github.com/synthetix-institute/fieldbridge}; \href{https://synthetix-institute.github.io/fieldbridge/}{interactive demonstration}; \href{https://github.com/synthetix-institute/fieldbridge/blob/main/docs/tutorial/index.md}{tutorial and executable examples}.

\bibitem{Welling2011SGLD} M. Welling and Y. W. Teh, Bayesian learning via stochastic gradient Langevin dynamics, \emph{Proc. ICML} 681--688 (2011).

\bibitem{Shalizi2001} C. R. Shalizi and J. P. Crutchfield, Computational mechanics: Pattern and prediction, structure and simplicity, \emph{J. Stat. Phys.} \textbf{104}, 817--879 (2001).

\bibitem{HoKalman1966} B. L. Ho and R. E. Kalman, Effective construction of linear state-variable models from input/output functions, \emph{Regelungstechnik} \textbf{14}, 545--548 (1966).

\bibitem{Petreczky2011} M. Petreczky, L. Bak\'o and J. H. van Schuppen, Realization theory of discrete-time linear switched systems, arXiv:1103.1343 (2011).

\bibitem{Mori1965} H. Mori, Transport, collective motion, and Brownian motion, \emph{Prog. Theor. Phys.} \textbf{33}, 423--455 (1965).

\bibitem{Zwanzig1961} R. Zwanzig, Memory effects in irreversible thermodynamics, \emph{Phys. Rev.} \textbf{124}, 983--992 (1961).

\bibitem{Hufnagel2020Bessel} N. Hufnagel and J. H. C. Woerner, Martingale estimation functions for Bessel processes, arXiv:2007.12521 (2020).

\bibitem{Murray1993Nonholonomic} R. M. Murray and S. S. Sastry, Nonholonomic motion planning: Steering using sinusoids, \emph{IEEE Trans. Automat. Control} \textbf{38}, 700--716 (1993).

\bibitem{RemleinSeifert2024} B. Remlein and U. Seifert, Nonequilibrium fluctuations of chemical reaction networks at criticality: The Schl\"ogl model as paradigmatic case, \emph{J. Chem. Phys.} \textbf{160}, 134103 (2024).
\bibitem{BaulinColloidMemory} V. A. Baulin, Geometry-Encoded Hidden Orientational Memory, arXiv:2610.19090 (2026). \href{https://arxiv.org/abs/2610.19090}{arxiv.org/abs/2610.19090}.

\bibitem{KondepudiNelson1985} D. K. Kondepudi and G. W. Nelson, Weak neutral currents and the origin of biomolecular chirality, \emph{Nature} \textbf{314}, 438--441 (1985).

\bibitem{Rossler1976} O. E. R\"ossler, An equation for continuous chaos, \emph{Phys. Lett. A} \textbf{57}, 397--398 (1976).

\bibitem{Gardner2005} F. M. Gardner, \emph{Phaselock Techniques}, 3rd ed. (Wiley, Hoboken, NJ, 2005).

\bibitem{Smerzi1997} A. Smerzi, S. Fantoni, S. Giovanazzi and S. R. Shenoy, Quantum coherent atomic tunneling between two trapped Bose--Einstein condensates, \emph{Phys. Rev. Lett.} \textbf{79}, 4950--4953 (1997).

\bibitem{Wouters2007} M. Wouters and I. Carusotto, Excitations in a nonequilibrium Bose--Einstein condensate of exciton polaritons, \emph{Phys. Rev. Lett.} \textbf{99}, 140402 (2007).

\bibitem{GuckenheimerHolmes1983} J. Guckenheimer and P. Holmes, \emph{Nonlinear Oscillations, Dynamical Systems, and Bifurcations of Vector Fields} (Springer, New York, 1983).

\bibitem{Bender2021Parrots} E. M. Bender, T. Gebru, A. McMillan-Major and S. Shmitchell, On the dangers of stochastic parrots: can language models be too big?, in \emph{Proc. 2021 ACM Conference on Fairness, Accountability, and Transparency} 610--623 (2021).

\bibitem{Buckingham1914} E. Buckingham, On physically similar systems; illustrations of the use of dimensional equations, \emph{Phys. Rev.} \textbf{4}, 345--376 (1914).

\bibitem{Lin2009Synthetic} Y.-J. Lin, R. L. Compton, K. Jim\'enez-Garc\'ia, J. V. Porto and I. B. Spielman, Synthetic magnetic fields for ultracold neutral atoms, \emph{Nature} \textbf{462}, 628--632 (2009).

\bibitem{Kirkpatrick1977} S. Kirkpatrick, Frustration and ground-state degeneracy in spin glasses, \emph{Phys. Rev. B} \textbf{16}, 4630--4641 (1977).

\bibitem{Haberman1979} R. Haberman, Slowly varying jump and transition phenomena associated with algebraic bifurcation problems, \emph{SIAM J. Appl. Math.} \textbf{37}, 69--106 (1979).

\bibitem{Jung1990} P. Jung, G. Gray, R. Roy and P. Mandel, Scaling law for dynamical hysteresis, \emph{Phys. Rev. Lett.} \textbf{65}, 1873--1876 (1990).

\bibitem{Ferrillo2018Cattaneo} F. Ferrillo, R. Spigler and M. Concezzi, Comparing Cattaneo and fractional derivative models for heat transfer processes, \emph{SIAM J. Appl. Math.} \textbf{78}, 1450--1469 (2018), \href{https://doi.org/10.1137/17M1135918}{doi:10.1137/17M1135918}.

\bibitem{DeGiorgio1970} V. DeGiorgio and M. O. Scully, Analogy between the laser threshold region and a second-order phase transition, \emph{Phys. Rev. A} \textbf{2}, 1170--1177 (1970).

\bibitem{Gardner2000} T. S. Gardner, C. R. Cantor and J. J. Collins, Construction of a genetic toggle switch in \emph{Escherichia coli}, \emph{Nature} \textbf{403}, 339--342 (2000).

\bibitem{GolubitskyStewartSchaeffer1988} M. Golubitsky, I. Stewart and D. G. Schaeffer, \emph{Singularities and Groups in Bifurcation Theory}, Vol. II (Springer, New York, 1988).

\bibitem{ChenGoldenfeldOono1996} L.-Y. Chen, N. Goldenfeld and Y. Oono, Renormalization group and singular perturbations: multiple scales, boundary layers, and reductive perturbation theory, \emph{Phys. Rev. E} \textbf{54}, 376--394 (1996).

\bibitem{Lamb1964} W. E. Lamb, Jr., Theory of an optical maser, \emph{Phys. Rev.} \textbf{134}, A1429--A1450 (1964).

\bibitem{Landau1937} L. D. Landau, On the theory of phase transitions, \emph{Zh. Eksp. Teor. Fiz.} \textbf{7}, 19--32 (1937).

\bibitem{Brazovskii1975} S. A. Brazovskii, Phase transition of an isotropic system to a nonuniform state, \emph{Sov. Phys. JETP} \textbf{41}, 85--89 (1975).

\bibitem{SwiftHohenberg1977} J. Swift and P. C. Hohenberg, Hydrodynamic fluctuations at the convective instability, \emph{Phys. Rev. A} \textbf{15}, 319--328 (1977).

\bibitem{Sethna1993} J. P. Sethna, K. Dahmen, S. Kartha, J. A. Krumhansl, B. W. Roberts and J. D. Shore, Hysteresis and hierarchies: dynamics of disorder-driven first-order phase transformations, \emph{Phys. Rev. Lett.} \textbf{70}, 3347--3350 (1993).

\bibitem{Francis1976} B. A. Francis and W. M. Wonham, The internal model principle of control theory, \emph{Automatica} \textbf{12}, 457--465 (1976).

\bibitem{Kramers1940} H. A. Kramers, Brownian motion in a field of force and the diffusion model of chemical reactions, \emph{Physica} \textbf{7}, 284--304 (1940).

\bibitem{Purcell1977} E. M. Purcell, Life at low Reynolds number, \emph{Am. J. Phys.} 45, 3--11 (1977).

\bibitem{Krantz2019} P. Krantz, M. Kjaergaard, F. Yan, T. P. Orlando, S. Gustavsson and W. D. Oliver, A quantum engineer's guide to superconducting qubits, \emph{Appl. Phys. Rev.} 6, 021318 (2019).

\bibitem{Blais2021} A. Blais, A. L. Grimsmo, S. M. Girvin and A. Wallraff, Circuit quantum electrodynamics, \emph{Rev. Mod. Phys.} 93, 025005 (2021).

\bibitem{Burkard2023} G. Burkard, T. D. Ladd, A. Pan, J. M. Nichol and J. R. Petta, Semiconductor spin qubits, \emph{Rev. Mod. Phys.} 95, 025003 (2023).

\bibitem{Bruzewicz2019} C. D. Bruzewicz, J. Chiaverini, R. McConnell and J. M. Sage, Trapped-ion quantum computing: progress and challenges, \emph{Appl. Phys. Rev.} 6, 021314 (2019).

\bibitem{Browaeys2020} A. Browaeys and T. Lahaye, Many-body physics with individually controlled Rydberg atoms, \emph{Nat. Phys.} 16, 132--142 (2020).

\bibitem{Slussarenko2019} S. Slussarenko and G. J. Pryde, Photonic quantum information processing: a concise review, \emph{Appl. Phys. Rev.} 6, 041303 (2019).

\bibitem{Aspelmeyer2014} M. Aspelmeyer, T. J. Kippenberg and F. Marquardt, Cavity optomechanics, \emph{Rev. Mod. Phys.} 86, 1391--1452 (2014).

\bibitem{Goto2018Boltzmann} H. Goto, Z. Lin and Y. Nakamura, Boltzmann sampling from the Ising model using quantum heating of coupled nonlinear oscillators, \emph{Sci. Rep.} 8, 7154 (2018).

\end{thebibliography}
\end{document}